%% file: acl_latex.tex
\pdfoutput=1
\PassOptionsToPackage{table}{xcolor}
\documentclass[11pt]{article}

\usepackage[final]{acl}

\usepackage{times}
\usepackage{latexsym}

\usepackage[T1]{fontenc}
\usepackage[utf8]{inputenc}

\usepackage{microtype}

\usepackage{inconsolata}

\usepackage{graphicx}

\usepackage{hyperref}
\usepackage{url}

\usepackage{amsmath}
\usepackage{amsfonts}
\usepackage{multirow}
\usepackage{booktabs,arydshln}
\usepackage{algorithm}
\usepackage{algorithmic}
\usepackage{tabu}
\usepackage{amssymb}
\usepackage{subfigure}
\usepackage{nicematrix}

\makeatletter
\def\adl@drawiv#1#2#3{%
        \hskip.5\tabcolsep
        \xleaders#3{#2.5\@tempdimb #1{1}#2.5\@tempdimb}%
                #2\z@ plus1fil minus1fil\relax
        \hskip.5\tabcolsep}
\newcommand{\cdashlinelr}[1]{%
  \noalign{\vskip\aboverulesep
           \global\let\@dashdrawstore\adl@draw
           \global\let\adl@draw\adl@drawiv}
  \cdashline{#1}
  \noalign{\global\let\adl@draw\@dashdrawstore
           \vskip\belowrulesep}}
\makeatother

\newcommand{\rom}[1]{\uppercase\expandafter{\romannumeral #1\relax}}

\title{Group then Scale: Dynamic Mixture-of-Experts Multilingual Language Model}

\author{Chong Li, Yingzhuo Deng, Jiajun Zhang, Chengqing Zong\footnotemark[1] \\
        State Key Laboratory of Multimodal Artificial Intelligence Systems, \\
        Institute of Automation, CAS, Beijing, China\\
        School of Artificial Intelligence, University of Chinese Academy of Sciences, Beijing, China\\
        \{lichong2021, dengyingzhuo2024\}@ia.ac.cn, \{jjzhang, cqzong\}@nlpr.ia.ac.cn
        }

\begin{document}
\maketitle

\renewcommand{\thefootnote}{\fnsymbol{footnote}} %将脚注符号设置为fnsymbol类型，即特殊符号表示
\footnotetext[1]{Corresponding author.}

\renewcommand{\thefootnote}{\arabic{footnote}}

\begin{abstract}
% Large Language Model (LLM) has shown impressive multilingual performance through pre-training on large-scale multilingual corpus. 
The curse of multilinguality phenomenon is a fundamental problem of multilingual Large Language Models (LLMs), where the competition between massive languages results in inferior performance. 
It mainly comes from limited capacity and negative transfer between dissimilar languages. 
% Researchers mainly attribute it to the limited capacity of model. 
To address this issue, we propose a method to dynamically group and scale up the parameters of multilingual LLM while boosting positive transfer among similar languages. 
Specifically, the model is first tuned on monolingual corpus to determine the parameter deviation in each layer and quantify the similarity between languages. 
Layers with more deviations are extended to mixture-of-experts layers to reduce competition between languages, where one expert module serves one group of similar languages. 
Experimental results on 18 to 128 languages show that our method reduces the negative transfer between languages and significantly boosts multilingual performance with fewer parameters. 
Such language group specialization on experts benefits the new language adaptation and reduces the inference on the previous multilingual knowledge learned. \footnote{Our code and model weights are available at \href{https://github.com/ZNLP/DMoE}{https://git\\hub.com/ZNLP/DMoE}}

% After tuning LLM on monolingual cethod, orpus, we find a hourglass shape distribution for parameter deviation. 
\end{abstract}

\section{Introduction}
After training on the massive multilingual corpus, large language models obtain impressive multilingual abilities, e.g., cross-lingual natural language understanding \citep{xue-etal-2021-mt5} and in-context learning \citep{lin-etal-2022-shot, workshop2023bloom, wei2023polylm, anil2023palm2, ustun-etal-2024-aya}. 
However, their performance in medium- to low-resource languages, still lags behind that of high-resource languages \citep{lai-etal-2023-okapi, asai-etal-2024-buffet, li-etal-2024-improving-context}, and is hindered by the \textit{curse of multilinguality} phenomenon \citep{aharoni-etal-2019-massively, wu-dredze-2020-languages}. 
\begin{figure}[th]
\centering
\subfigure[Calculate the parameter deviation for each language.]{\label{fig:delta_dist}\includegraphics [width=0.48\textwidth]{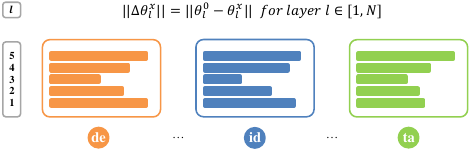}}
\vspace{-1mm}
\subfigure[Dynamically scale up multilingual LLM.]{\label{fig:dyn_scale}\includegraphics [scale=0.98]{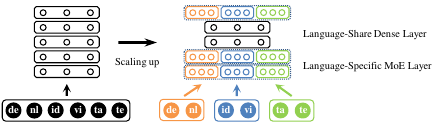}}
\vspace{-2mm}
\caption{(a) We first statisticize layer-wise parameter deviation of the multilingual language model for each language, (b) then dynamically scale up layers with more deviations into mixture-of-experts layers for language groups divided by language similarity.}
\vspace{-5mm}
\end{figure}
It is found that the limited capacity and negative language transfer mainly contribute to the curse of multilinguality phenomenon \citep{chang-etal-2024-multilinguality}. 
Thus, our key research problem lies on: \textit{How to flexibly augment the capacity of LLM for massive languages?}

To address the research problem above, \citet{pfeiffer-etal-2022-lifting} fine-tuned a module for each new language to augment parameters. 
The additional language identification process hinders its general application and affects the inference performance if misclassification. 
\citet{blevins-etal-2024-breaking} trained models for new languages using the multilingual base model as initialization, and assembled them with vanilla models during inference. 
It largely increases the inference cost and the amount of model parameters which linearly grows with the number of languages involved. 

In contrast, we introduce language specialization to the mixture-of-experts structure to scale up the parameters of the model. 
% We first investigate the impact of additional corpora on the parameters of LLM. 
In particular, monolingual corpus from each language $x$ is first adopted to tune the model and obtain the layer-wise parameter deviation $\Delta \theta^{\textit{x}}_l$ like Figure \ref{fig:delta_dist}. 
Layers near the input and output of LLM are often found with more derivation than the others (refer to Figure \ref{fig:delta_bloom} in Appendix \ref{app:lang_delta} for more details). 
We argue that layers with more deviation require more capacity to contain language-specific knowledge, while the other layers can be shared with all languages, like the ``concept space'' in the multilingual LLM \citep{wendler-etal-2024-llamas}. 
Thus, the former is extended to the mixture-of-experts layer, and the parameter of each expert is tuned by a group of similar languages like Figure \ref{fig:dyn_scale}. 
It aims to precisely exploit parameters during scaling up and keep a similar inference cost for each token. 
Such designation is also beneficial for extending new languages while reducing the effect on the previously learned languages. 
Given a new language to adapt, we first determine its similarity between existing language groups, then copy and fine-tune the expert for the most similar language group to achieve a better transferring performance and alleviate catastrophic forgetting. 
The experimental results on 18 to 128 languages show that our method significantly improves multilingual performance and mitigates the curse of multilinguality phenomenon. 
The improvement in perplexity reaches 11.4\% over the continual pre-training method and even surpasses X-ELM \citep{blevins-etal-2024-breaking} 9.6\% with 3.6x fewer parameters on average. 
In summary, our contributions lie in the following:
% Advantages: 
%   1. Dynamic Grouping 
%   2. Dynamic Extension 
%   3. Dynamic Adaptation
\begin{itemize}
    \item We propose a mixture-of-experts training framework to flexibly group languages and dynamically augment the capacity of multilingual large language models. 
    \item We formalize language grouping into a maximin optimization problem and introduce a token-level language classification loss to specialize mixture-of-experts layers. 
    % \item We empirically find that language 
    \item Extensive experiments on 18 to 128 languages demonstrate the effectiveness of our method which largely mitigates the curse of multilinguality phenomenon. 
\end{itemize}

\section{Related Works}

\subsection{Quantify Language Similarity}
The LANG2VEC method \citep{littell-etal-2017-uriel} represents languages as typological, geographical, and phylogenetic vectors to calculate the similarity between them and has been widely adopted \citep{blevins-etal-2024-breaking, chang-etal-2024-multilinguality}. However, they rely exclusively on language- or data-intrinsic features, ignoring the characteristics of the downstream models. To address this limitation, prior works have proposed model-specific representations, such as learnable language vectors \citep{tsvetkov-etal-2016-polyglot, ostling-tiedemann-2017-continuous, johnson-etal-2017-googles} and leveraging hidden states of the model \citep{malaviya-etal-2017-learning} or gradients of the loss function \citep{wang2022parameter} to derive language representations. These model-specific approaches often require training from scratch or incur high computational costs by recalculating similarity during training. In contrast, our method utilizes parameter deviations as language representations, enabling stable similarity estimation through fine-tuning the downstream model on a small dataset in the preparatory phase.

\subsection{Mixture of Experts}
% TODO Chong
Since the concept of mixture-of-experts proposed \citep{jacobs1991adaptive, jordan1994hierarchical}, it has been widely applied to SVM \citep{Ronan2001svm}, Gaussian process \citep{Volker2001Gaussian}, Dirichlet process \citep{shahbaba2009nonlinear}, LSTM \citep{Lucas2015Generative, shazeer2017out}, and Transformer \citep{lepikhin2021gshard, roller2021hash, fedus2022switch, dai-etal-2022-stablemoe, mistral2023mixtral, dai2024deepseekmoe, zhou2025moelpr}. 
Adding more experts scales up the total capacity of the model while keeping similar inference costs on each token. 
Previous studies mainly focus on designing a better load-balancing routing strategy \citep{roller2021hash, fedus2022switch, zhou2022moe} and a training method \citep{sukhbaatar2024btm}. 
Our work is similar to X-MOD \citep{pfeiffer-etal-2022-lifting}, which trains an adapter module for each language in all layers. 
The main differences lie in 
% 1) scaling up capacity-demanding layers in our works rather than all layers in X-MOD. 
1) grouping similar languages in each expert to boost cross-lingual transfer rather than allocating one adapter for each language. 2) Text for inference can be directly input to our model without specifying languages which is inflexible and required for X-MOD. 
% Our work aims to exploit the advantage of conditional computation in mixture of expert and cross-lingual transfer between similar languages to improve the performance of multilingual mixture of expert model.

% A large-scale LSTM-based MoE model \citep{shazeer2017out}
% Sequentially add experts \citep{Aljundi2017expert}
% GShard \citep{lepikhin2021gshard}
% Switch Transformer \citep{fedus2022switch}
% Hash Layer \citep{roller2021hash}
% StableMoE \citep{dai-etal-2022-stablemoe}
% Mixtral 8x7B \citep{mistral2023mixtral}
% DeepSeekMoE \citep{dai2024deepseekmoe}

\begin{figure*}[th]
\centering
\includegraphics[width=0.98\textwidth]{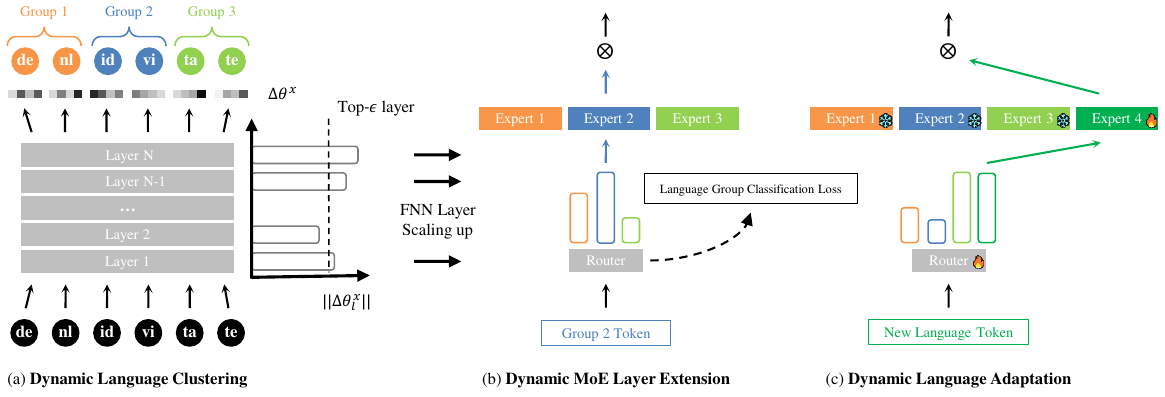}
\vspace{-2mm}
\caption{\label{fig:main}The overview of our method to group and scale up the multilingual LLM. (a) Given pre-training languages, we first determine their parameter deviation $\Delta \theta^{x}$ on the model, then group similar languages by the similarity of $\Delta \theta^{x}$. (b) These layers with higher $\| \Delta \theta^{x}_{l} \|$ are extended to MoE layers, where each expert is tuned with tokens from the corresponding language group to specialize. (c) To adapt to the new language, we copy the multilingual expert from the most similar language group, then only fine-tune the router and expert added.}
\vspace{-5mm}
\end{figure*}

\subsection{Multilingual Large Language Model}
The pre-training methods of multilingual large language models \citep{conneau-etal-2020-unsupervised, lin-etal-2022-shot, workshop2023bloom, yang2023bigtrans, wei2023polylm, ustun-etal-2024-aya, dang2024ayaexpanse, ming2024marcollm, ji2025emma500} are mainly extended from the one for the monolingual corpus \citep{radford2018gpt, devlin-etal-2019-bert, lewis-etal-2020-bart, raffel2020t5}, and relied on a balanced sampling method to mitigate the performance gap between languages. 

To mitigate the curse of multilinguality, \citet{blevins-etal-2024-breaking} applied the Branch-Train-Merge method \citep{li2022btm} on the training of multilingual language model, where one model serves for a group of languages, and assembled output of top-m models after language identification during inference. 
In contrast, our method is motivated by the distribution of parameter deviation during the training of multilingual large language models and strives to scale up the parameter on the language-specific layers. 
It keeps a similar cost without additional language classification while augmenting the capacity of the multilingual language model during inference. 

\section{Method}
As shown in Figure \ref{fig:main}, to train a \textbf{D}ynamic \textbf{M}ixture-\textbf{o}f-\textbf{E}xperts model (\textbf{DMoE}), we first fine-tune the model on the monolingual corpus to obtain the parameter deviation for each language. 
Then the parameter deviation is used to cluster similar languages (Section \ref{sec:lang_cluster}) and determine layers to extend parameters (Section \ref{sec:layer_exten}). 
Besides, new languages for adaption are also dynamically assigned to the most similar language cluster to mitigate catastrophic forgetting (Section \ref{sec:lang_adaptation}). 

\subsection{Dynamic Language Clustering}
\label{sec:lang_cluster}
The quality of the clustering method is primarily influenced by the choice of similarity metric, making the determination of an appropriate metric central to its effectiveness. 
We first obtain the parameter deviation of the model by fine-tuning only ten steps, investigated in Appendix \ref{app:lang_delta}, and take it as a representation of distinctive characteristics for each language. Given the high-dimensional nature of the parameter deviation, we employ cosine similarity as the metric to measure the similarity between languages. To satisfy the clustering process, we define the intra-group language similarity as follows:
\begin{equation}
\label{sim}
% \scriptstyle
\text{Sim}(\theta, G_k) = \min_{x,y \in G_k} \frac{\Delta \theta^{x} \cdot \Delta \theta^{y}}{\|\Delta \theta^{x}\| \|\Delta \theta^{y}\|}
\end{equation}
where $G_k$ is the $k$-th group of languages, $\Delta \theta^{x}$ and $\Delta \theta^{y}$ are the parameter deviation of language $x$ and $y$ respectively on the parameter $\theta$, and $\Delta\theta^{x} = [\Delta\theta^{x}_1, \Delta \theta^{x}_2, \cdots, \Delta\theta^{x}_N]$ is the concatenation of the parameter deviation from all $N$ layers after fine-tuning on language $x$. 

A higher intra-group similarity indicates that the languages within the group are more similar, resulting in less conflict between them. This reduces the potential for gradient conflicts during the continuous pre-training on different languages, making it more appropriate to share parameters with the same expert. Therefore, we can perform clustering by maximizing the similarity within each group, which can be formalized as follows:
\begin{equation}
% \scriptstyle
\max_{G_1,G_2,\dots,G_K} \sum_{k=1}^K \text{Sim}(\theta, G_k)
\end{equation}
However, obtaining the global optimal solution to this problem is NP-Hard. Additionally, the number of languages in each group needs to be balanced to enhance the utilization of experts. To address these challenges, we employ a greedy algorithm. The pseudo-code is provided in Algorithm \ref{alg:cluster}.

\begin{algorithm}
\caption{Balanced Language Clustering}
\label{alg:cluster}
\begin{algorithmic}[1]
\REQUIRE Parameter deviations for different languages $\Delta \Theta = \{\Delta \theta^{1}, \Delta \theta^2, \dots, \Delta \theta^x\}$, Number of groups $K$ \\
\ENSURE Language clustering result $\textit{Groups}$ \\
\STATE Initialize $ \textit{Groups} = \{\} $
\WHILE{$\Delta \Theta$ is not empty}
    \STATE Compute the cosine similarity between languages $(i, j)$ for all $\Delta \theta^i, \Delta \theta^j \in \Delta \Theta$
    \STATE Find the most similar pair of languages $(i^*, j^*)$
    \STATE Merge languages $i^*$ and $j^*$ to form a group: $G = \{i^*, j^*\}$
    \STATE Remove $i^*$ and $j^*$ from $\Delta \Theta$
    \WHILE{$|G| < \frac{|\Delta \Theta|}{K}$}
    \STATE Compute the intra-group similarity (Eq. \ref{sim}) of $G \cup \{m\}$ for all $\Delta \theta^m \in \Delta \Theta$
    \STATE Find the group $G \cup \{m^*\}$ that maximizes the intra-group similarity
    \STATE Add $m^*$ to $G$ 
    \STATE Remove $m^*$ from $\Delta \Theta$
    \ENDWHILE
    \STATE Add group $G$ to $\textit{Groups}$
\ENDWHILE
\STATE \textbf{Return:} $\textit{Groups}$
\end{algorithmic}
\end{algorithm}
% \vspace{-4mm}

\subsection{Dynamic MoE Layer Extension}
\label{sec:layer_exten}
We assume that the layers with large parameter deviations are important and language-specific during fine-tuning, requiring additional capacities to mitigate the conflicts between languages. 
Thus the top-$\epsilon$ of dense layers and extended to the mixture-of-experts layers with $g$ experts, where $\epsilon \in [0, 1]$ and $g \in \mathbb{N}^{+}$ are hyper-parameters. 
Each expert is initialized from the parameters of the original dense layer, while the ones of router are randomly initialized. 
Corpora from the same language group are organized in the same batch and used to fine-tune the parameters of the corresponding expert. 
We also train the router with the following language group classification loss:
\begin{equation}
% \scriptstyle
\mathcal{L}_{\textit{RC}}(\theta) = -\sum_{x} \sum_{i=1}^{M}  \left[ \text{log}\left(\text{P}_{i}(l|x;\theta)\right) \right]
\end{equation}
where $x$ is a token from the language group $l$, and $\text{P}_i(\cdot)$ is the probability estimated by the router at the $i$-th MoE layer. 
Thus the training loss comes to the weighted sum of \textbf{C}ausal \textbf{L}anguage \textbf{M}odeling (\textbf{CLM}) loss and the above language group classification loss: 
\begin{equation}
% \scriptstyle
\label{eq:loss}
\mathcal{L}(\theta) = \mathcal{L}_{\textit{CLM}}(\theta) + \alpha \mathcal{L}_{\textit{RC}}(\theta)
\end{equation}
where $\alpha \in \mathbb{R}^{+}$ is a hyper-parameter.

% Add language classification loss in the first stage.

\subsection{Dynamic Language Adaptation}
\label{sec:lang_adaptation}
Given new languages to adapt, we introduce a method to augment their capacity while reducing the inference to other languages learned. 
Specifically, samples from the new language are first input to experts through a hard routing strategy. 
The multilingual expert with the lowest perplexity is considered the most similar one, which is copied and only fine-tuned for fast adaptation. 
It is noted that the other part of parameters, like the shared dense layers and the other experts, are frozen to avoid catastrophic forgetting during the new language learning \citep{winata-etal-2023-overcoming}. 

\section{Experiments}
\subsection{Experiments Settings}

\paragraph{Large Language Models} We adopt the multilingual Bloom \citep{workshop2023bloom} and English-centric Gemma \citep{gemmateam2024gemma} series models in this work. 

\paragraph{Corpus} Two multilingual corpora, CulturaX \citep{nguyen-etal-2024-culturax} and MADLAD-400 \citep{kudugunta2023madlad}, are used in this work. 
% and Dolma \citep{soldaini-etal-2024-dolma} are chosen as general English corpus. 
We set the language sampling exponent to 0.3 following mT5 \citep{xue-etal-2021-mt5}. 

\paragraph{Evaluation Tasks} There are five multilingual tasks, covering natural language inference \citep{conneau-etal-2018-xnli}, paraphrase detection \citep{yang-etal-2019-paws}, and multilingual reasoning tasks \citep{ponti-etal-2020-xcopa, lin-etal-2022-shot, tikhonov-ryabinin-2021-heads}, selected to evaluate the performance of multilingual LLMs. 
To reduce the variability of prompt and evaluation method, we choose the default prompt from the language model evaluation harness framework \citep{eval-harness}. 

% XNLI \citep{conneau-etal-2018-xnli}, XCOPA \citep{ponti-etal-2020-xcopa}, PAWS-X \citep{yang-etal-2019-paws}, and XWinograd \citep{tikhonov-ryabinin-2021-heads}

% MMMLU, Multilingual ARC, Multilingual HellaSwag\citep{lai-etal-2023-okapi},

\paragraph{Baselines} 
\begin{itemize}
    \item \textbf{+ Pre-train}, where the base model continues to pre-train on the same multilingual corpus. It denotes the performance of the vanilla continual multilingual pre-training method. 
    % \item The mixture-of-expert model with the same parameter amount and multilingual tokens budgets.
    \item \textbf{X-ELM} \citep{blevins-etal-2024-breaking} trains a model for two similar languages, and ensembles outputs from top-m models during inference, where m is set to 2 in this work. 
    \item \textbf{Branch-Train-Mix} \citep{sukhbaatar2024btm} trains models specialized for one domain and merges them to obtain a mixture-of-experts model, which shows significantly better performance than Branch-Train-Merge \citep{li2022btm}. We apply our dynamic language clustering results to it, serving as a strong multilingual mixture-of-experts baseline. 
\end{itemize}

To conduct a fair comparison, the total training token amount is the same for all methods. 
Hyper-parameters are reported in Appendix \ref{app:hyper}. 

\subsection{Results on 18 Languages}
\begin{figure}[th]
\centering
\includegraphics[width=0.47\textwidth]{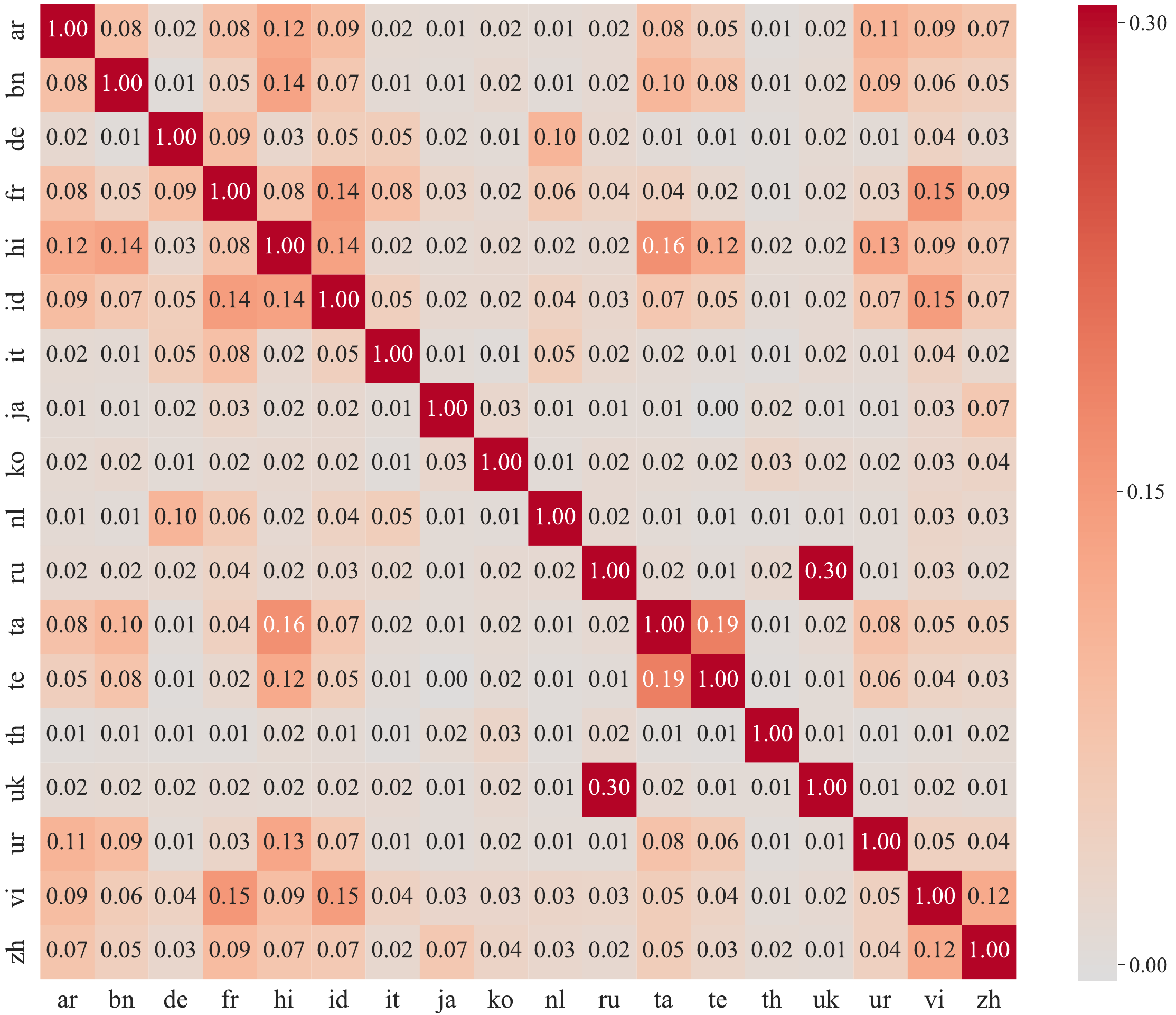}
\vspace{-2mm}
\caption{\label{fig:18langs_560m}The cosine similarity between 18 languages for $\text{BLOOM}_{\text{560M}}$.}
\vspace{-4mm}
\end{figure}

We first conduct experiments on 18 languages from 9 language families. 
Figure \ref{fig:18langs_560m} illustrates the pair-wise language similarity calculated by the $\Delta \theta$ of $\text{BLOOM}_{\text{560M}}$, and more details of other models refer to Appendix \ref{app:lang_similarity}. 
It mostly exhibits some linguistic characteristics. 
For example, Tamil (ta) and Telugu (te), which both come from the Dravidian language family, show a similar trend among languages and have higher similarity than the other languages. 
Based on the pair-wise similarity, languages are divided into multiple groups by Algorithm \ref{alg:cluster}, and results are reported in Table \ref{tab:18langs_group}. 
Appendix \ref{app:lang_similarity} reports the results of other language models investigated. 

\input{tabs/bloom560M_18langs_group}

\input{tabs/main_18langs_ppl}

After language clustering, we continue pre-training on 18 languages under a 6.5B tokens amount budget from CulturaX. 
The perplexity results of models across different parameter amounts are shown in Table \ref{tab:18langs_ppl}. 
We can find that scaling up model parameters brings better language modeling performance compared with the continued pre-training method (``+ Pre-train''). 
DMoE obtains the highest average improvement on perplexity (+11.4\% over ``+ Pre-train'') than the other two strong baseline methods (+0.8\% and +2.2\% respectively) and requires the least additional parameters. 
It is noted that DMoE with 9 experts outperforms X-ELM \citep{blevins-etal-2024-breaking} using 3.6x less parameters. 
The improvement mostly comes from unseen languages like German (+17.9\%) and low-resource languages like Urdu (+13.6\%). 
Appendix \ref{app:qwen} shows similar results of Qwen2.5 base models. 

\begin{figure}[th]
\centering
\includegraphics[width=0.46\textwidth]{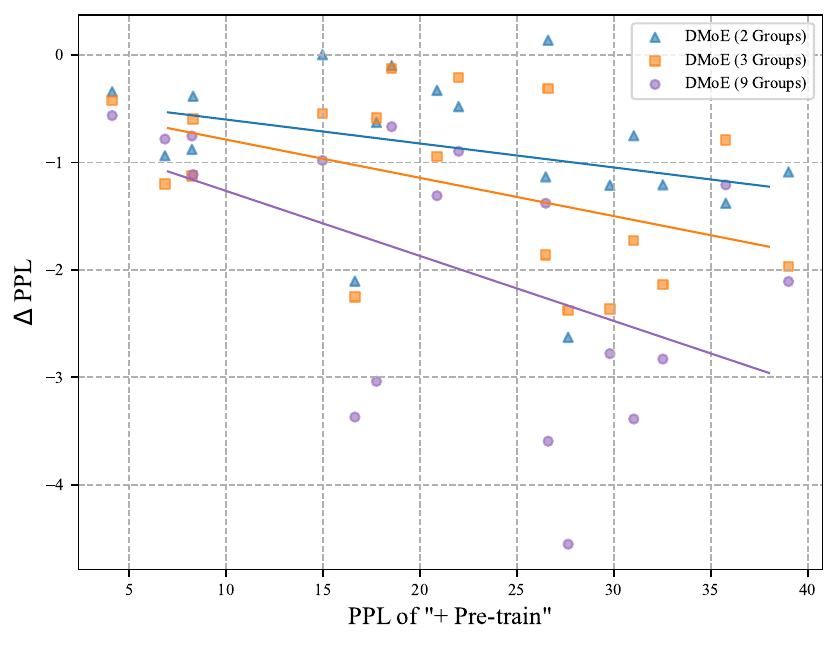}
\caption{\label{fig:ppl_fit}The improvement of DMoE comparing to the continual pre-training baseline method on BLOOM${}_{\text{560M}}$.}
\vspace{-3mm}
\end{figure}

Figure \ref{fig:ppl_fit} illustrates the trend of perplexity improvement over the continual pre-training baseline using DMoE across 18 languages. 
It can be found that languages with higher perplexity benefit more from our method. 
Moreover, with more language groups divided, DMoE shows better multilingual language modeling performance. 

% \paragraph{Number of experts}

% \paragraph{Clustering results}

\paragraph{Transfer language similarities.}
To evaluate the effectiveness of dynamic language clustering (Section \ref{sec:lang_cluster}), we replace the 6-group dividing into the LANG2VEC \citep{littell-etal-2017-uriel} and random grouping results in Table \ref{tab:18langs_group}. 
The ``w/ Random Cluster'' row reports the language modeling result on $\text{BLOOM}_{\text{560M}}$, which is worse than the DMoE model (+1.4 PPL on average). 
We argue that the poor result arises from the negative transfer between dissimilar languages, especially deteriorating the performance of low-resource languages like Urdu (+2.6 PPL). 
And LANG2VEC brings an inferior result, +0.3 PPL on average, comparing our model-specific method. 
It demonstrates that better language clustering results can bring better cross-lingual transfer to the low-resource languages. 
The language clustering result of $\text{Gemma}_{\text{2B}}$ is further applied on $\text{BLOOM}_{\text{560M}}$ to investigate the transferability of our method. 
It is interesting to find that $\text{BLOOM}_{\text{560M}}$ with Gemma clusters is slightly worse than the vanilla model in Table \ref{tab:18langs_ppl}. 
Although our method shows some transferability, we still recommend using language similarity classification based on its parameter derivation for better results. 

\paragraph{Trade-off between learning and forgetting.}
% Evaluate the fast adapting new language
When new languages come for multilingual models to adapt, it is better to achieve fast adaptation while alleviating the catastrophic forgetting of languages learned \citep{sun-etal-2020-distill, zhao2022life, wu-etal-2024-f}. 
We adopt 4 unseen languages: Belarusian (be), Malayalam (ml), Marathi (mr), and Serbian (sr) to evaluate the performance of models. 
As shown in Table \ref{tab:18langs_add4}, the dense model suffers a catastrophic forgetting of the 18 languages learned after \textbf{L}anguage \textbf{A}daptation \textbf{P}re-\textbf{T}raining (\textbf{LAPT}), especially on the medium and low resource languages (+2.0 PPL). 
In contrast, the proposed \textbf{D}ynamic \textbf{L}anguage \textbf{A}daptation (\textbf{DLA}) method (Section \ref{sec:lang_adaptation}) for DMoE achieves better adaptation results on new languages, and mitigates the catastrophic forgetting of the languages learned (only +0.7 PPL). 
It benefits from language-specific expert design and fine-tuning method, which provides a better module for initialization and reduces the inference to the modules learned. 

\begin{figure}[ht]
\centering
\includegraphics[width=0.47\textwidth]{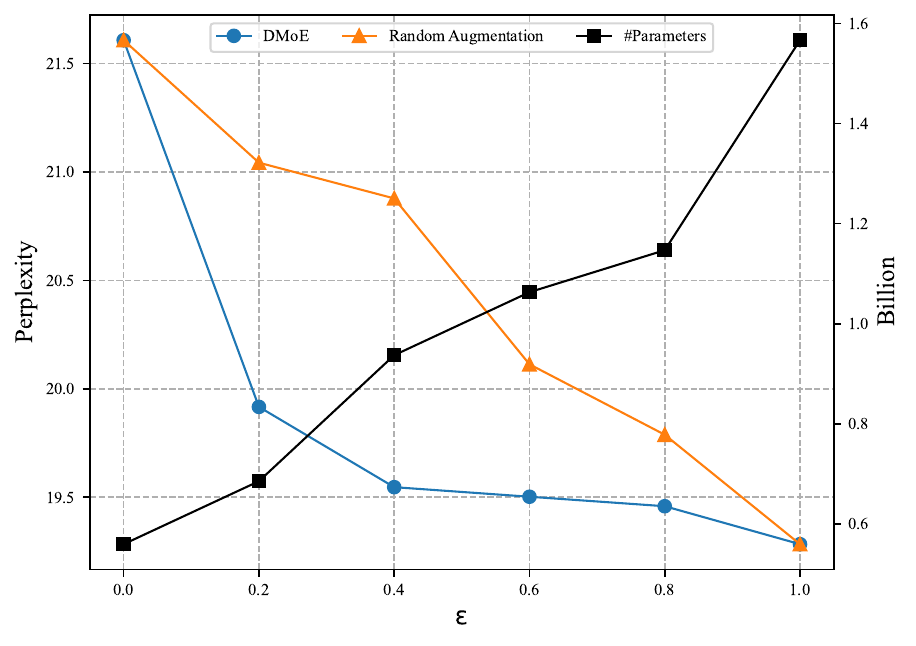}
\caption{\label{fig:18langs_ppl_eps}The average perplexity of DMoE across 18 languages under different $\epsilon$ using $\text{BLOOM}_{\text{560M}}$, where $\epsilon$ = 1 denotes all layers are extended to MoE layers.}
\vspace{-4mm}
\end{figure}

\input{tabs/add4langs}

\begin{figure*}[th]
\centering
\vspace{-3mm}
\subfigure[DMoE (6 Groups) w/o language group classification loss.]{\label{fig:tok_lang_dist}\includegraphics [scale=0.7]{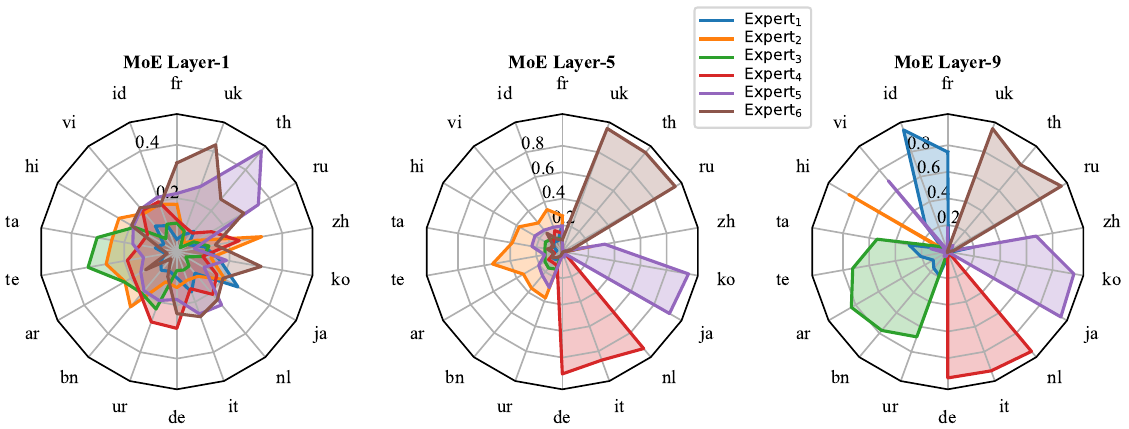}}
\vspace{-3mm}
\subfigure[DMoE (6 Groups) w/ language group classification loss.]{\label{fig:tok_lang_add_loss_dist}\includegraphics [scale=0.7]{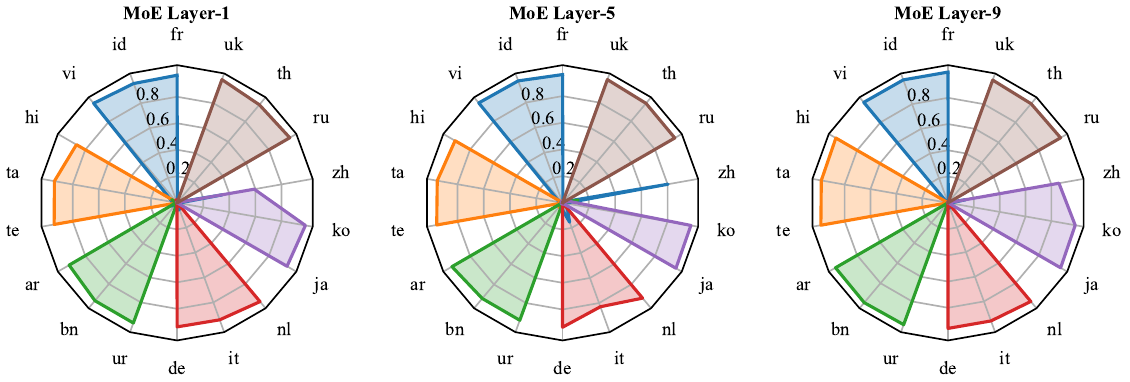}}
\vspace{-3mm}
\caption{\label{fig:tok_moe_dist}The router distribution of top-1 expert for texts in different languages. (a) DMoE trained with randomly initialized router. (b) DMoE trained with language classification loss. Refer to Appendix \ref{app:expert_dist} for more details.}
\vspace{-3mm}
\end{figure*}

\paragraph{Ablation study}
We first modify the hyper-parameter $\epsilon$ to determine the effect of augmenting the layer with higher parameter deviation. 
Figure \ref{fig:18langs_ppl_eps} shows that scaling up layers with higher derivation is much better than the random augmentation baseline when $\epsilon$ is less than 0.5. 
To balance the parameter amount and performance, we set $\epsilon$ to 0.4 in this work. 

The router classification loss is removed to quantify its contribution. 
As shown in the ``w.o/ Class. Loss'' row of Table \ref{tab:18langs_ppl}, the perplexity increases by 0.5 on average. 
Figure \ref{fig:tok_moe_dist} illustrates the statistics of token distribution assigned to the top-1 expert. 
It can be found that the bottom layer like the first layer does not show language specification without router classification loss (Figure \ref{fig:tok_lang_dist}). 
Tokens are mostly assigned to the expert tuned in the same language with router classification loss as expected (Figure \ref{fig:tok_lang_add_loss_dist}). 
% In addition, better language clustering brings higher multilingual performance. 
% DaMoE with random language clusters lags behind 1.4 than the vanilla method. 
% It mostly comes from the worse results in low-resource languages, which shows that a better language clustering result can bring better cross-lingual transfer to the low-resource languages. 

% \input{tabs/main_18langs_in-context}
% xnli, xcopa, paws-x, xwinograde
% translation

\input{tabs/main_128langs_ppl}
\input{tabs/main_128langs_in-context}

\subsection{Extend to 128 Languages}
In this section, we further scale up the number of languages from 18 to 128 and increase the amount of pre-training tokens to 17.7B. 
Following previous findings, the number of language groups is set to 16, and refer to Table \ref{tab:128lang_16G} in Appendix \ref{app:lang_similarity} for more details of language dividing. 

Table \ref{tab:128_ppl} reports the perplexity of 20 languages across different resources and the average result of 128 languages. 
It can be found that DMoE significantly mitigates the curse of multilinguality phenomenon and outperforms Branch-Train-Mix 1.1 PPL on average across 128 languages. 
The improvement mostly comes from unseen languages and low-resource languages, which reach 2.1 PPL on average for the five extremely low-resource languages in Table \ref{tab:128_ppl}. 
The eighty languages with Latin script improved by 2.9 PPL over the continual pre-training model on average, while the other non-Latin languages improved by only 0.7 PPL. 

\begin{figure}[th]
\centering
\includegraphics[width=0.46\textwidth]{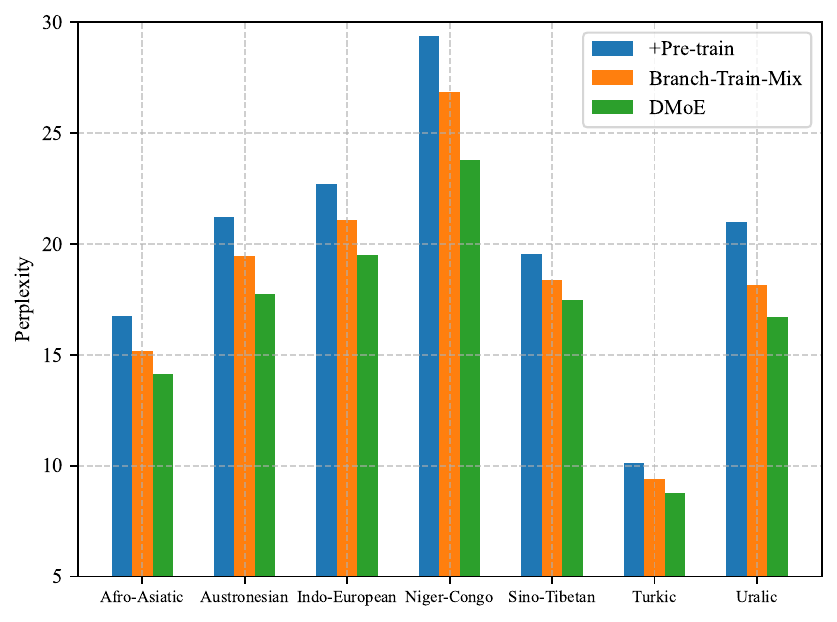}
\vspace{-2mm}
\caption{\label{fig:128ppl_family}The average perplexity of BLOOM${}_{\text{560M}}$ across language families after training on the 128 languages.}
\vspace{-5mm}
\end{figure}

We calculate the improvement across language families and find that the trend is similar where our method outperforms the baseline methods. 
It is interesting to find that languages belonging to the Niger-Congo family, which only take up 0.4GB in the pre-training corpus of BLOOM, benefit the most from our method (Figure \ref{fig:128ppl_family}). 

\paragraph{In-context learning} results on five multilingual datasets are shown in Table \ref{tab:128langs_incontext}. Appendix \ref{app:128langs_all_res} reports the results of all languages on these tasks. 
Similar to the language modeling results, DMoE also boosts the in-context learning performance and outperforms baseline models across two parameter amounts under zero-shot and few-shot settings. 
The performance on the multilingual reasoning task XWinograd benifits most from our method, which improves 1.6\% on average over the base model. 
It further demonstrates the effectiveness of our method in improving multilingual large language models. 

% 128 languages experiments with 16 experts
% baseline: dense, btm(multilingual), DynMMOE

% Add new language

% Analysis

% representation

% Adjust \alpha

% Number of experts 

% Ablation Study
% - Router Initialization
% - Language Distribution -> randomly allocated

% Footnotes are inserted with the \verb|\footnote| command.\footnote{This is a footnote.}

% \section{Ablation Study}
% \label{sec:bibtex}

% Please ensure that Bib\TeX{} records contain DOIs or URLs when possible, and for all the ACL materials that you reference.
% Use the \verb|doi| field for DOIs and the \verb|url| field for URLs.
% If a Bib\TeX{} entry has a URL or DOI field, the paper title in the references section will appear as a hyperlink to the paper, using the hyperref \LaTeX{} package.

\section{Conclusion and Future Work}
In this paper, we propose a method to mitigate the curse of multilinguality by augmenting parameters and boosting cross-lingual transfer. 
Multilingual large language models trained with our method achieve better language modeling and in-context learning performance than the continued pre-trained dense model and other scaling methods. 
These language-specialized experts make it easier to learn new languages and keep multilingual knowledge learned. 

The specialization of experts can be further improved in the future, e.g., a shared expert learning general knowledge and other experts specializing in language-related knowledge. 
Designing a method to determine the language similarity with less cost or calculate better language clustering results is another direction. 
We hope this work can motivate more studies on the curse of multilinguality phenomenon and put forward the development of multilingual language models.

\section*{Limitations}
The first limitation lies in the additional computation to fine-tune and determine the parameter derivation for each language, which will linearly increase with the number of languages involved and the parameter amount of the model. 
Transferring the language similarity calculated from the small model into the larger model is a promising method to save computation. 
% The promising results of the transferability of language similarity show that it can reuse 

% language coverage
The coverage of training and evaluation languages is another limitation. 
For example, languages from the Trans-New Guinea language family are not involved in this work. 
It is mainly due to the constrain of languages provided by the multilingual corpora and datasets used. 

Our method relies on dynamic grouping languages and scaling parameters, which brings a higher training cost than the dense model. 
Due to the limited computation budget, the parameter amount of LLMs investigated in this work is less than 22.6B, and the token amount in training is restrained at 17.7B. 

\section*{Acknowledgements}
We would like to thank Junhong Wu and the anonymous reviewers for their helpful discussions and valuable comments. 
The research work was supported by the Natural Science Foundation of China (No. 62336008) and the Strategic Priority Research Program of Chinese Academy of Sciences (No. XDA04080400).

\bibliography{anthology,custom}
% \bibliography{custom}

\appendix

% \newpage

\section{Hyper-parameters}
\label{app:hyper}
Following \citet{workshop2023bloom}, the global batch size is set to 512 samples with 2048 tokens during the language adaptation pre-training stage. 
AdamW optimizer \citep{loshchilov2019adamw} with $\beta_1 = 0.9$ and $\beta_2 = 0.999$ is used in this work. 
We empirically set the learning rate to 2e-5, adopt bf16 mixed precision training \citep{micikevicius2018mixed} and ZeRO-3 \citep{rasley2020deepspeed} to save GPU memory cost. 
And the $\alpha$ in the loss function (Equation \ref{eq:loss}) is empirically set to 1.28. 
The model learns language-specialized experts and routers in the dynamic MoE layer extension stage, and is trained normally after that. 
All MoE models adopt top-2 routing during inference in this work. 
% A server with 8*NVIDIA A100 80GB RAM is used to conduct all experiments. 

\section{Language Delta}
\label{app:lang_delta}
\begin{figure}[ht]
\centering
\includegraphics[width=0.48\textwidth]{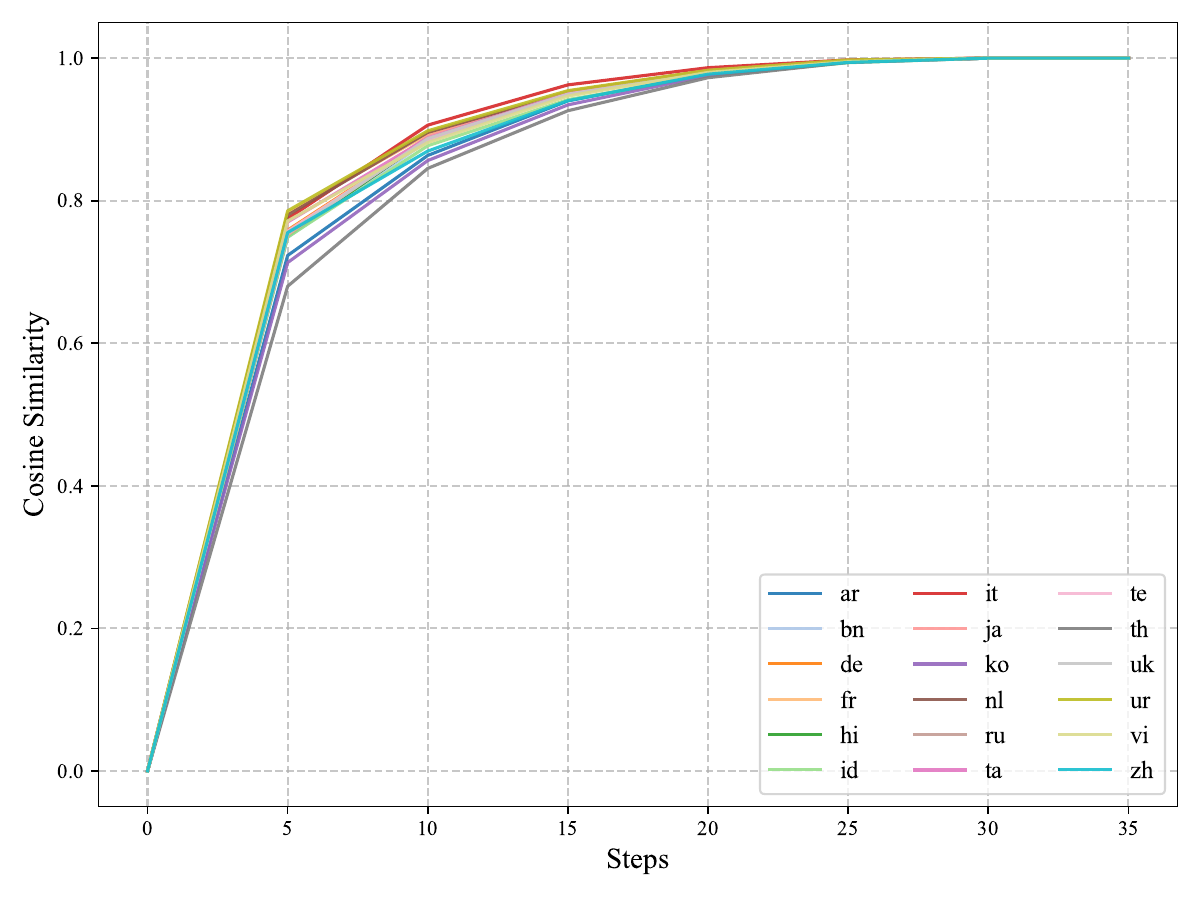}
\caption{\label{fig:18langs_cos_step}The cosine similarity between $\Delta \theta^{x}$ at the i-th step and the one at the 40th step for each language using BLOOM${}_{\text{560M}}$.}
\vspace{-2mm}
\end{figure}

\begin{figure*}[ht]
\centering
\subfigure[5th step]{\label{fig:sim_5step}\includegraphics [width=0.48\textwidth]{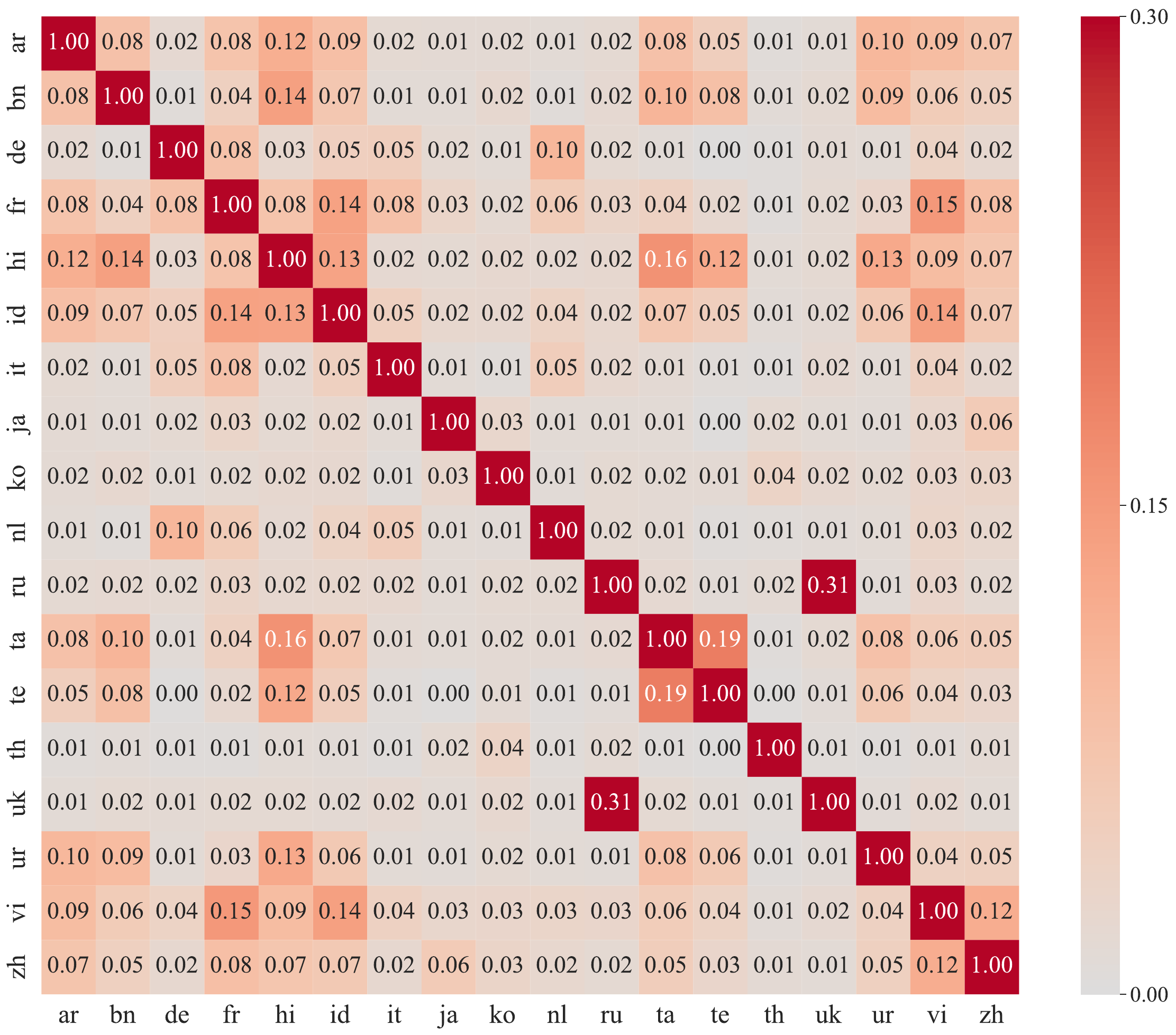}}
\subfigure[10th step]{\label{fig:sim_10step}\includegraphics [width=0.48\textwidth]{imgs/bloom560m_heat_sim_10_last3_18.pdf}}
\subfigure[20th step]{\label{fig:sim_20step}\includegraphics [width=0.48\textwidth]{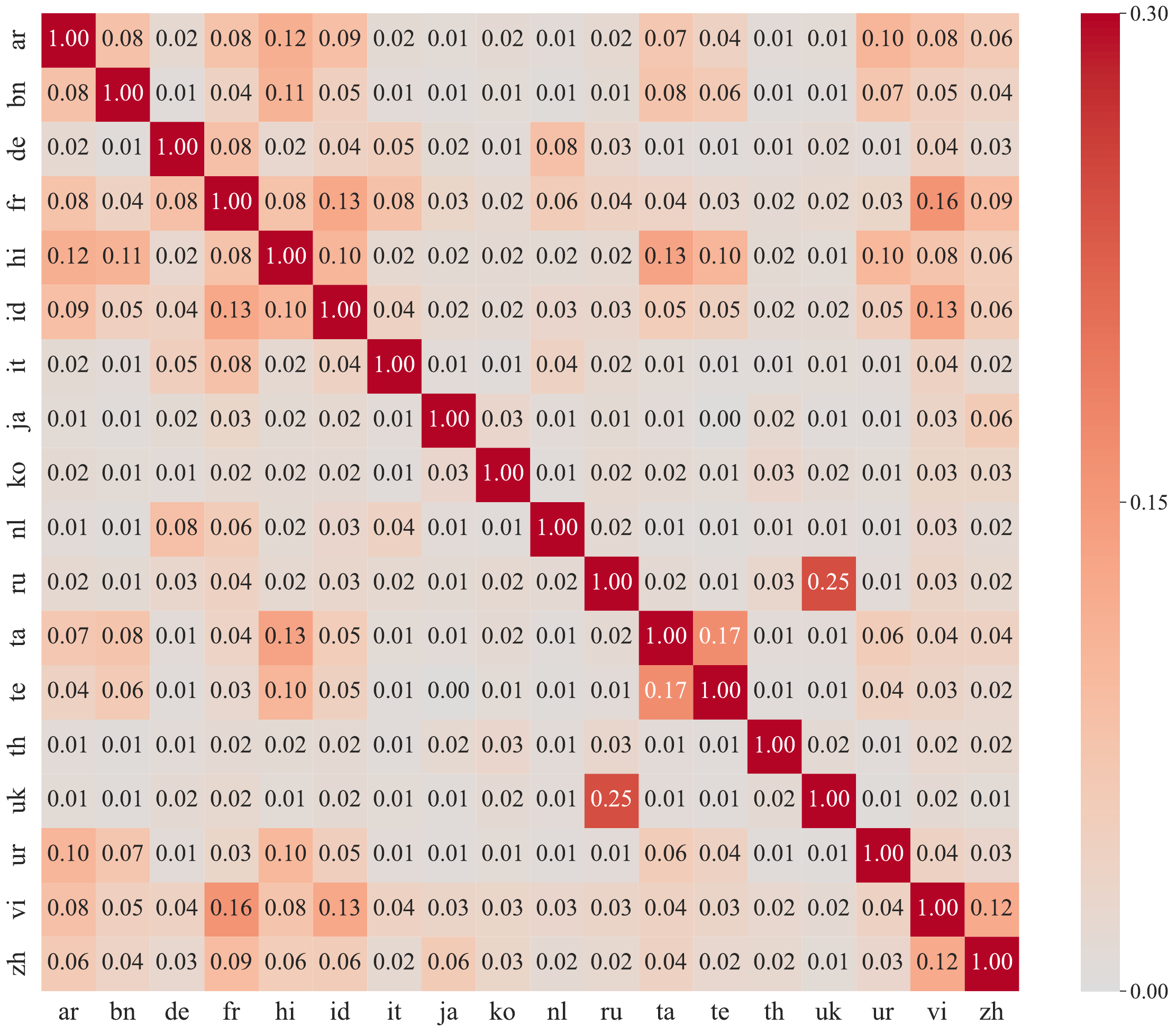}}
\subfigure[40th step]{\label{fig:sim_40step}\includegraphics [width=0.48\textwidth]{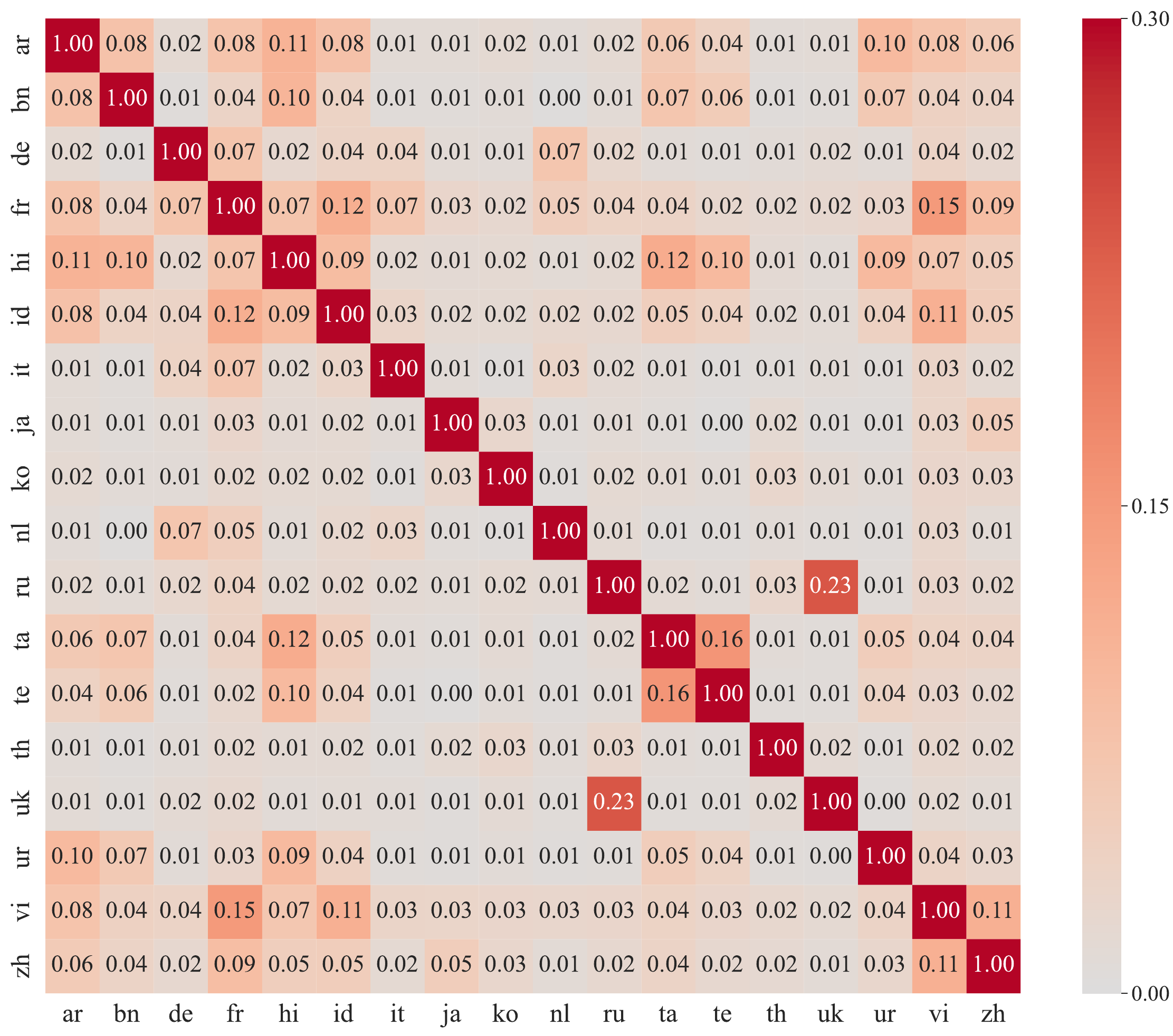}}
\caption{\label{fig:sim_matrices}The language similarity matrices calculated by the parameter derivation at different fine-tuning steps using BLOOM${}_{\text{560M}}$.}
\vspace{-2mm}
\end{figure*}

% TODO: Add experimental results to show that 10 fine-tuning steps are enough. 
Figure \ref{fig:18langs_cos_step} illustrates the cosine similarity of the parameter deviation during fine-tuning. 
It can be found that the deviation is relatively small after 10 tuning steps, and the cosine similarity of the one at the 10th step between the parameter deviation at the 40th step is higher than 80\% for all languages. 
The language similarity matrices are similar using the parameter deviation at different steps (Figure \ref{fig:sim_matrices}). 
Therefore, we only fine-tune 10 steps to determine the parameter derivation of models for each language. 

\begin{figure*}[th]
\centering
\subfigure[ar]{\label{fig:delta_ar}\includegraphics [scale=0.25]{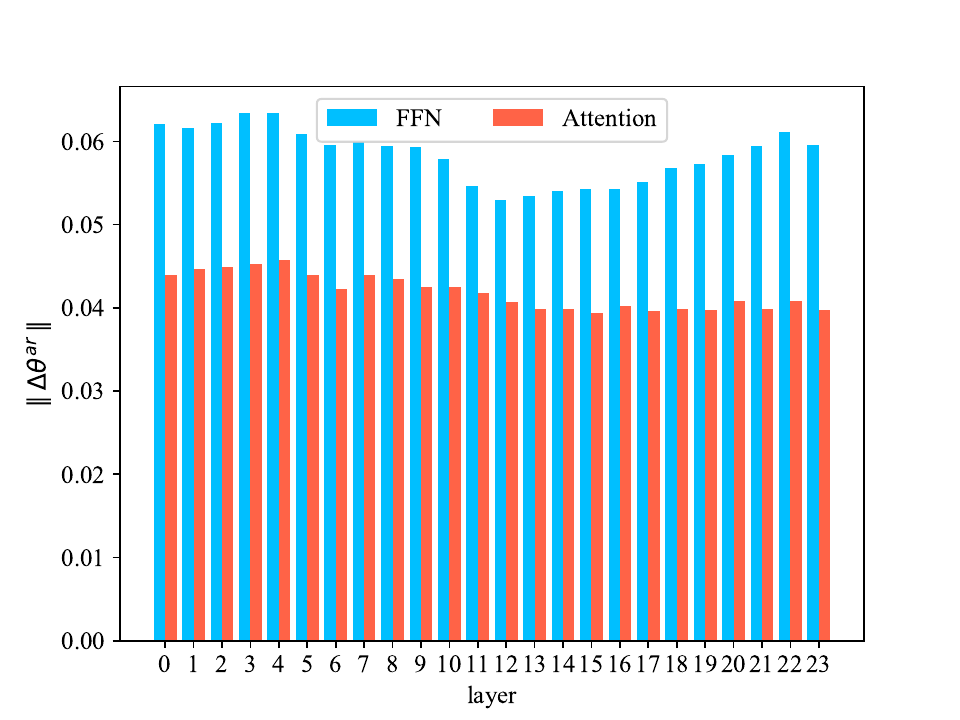}}
\subfigure[bn]{\label{fig:delta_bn}\includegraphics [scale=0.25]{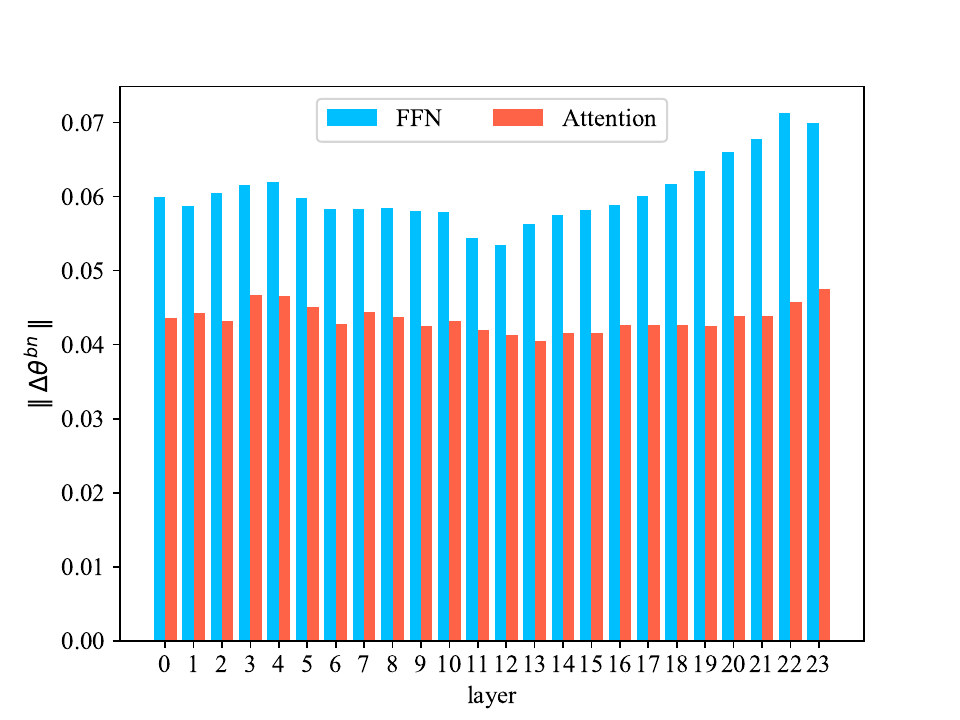}}
\subfigure[de]{\label{fig:delta_de}\includegraphics [scale=0.25]{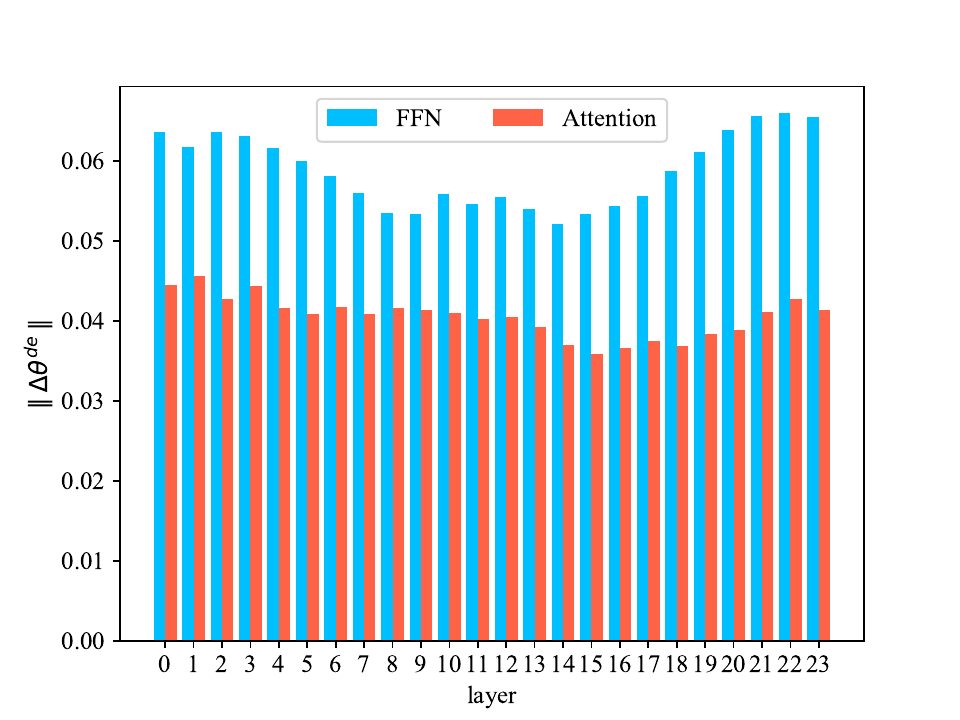}}
\subfigure[fr]{\label{fig:delta_fr}\includegraphics [scale=0.25]{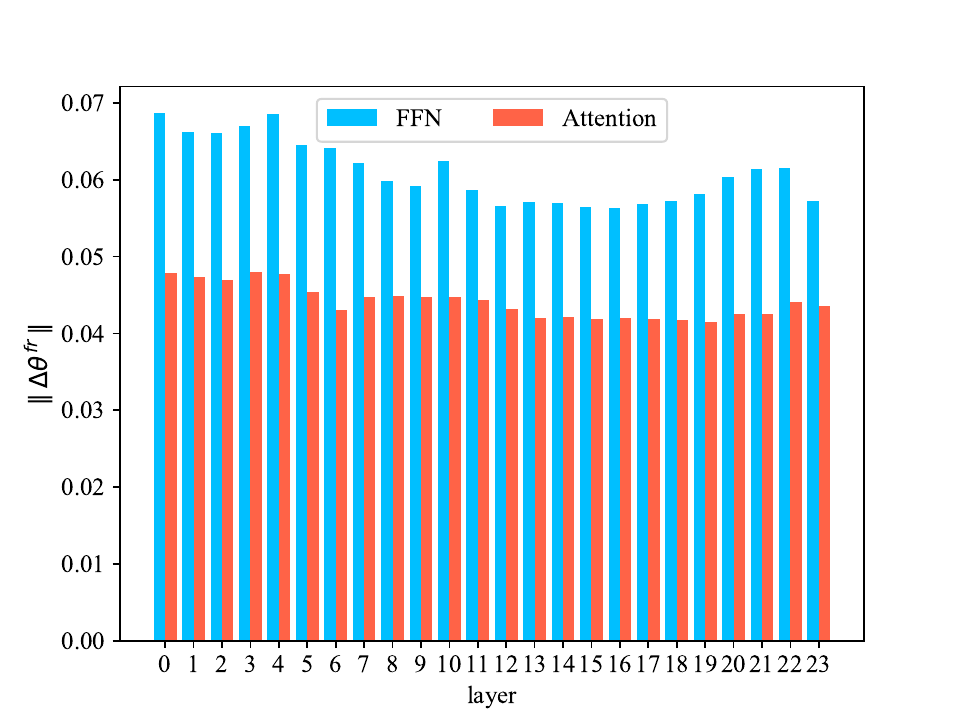}}
\subfigure[hi]{\label{fig:delta_hi}\includegraphics [scale=0.25]{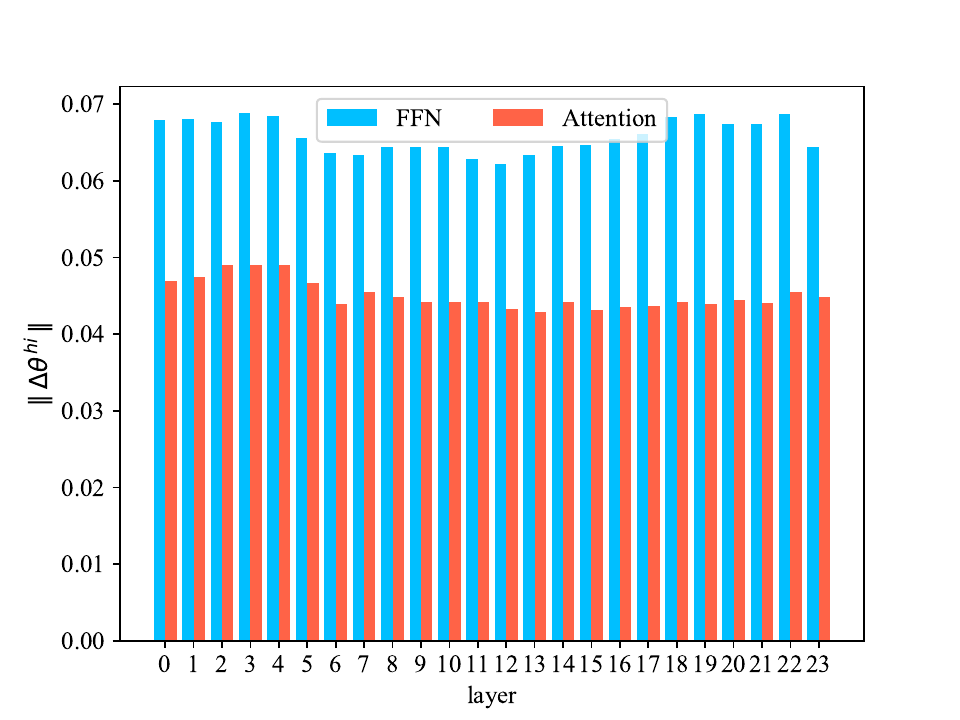}}
\subfigure[id]{\label{fig:delta_id}\includegraphics [scale=0.25]{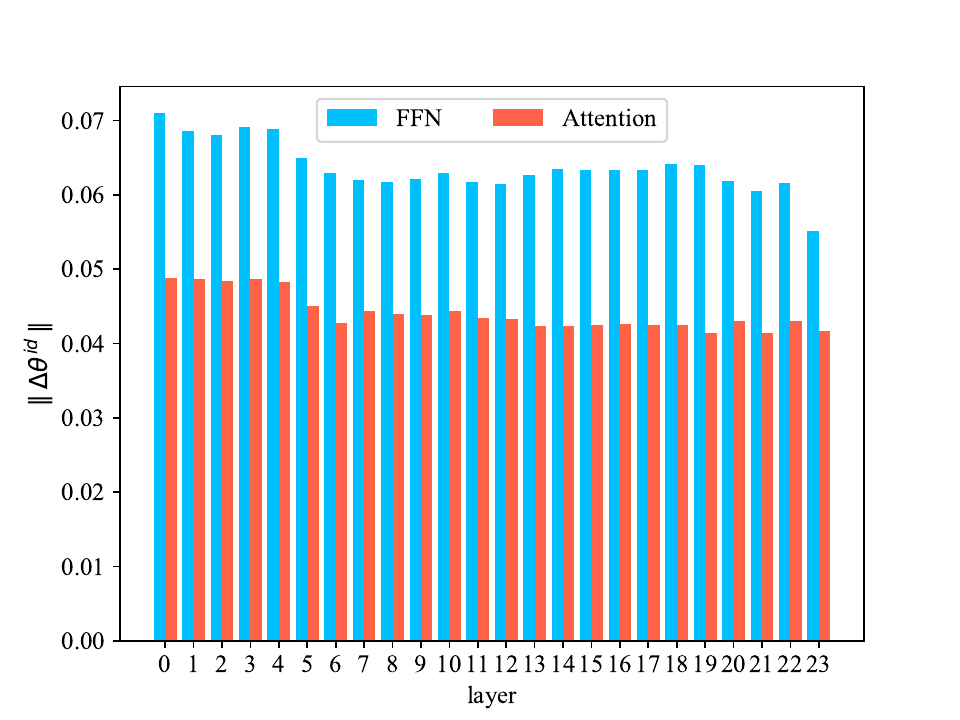}}
\subfigure[it]{\label{fig:delta_it}\includegraphics [scale=0.25]{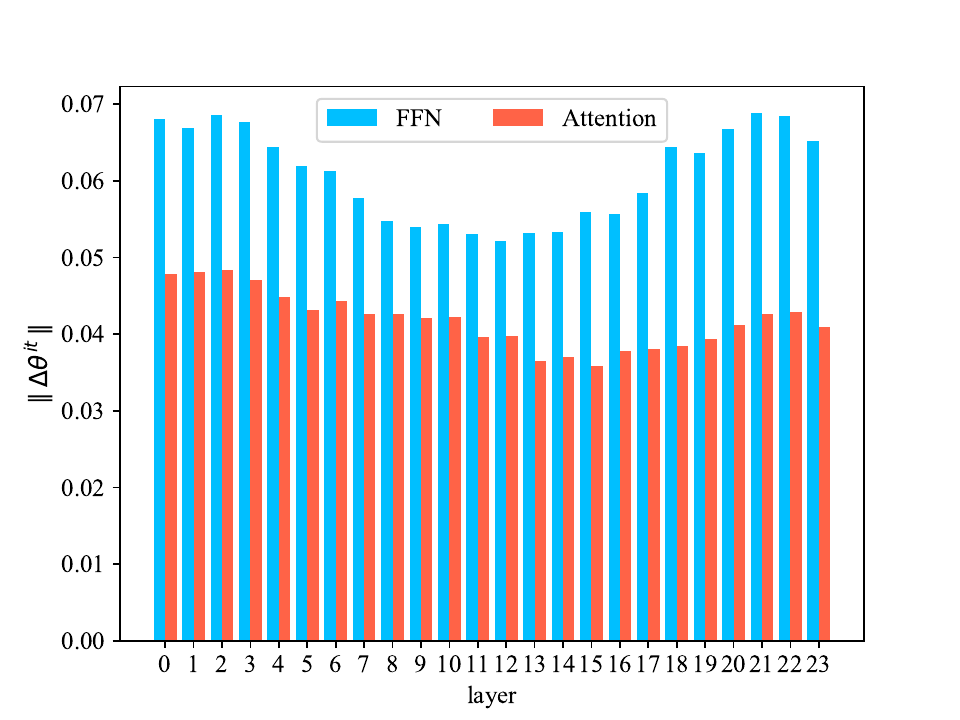}}
\subfigure[ja]{\label{fig:delta_ja}\includegraphics [scale=0.25]{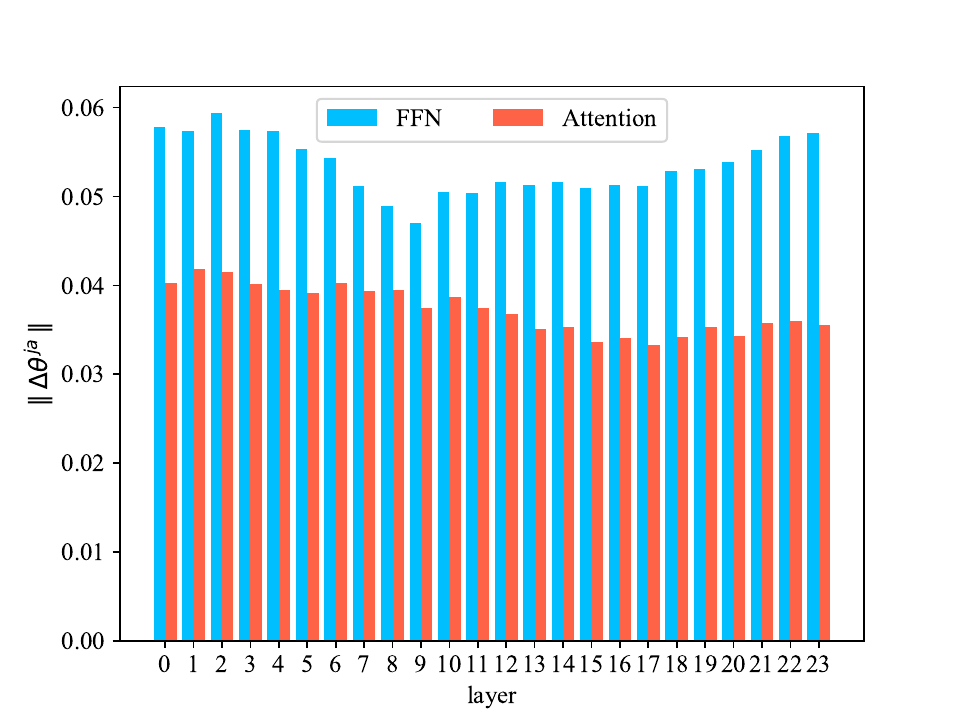}}
\subfigure[ko]{\label{fig:delta_ko}\includegraphics [scale=0.25]{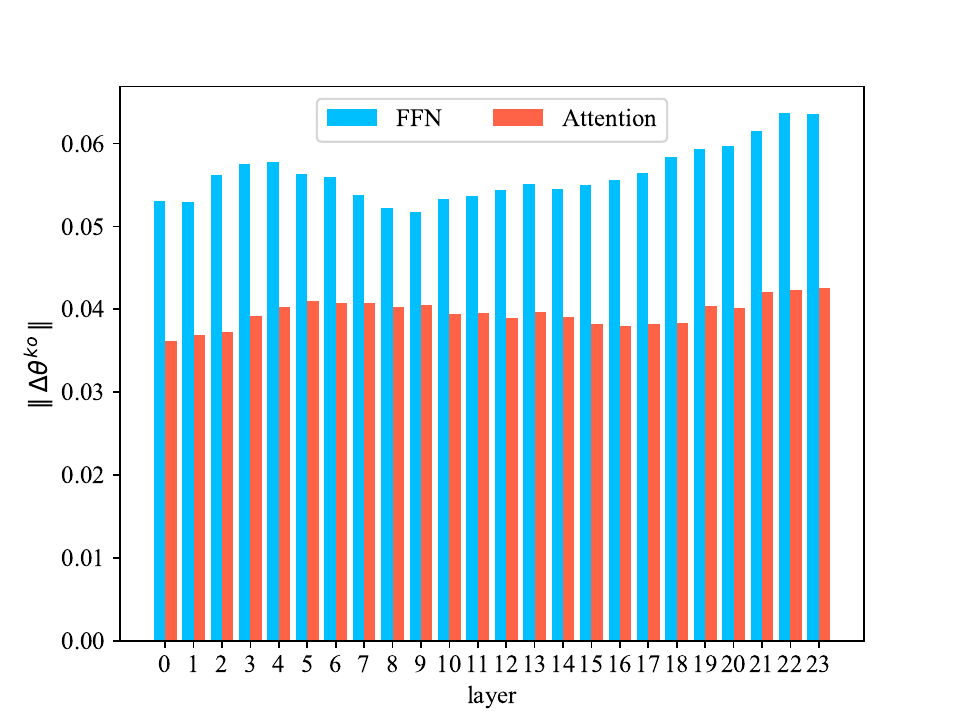}}
\subfigure[nl]{\label{fig:delta_nl}\includegraphics [scale=0.25]{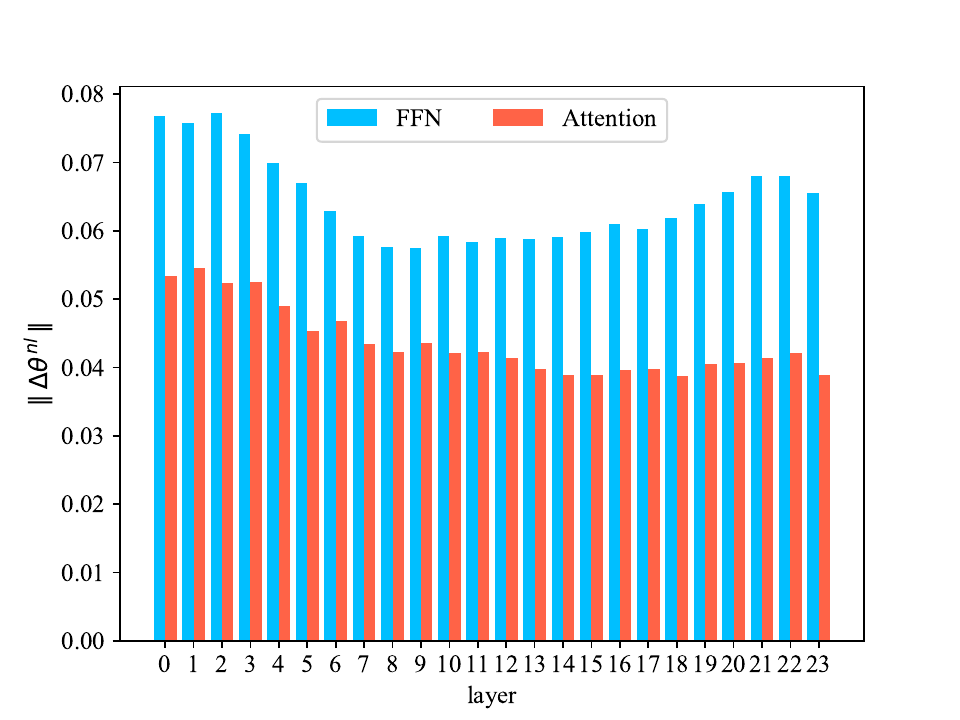}}
\subfigure[ru]{\label{fig:delta_ru}\includegraphics [scale=0.25]{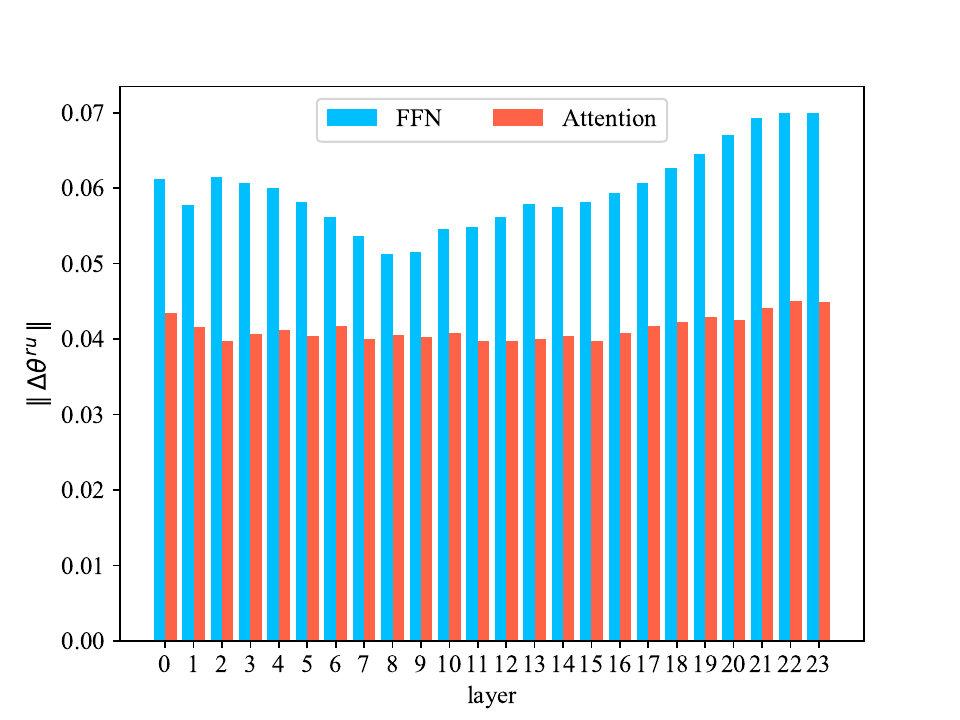}}
\subfigure[ta]{\label{fig:delta_ta}\includegraphics [scale=0.25]{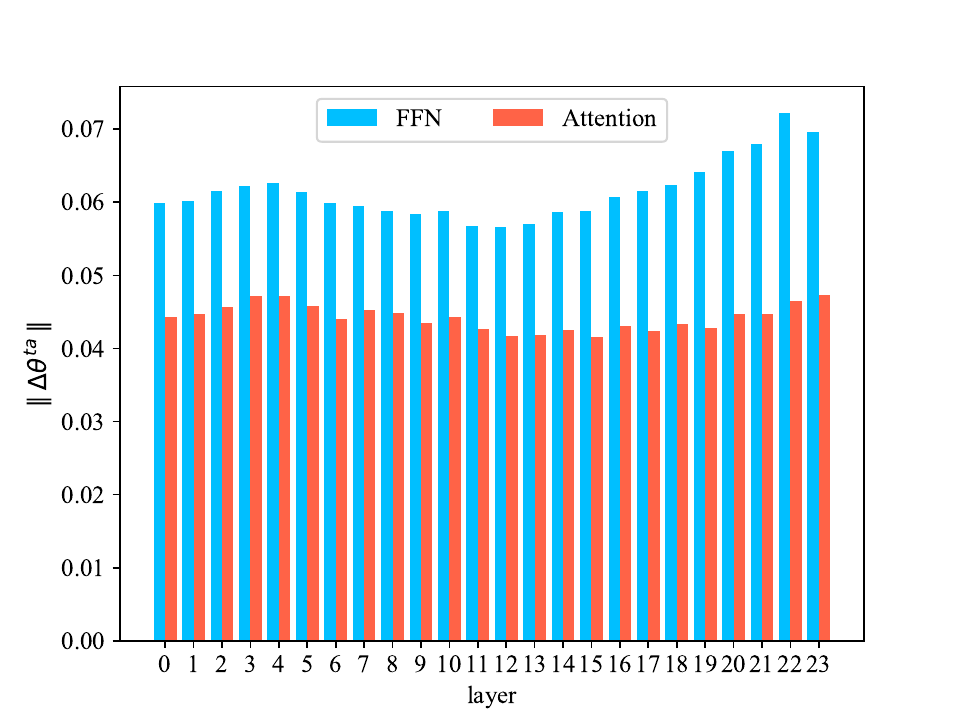}}
\subfigure[te]{\label{fig:delta_te}\includegraphics [scale=0.25]{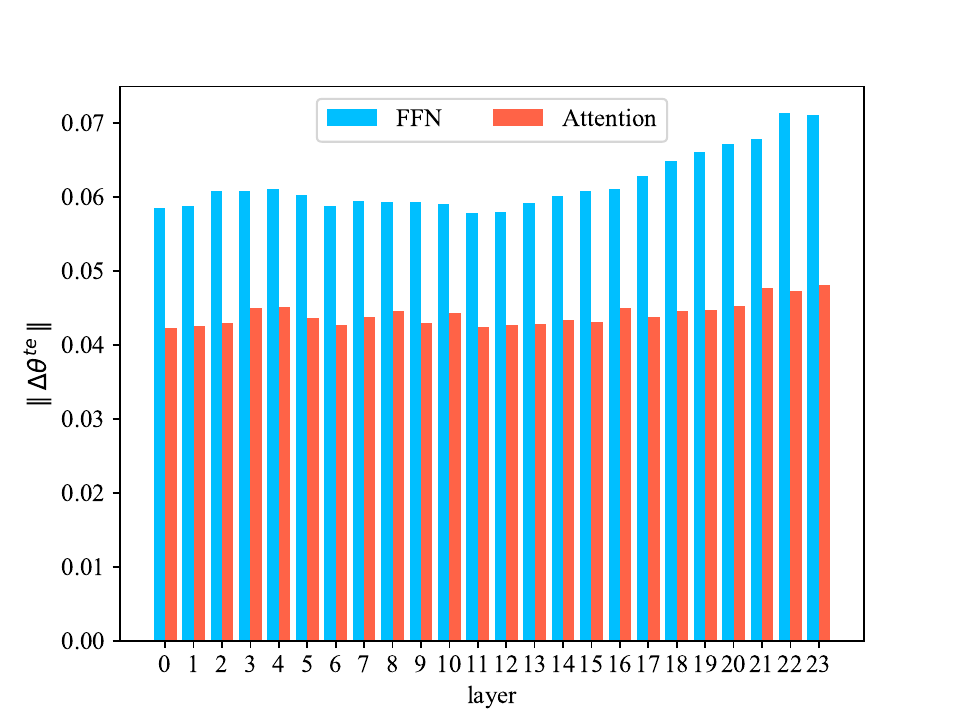}}
\subfigure[th]{\label{fig:delta_th}\includegraphics [scale=0.25]{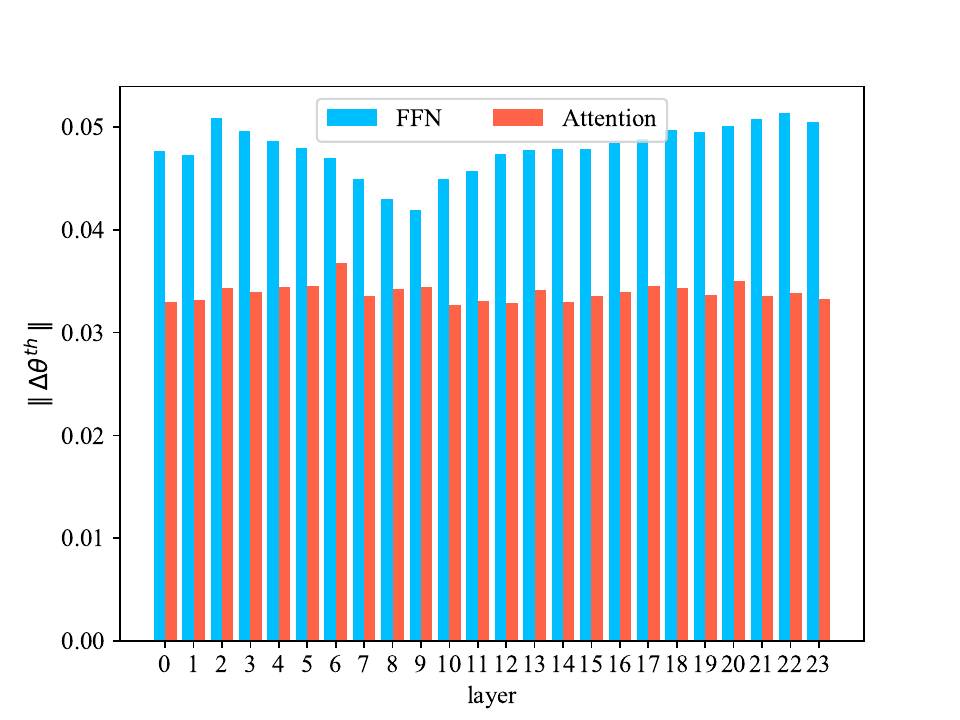}}
\subfigure[uk]{\label{fig:delta_uk}\includegraphics [scale=0.25]{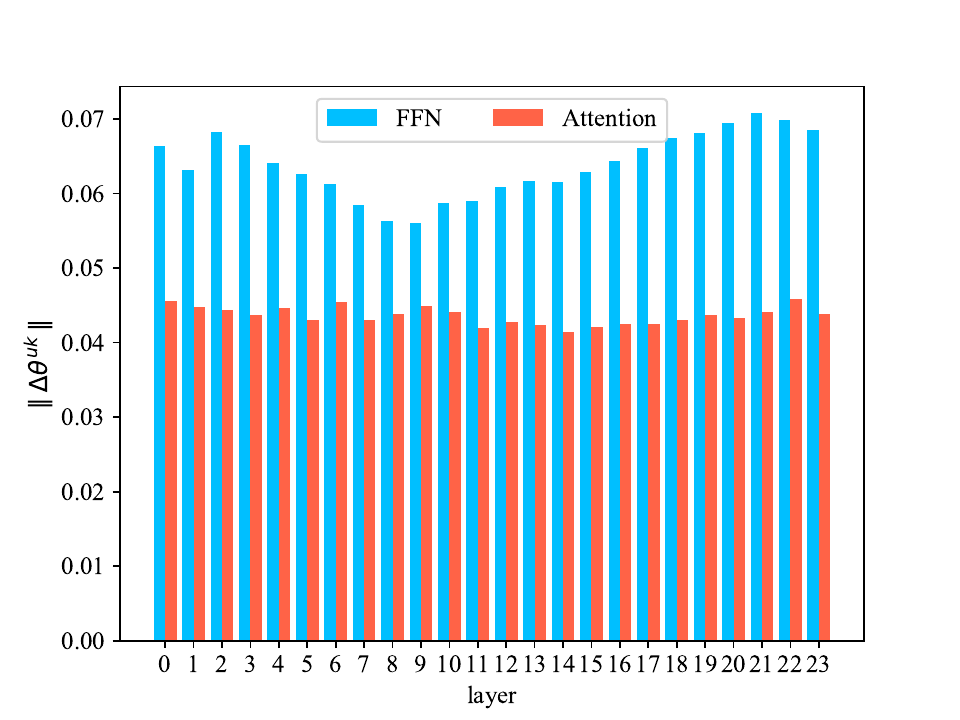}}
\subfigure[ur]{\label{fig:delta_ur}\includegraphics [scale=0.25]{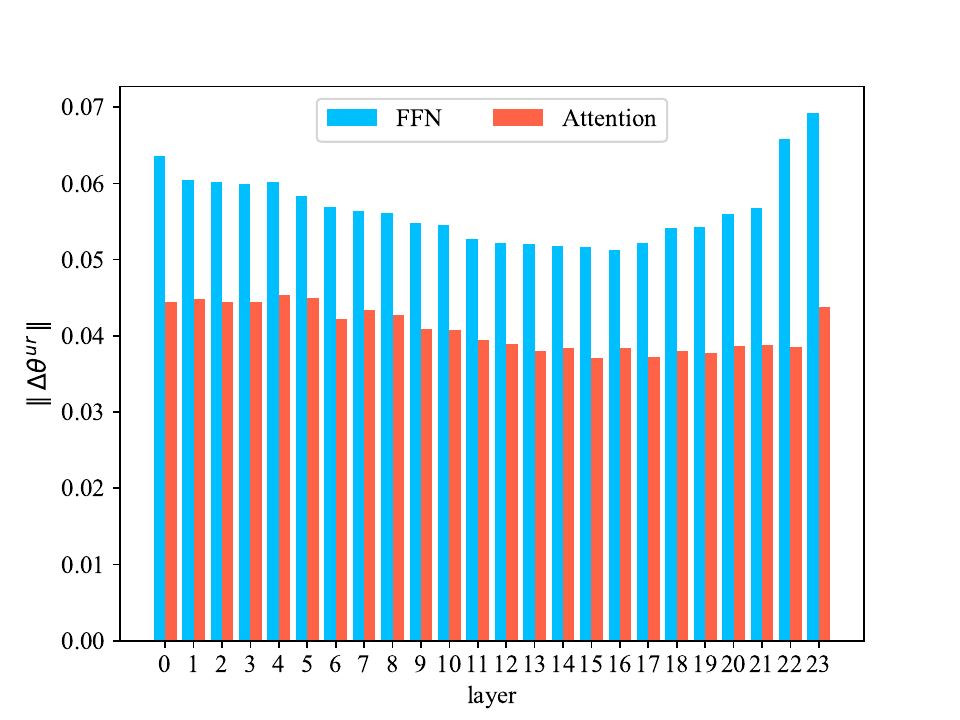}}
\subfigure[vi]{\label{fig:delta_vi}\includegraphics [scale=0.25]{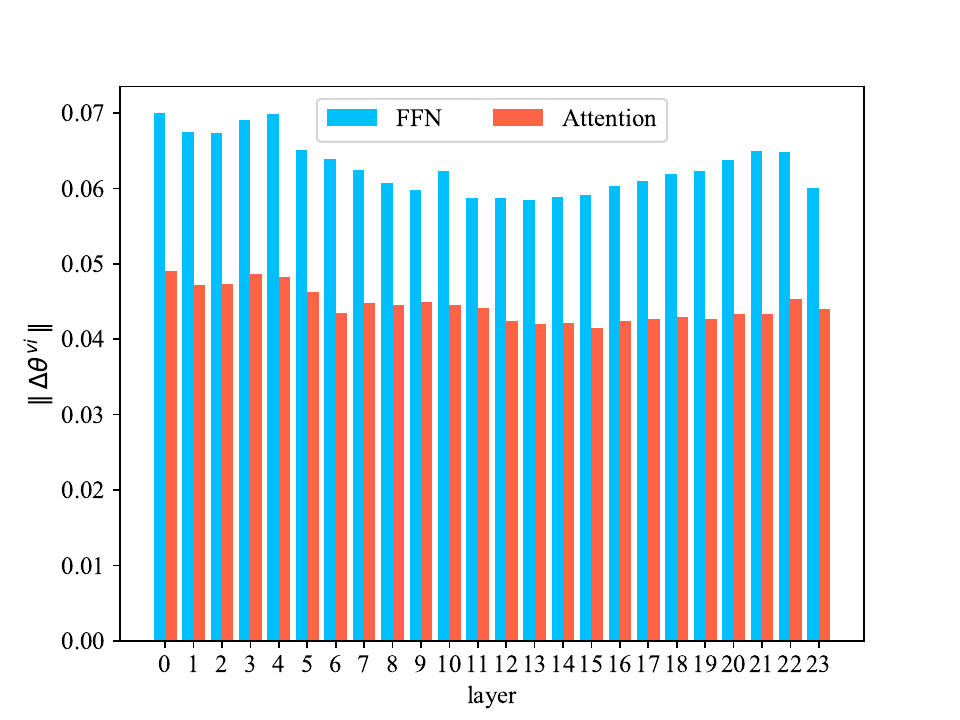}}
\subfigure[zh]{\label{fig:delta_zh}\includegraphics [scale=0.25]{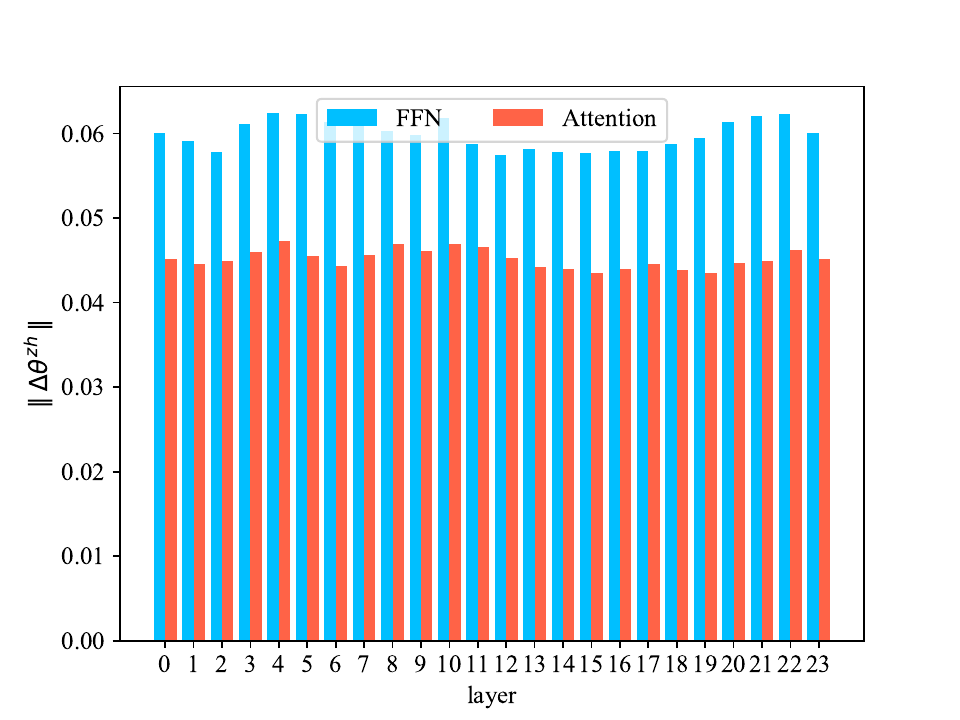}}
\vspace{-2mm}
\caption{\label{fig:delta_bloom}The distribution of parameter deviation $\|\Delta \theta^{x}\|$ across layers of BLOOM${}_{\text{560M}}$ for 18 languages.}
\vspace{-2mm}
\end{figure*}

Figure \ref{fig:delta_bloom} shows the distribution of parameter deviation $\|\Delta \theta^{x}\|$ across layers of BLOOM${}_{\text{560M}}$ for 18 languages. 
It is interesting to find that layers near the embedding or output layer often have a relatively high parameter derivation $\|\Delta \theta^{x}\|$. 

\section{Additional Results}

\subsection{Language Similarity and Grouping}
\label{app:lang_similarity}
\begin{figure}[ht]
\centering
\includegraphics[width=0.48\textwidth]{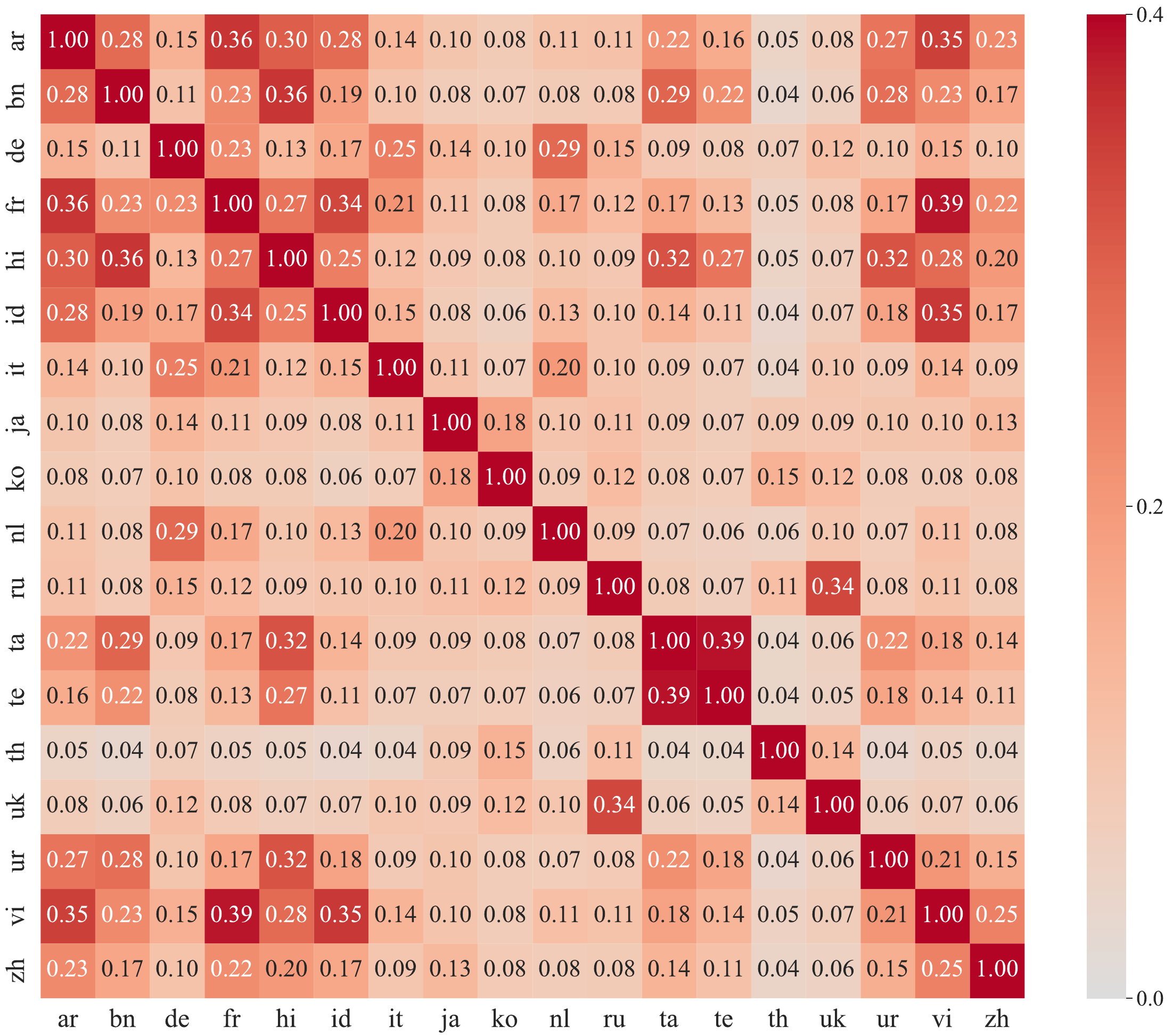}
\caption{\label{fig:18langs_1.7b_all}The cosine similarity between 18 languages using all parameter deviation of $\text{BLOOM}_{\text{1.7B}}$.}
\vspace{-2mm}
\end{figure}

\begin{figure}[ht]
\centering
\includegraphics[width=0.48\textwidth]{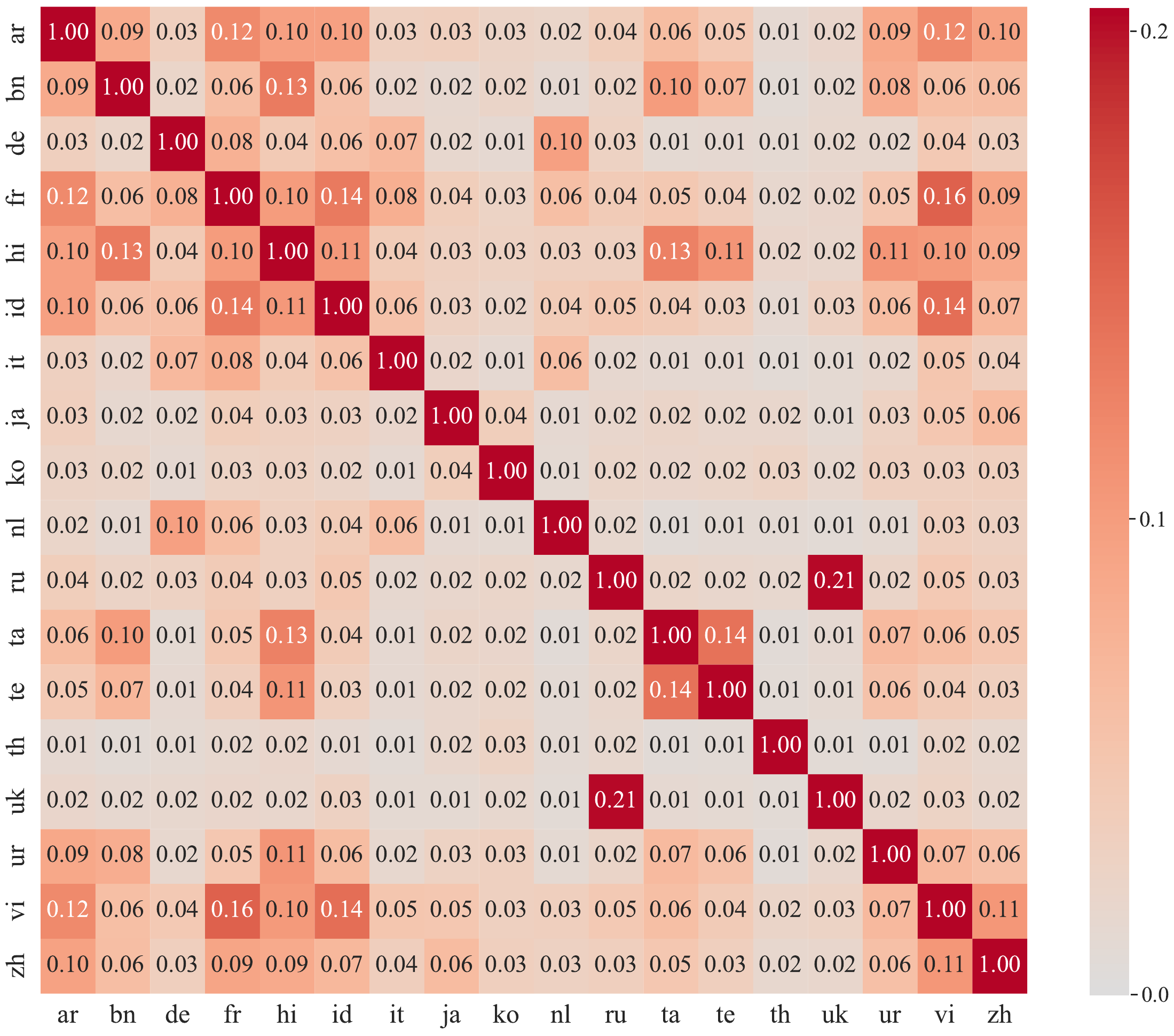}
\caption{\label{fig:18langs_1.7b}The cosine similarity between 18 languages using the parameter deviation of $\text{BLOOM}_{\text{1.7B}}$ at the last 3 layers.}
\vspace{-2mm}
\end{figure}

Figure \ref{fig:18langs_1.7b_all} and \ref{fig:18langs_1.7b} illustrates the language similarity matrix calculated by all layers and the last 3 layers of the parameter derivation of BLOOM${}_{\text{1.7B}}$. 
Except for the difference in absolute value, they are very similar in relative trend. 
Therefore, we adopt the last 3 layers of the parameter derivation to calculate the language similarity by default. 

\begin{figure}[ht]
\centering
\includegraphics[width=0.48\textwidth]{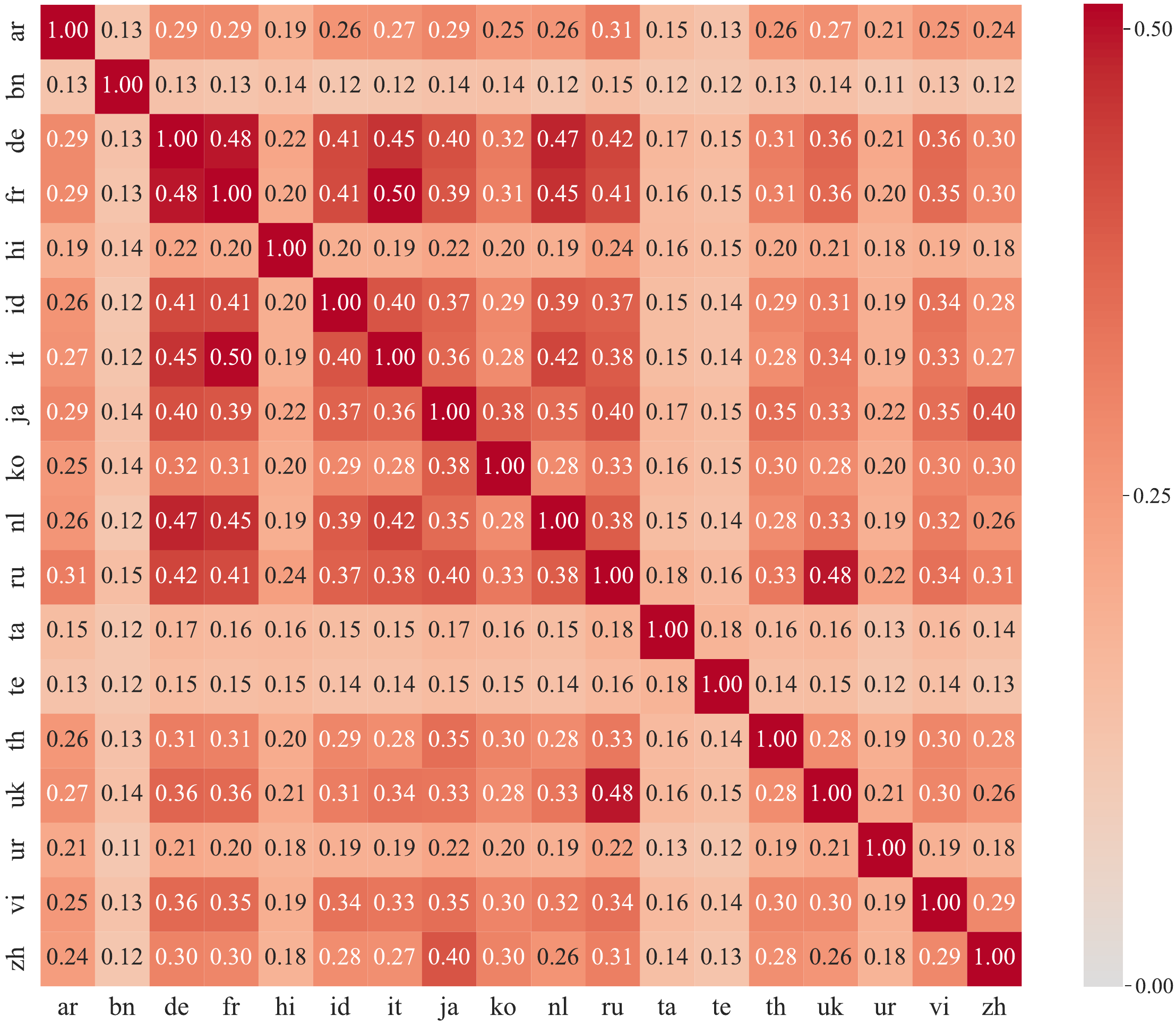}
\caption{\label{fig:18langs_2b}The cosine similarity between 18 languages using the parameter deviation of $\text{Gemma}_{\text{2B}}$ at the last 3 layers.}
\vspace{-2mm}
\end{figure}

\begin{figure}[ht]
\centering
\includegraphics[width=0.48\textwidth]{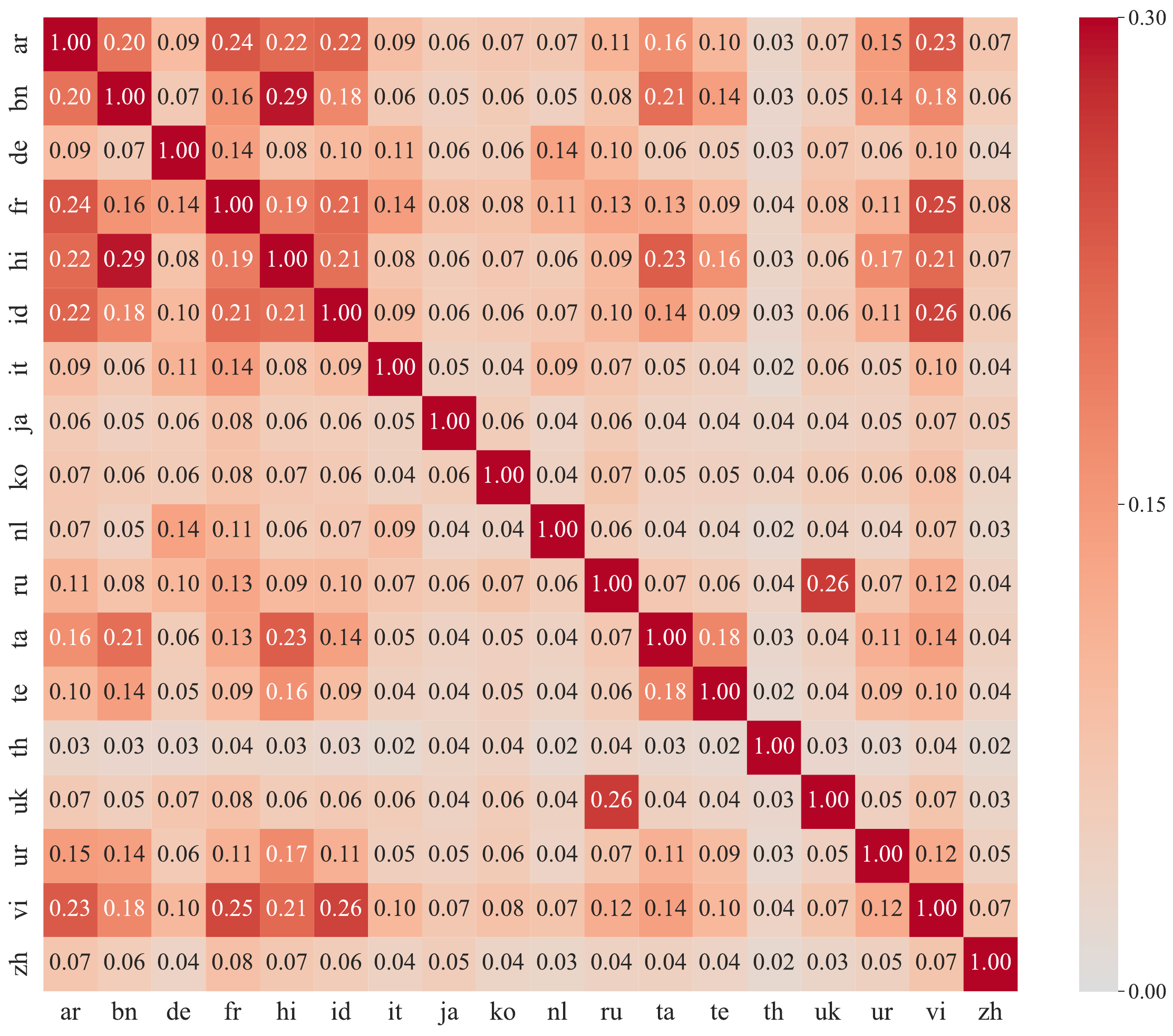}
\caption{\label{fig:18langs_1.7b_madlad400}The cosine similarity between 18 languages using the parameter deviation of $\text{BLOOM}_{\text{1.7B}}$ on the MADLAD-400 multilingual corpus.}
\vspace{-2mm}
\end{figure}

Figure \ref{fig:18langs_2b} shows pair-wise language similarity matrices of Gemma${}_{\text{2B}}$. 
It can be found that the one of BLOOM${}_{\text{1.7B}}$ is similar to the one of BLOOM${}_{\text{560M}}$ (Figure \ref{fig:18langs_560m}), which may arise from the same pre-training corpus used. 
In contrast, the language similarity matrix of Gemma${}_{\text{2B}}$ has a higher average similarity value and different patterns between languages. 
As shown in Figure \ref{fig:18langs_1.7b_madlad400}, It is interesting to find that replacing the multilingual corpus from CulturaX to MADLAD-400 results in a similar matrix. 

Given the language similarity matrix calculated, we obtain the language grouping results for BLOOM${}_{\text{1.7B}}$ and Gemma${}_{\text{2B}}$ in the 18-language experiments using Algorithm \ref{alg:cluster} (Table \ref{tab:18langs_group_bloom1b7} and \ref{tab:18langs_group_gemma2b}). 
Similar languages like Tamil and Telugu are often grouped in the same language cluster. 
BLOOM${}_{\text{560M}}$ and BLOOM${}_{\text{1.7B}}$ have the same language clustering result under the six and two groups settings. 

\input{tabs/bloom1.7B_18langs_group}

\input{tabs/gemma2B_18langs_group}

Table \ref{tab:128lang_16G} reports the 16 language groups used in the 128 languages experiment, which is calculated by the parameter deviation of BLOOM${}_{\text{1.7B}}$. 
BLOOM${}_{\text{560M}}$ adopts this result to save computation for the similar trend with BLOOM${}_{\text{1.7B}}$ in the 18-language experiment. 

% \begin{figure*}[ht]
% \centering
% \subfigure[w/ all layers]{\label{fig:18langs_1.7b_all}\includegraphics [width=0.48\textwidth]{imgs/bloom1.7b_heat_sim_10_all_18.pdf}}
% \subfigure[w/ the last 3 layers]{\label{fig:18langs_1.7b}\includegraphics [width=0.48\textwidth]{imgs/bloom1.7b_heat_sim_10_last3_18.pdf}}
% \caption{The cosine similarity between 18 languages using the parameter deviation of $\text{BLOOM}_{\text{1.7B}}$ at all layers (a) or the last 3 layers (b).}
% \vspace{-2mm}
% \end{figure*}

\input{tabs/langs_128langs_16G}

% \begin{figure*}[ht]
% \centering
% \includegraphics[width=0.98\textwidth]{imgs/bloom1.7b_heat_sim_10_last3_128.png}
% \caption{\label{fig:128langs_1.7b}The cosine similarity between 128 languages for $\text{BLOOM}_{\text{1.7B}}$.}
% \vspace{-2mm}
% \end{figure*}

\subsection{More base models on 18 Languages}
\label{app:qwen}
We apply our method to Qwen2.5 base models and report the results in Table \ref{tab:18langs_ppl_qwen}, which are in line with the previous results in Table \ref{tab:18langs_ppl}.

\input{tabs/main_18langs_ppl_qwen}

\subsection{Token Router Distribution}
\label{app:expert_dist}
\begin{figure*}[th]
\centering
\vspace{-2mm}
\includegraphics[width=0.95\textwidth]{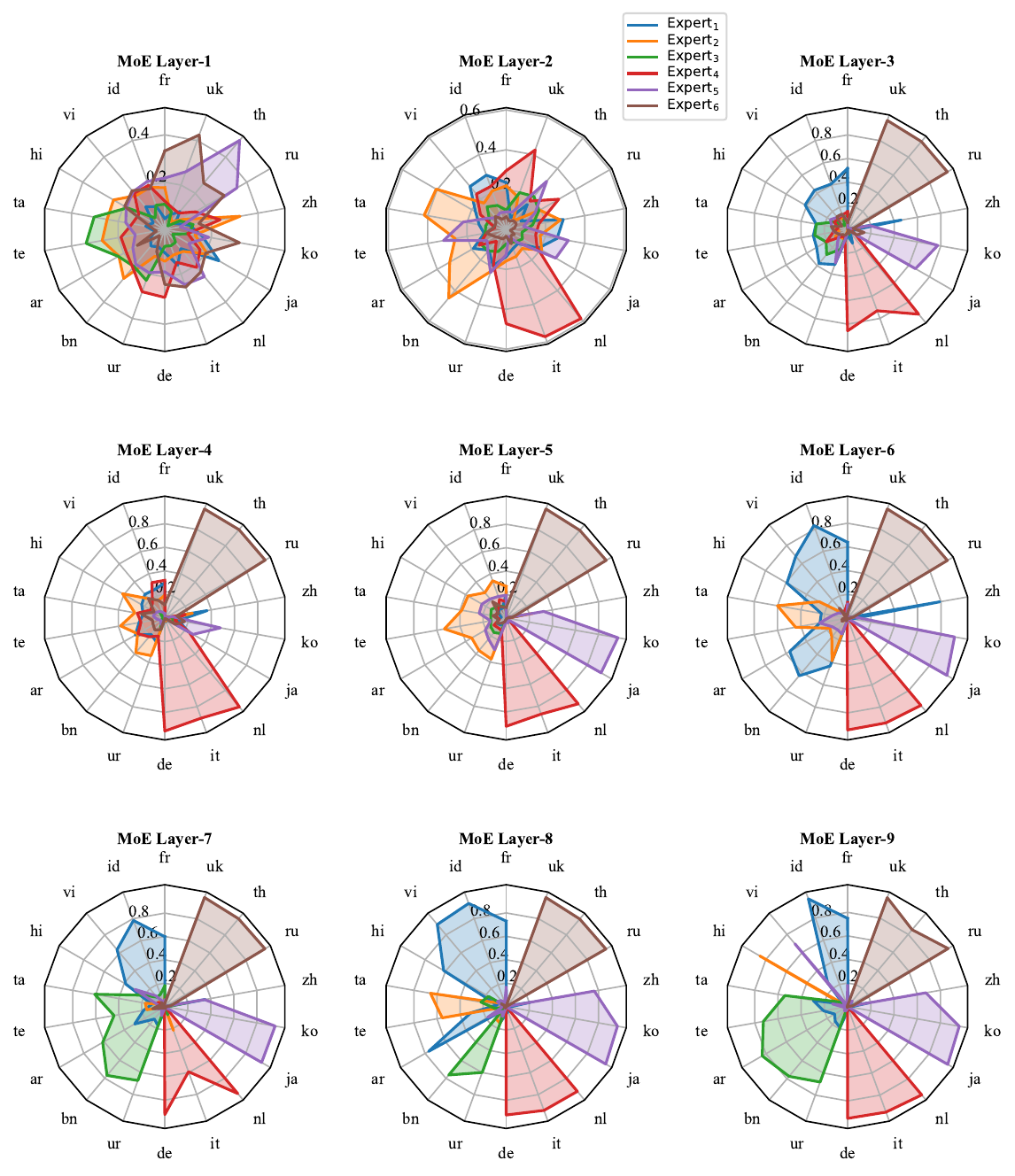}
\caption{\label{fig:tok_lang_dist_all}The router distribution of top-1 expert for texts in different languages on models trained with randomly initialized router.}
\vspace{-2mm}
\end{figure*}

\begin{figure*}[th]
\centering
\vspace{-2mm}
\includegraphics[width=0.95\textwidth]{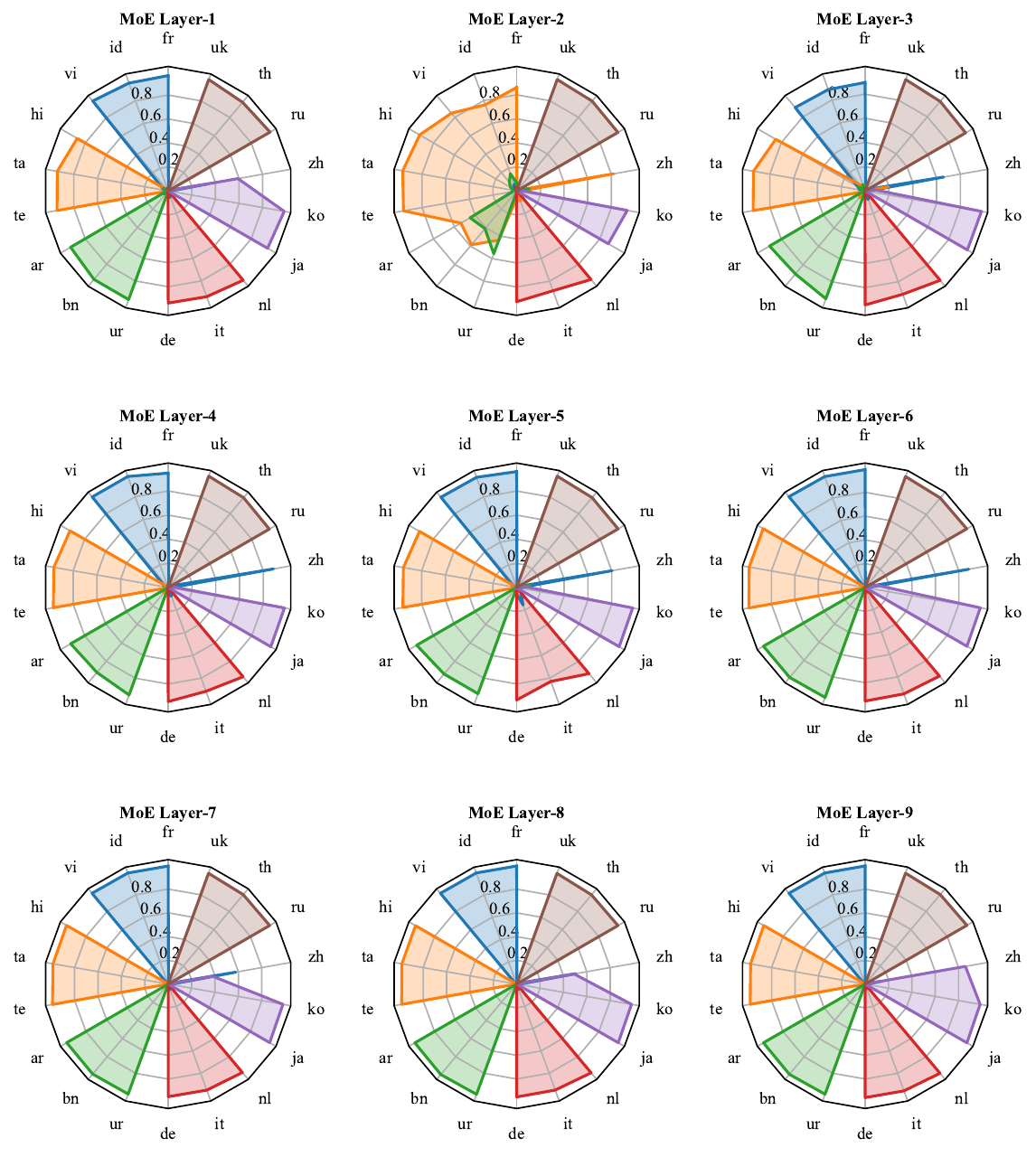}
\caption{\label{fig:tok_lang_add_loss_dist_all}The router distribution of top-1 expert for texts in different languages on models trained with router language classification loss.}
\vspace{-2mm}
\end{figure*}

We statisticize the top-1 expert distribution across the mixture-of-experts layers in Figure \ref{fig:tok_lang_dist_all} and \ref{fig:tok_lang_add_loss_dist_all}. 
As shown in Figure \ref{fig:tok_lang_dist_all}, the language specialization emerges at the last five MoE layers, while MoE layers often show language specialization with router language classification loss (Figure \ref{fig:tok_lang_add_loss_dist_all}). 

\subsection{Extend to 128 Languages}
\label{app:128langs_all_res}

We report the perplexity of all 128 languages on Table \ref{tab:128_ppl_p1} to \ref{tab:128_ppl_p7}. 
And in-context learning results on five multilingual datasets are shown in Table \ref{tab:all_xnli} to \ref{tab:all_xwinograd}. 

\input{tabs/128langs/part1}
\input{tabs/128langs/part2}
\input{tabs/128langs/part3}
\input{tabs/128langs/part4}
\input{tabs/128langs/part5}
\input{tabs/128langs/part6}
\input{tabs/128langs/part7}
\input{tabs/128langs/part8}

% TODO: Report the details of in-context learning results.

\input{tabs/in-context/xnli}

\input{tabs/in-context/pawsx}

\input{tabs/in-context/xcopa}

\input{tabs/in-context/xstorycloze}

\input{tabs/in-context/xwinograd}

\section{Licenses of Scientific Artifacts}
We follow and report the licenses of scientific artifacts involved in Table \ref{tab:license}. 

\input{tabs/licenses}

\section{Additional Information about Language Codes}
\label{app:lang_code}
Table \ref{tab:lang_codes} reports more information about the language codes involved in this work. 

\input{tabs/lang_codes}

\end{document}

%% file: tabs/bloom560M_18langs_group.tex
\begin{table}[htp]

\renewcommand\arraystretch{1.2}

\centering
\scriptsize

\setlength{\tabcolsep}{0.8mm}

 \begin{tabu}{l|cccccccccccccccccc}

 \toprule[1.2pt]

% high: ru, de, nl, ar, fr, it, ja, zh
% medium: vi, id, uk, hi, bn, ko, th
% low: ta, te, ur

 \multicolumn{1}{c}{} & \textbf{ar}&\textbf{ur}&\textbf{bn}&\textbf{it}&\textbf{de}&\textbf{nl}&\textbf{ta}&\textbf{te}&\textbf{hi}&\textbf{id}&\textbf{fr}&\textbf{vi}&\textbf{ru}&\textbf{uk}&\textbf{th}&\textbf{ko}&\textbf{ja}&\textbf{zh}\\

\midrule[0.8pt]

\textbf{9G}&\cellcolor[HTML]{31869B}1 &\cellcolor[HTML]{31869B}1&\cellcolor[HTML]{92CDDC}2&\cellcolor[HTML]{92CDDC}2&\cellcolor[HTML]{B7DEE8}3&\cellcolor[HTML]{B7DEE8}3&\cellcolor[HTML]{DAEEF3}4&\cellcolor[HTML]{DAEEF3}4&5&5&\cellcolor[HTML]{FDE9D9}6&\cellcolor[HTML]{FDE9D9}6&\cellcolor[HTML]{FCD5B4}7&\cellcolor[HTML]{FCD5B4}7&\cellcolor[HTML]{FAB58F}8&\cellcolor[HTML]{FAB58F}8&\cellcolor[HTML]{E26B0A}9&\cellcolor[HTML]{E26B0A}9\\

\textbf{6G}&\cellcolor[HTML]{31869B}1&\cellcolor[HTML]{31869B}1&\cellcolor[HTML]{31869B}1&\cellcolor[HTML]{92CDDC}2&\cellcolor[HTML]{92CDDC}2&\cellcolor[HTML]{92CDDC}2&\cellcolor[HTML]{B7DEE8}3&\cellcolor[HTML]{B7DEE8}3&\cellcolor[HTML]{B7DEE8}3&\cellcolor[HTML]{FCD5B4}4&\cellcolor[HTML]{FCD5B4}4&\cellcolor[HTML]{FCD5B4}4&\cellcolor[HTML]{FAB58F}5&\cellcolor[HTML]{FAB58F}5&\cellcolor[HTML]{FAB58F}5&\cellcolor[HTML]{E26B0A}6&\cellcolor[HTML]{E26B0A}6 &\cellcolor[HTML]{E26B0A}6\\

\textbf{6G${}^{(\text{L})}$}&\cellcolor[HTML]{31869B}1&\cellcolor[HTML]{31869B}1&\cellcolor[HTML]{92CDDC}2&\cellcolor[HTML]{B7DEE8}3&\cellcolor[HTML]{FCD5B4}4&\cellcolor[HTML]{B7DEE8}3&\cellcolor[HTML]{FAB58F}5&\cellcolor[HTML]{FAB58F}5&\cellcolor[HTML]{FAB58F}5&\cellcolor[HTML]{E26B0A}6&\cellcolor[HTML]{31869B}1&\cellcolor[HTML]{E26B0A}6&\cellcolor[HTML]{FCD5B4}4&\cellcolor[HTML]{FCD5B4}4&\cellcolor[HTML]{E26B0A}6&\cellcolor[HTML]{92CDDC}2&\cellcolor[HTML]{92CDDC}2&\cellcolor[HTML]{B7DEE8}3\\

\textbf{6G${}^{(\text{R})}$}&\cellcolor[HTML]{31869B}1&\cellcolor[HTML]{92CDDC}2&\cellcolor[HTML]{FAB58F}5&\cellcolor[HTML]{FCD5B4}4&\cellcolor[HTML]{E26B0A}6&\cellcolor[HTML]{E26B0A}6&\cellcolor[HTML]{92CDDC}2&\cellcolor[HTML]{B7DEE8}3&\cellcolor[HTML]{B7DEE8}3&\cellcolor[HTML]{FCD5B4}4&\cellcolor[HTML]{E26B0A}6&\cellcolor[HTML]{31869B}1&\cellcolor[HTML]{B7DEE8}3&\cellcolor[HTML]{FAB58F}5&\cellcolor[HTML]{FCD5B4}4&\cellcolor[HTML]{92CDDC}2&\cellcolor[HTML]{FAB58F}5&\cellcolor[HTML]{31869B}1\\

\textbf{3G}&\cellcolor[HTML]{31869B}1&\cellcolor[HTML]{31869B}1&\cellcolor[HTML]{31869B}1&2&2&2&\cellcolor[HTML]{31869B}1&\cellcolor[HTML]{E26B0A}3&\cellcolor[HTML]{31869B}1&\cellcolor[HTML]{31869B}1&2&2&\cellcolor[HTML]{E26B0A}3&\cellcolor[HTML]{E26B0A}3&\cellcolor[HTML]{E26B0A}3&\cellcolor[HTML]{E26B0A}3&\cellcolor[HTML]{E26B0A}3&2\\

\textbf{2G}&\cellcolor[HTML]{31869B}1&\cellcolor[HTML]{31869B}1&\cellcolor[HTML]{31869B}1&\cellcolor[HTML]{E26B0A}2&\cellcolor[HTML]{E26B0A}2&\cellcolor[HTML]{E26B0A}2&\cellcolor[HTML]{31869B}1&\cellcolor[HTML]{E26B0A}2&\cellcolor[HTML]{31869B}1&\cellcolor[HTML]{31869B}1&\cellcolor[HTML]{31869B}1&\cellcolor[HTML]{31869B}1&\cellcolor[HTML]{E26B0A}2&\cellcolor[HTML]{E26B0A}2&\cellcolor[HTML]{E26B0A}2&\cellcolor[HTML]{E26B0A}2&\cellcolor[HTML]{E26B0A}2&\cellcolor[HTML]{31869B}1\\

\bottomrule[1.2pt]
\end{tabu}
\vspace{-2mm}
\caption{\label{tab:18langs_group} The grouping results of $\text{BLOOM}_{\text{560M}}$, where ``\textbf{2G}'' denotes the result that divides into 2 groups. ``\textbf{6G${}^{(\text{L})}$}'' and ``\textbf{6G${}^{(\text{R})}$}'' indicate the LANG2VEC and random language clustering results, respectively. 
}
\vspace{-4mm}
\end{table}

%% file: tabs/main_18langs_ppl.tex
\begin{table*}[htp]

\renewcommand\arraystretch{1.1}

\centering
\scriptsize

\setlength{\tabcolsep}{0.6mm}

 \begin{tabu}{lr|cccccccc|ccccccc|ccc|c}
 
 \toprule[1.2pt]
  
  \multicolumn{2}{c}{ } & \multicolumn{8}{c}{\textbf{High}} & \multicolumn{7}{c}{\textbf{Medium}} & \multicolumn{3}{c}{\textbf{Low}} & \multicolumn{1}{c}{ } \\

  \cmidrule(r){3-10} \cmidrule(r){11-17} \cmidrule(r){18-20} \noalign{\smallskip}

% high: ru, de, nl, ar, fr, it, ja, zh
% medium: vi, id, uk, hi, bn, ko, th
% low: ta, te, ur

 \multicolumn{1}{c}{\textbf{Model}} & \multicolumn{1}{c}{$\#$\textbf{Param.}} & \textbf{ar} & \textbf{de}${}^{\dagger}$ & \textbf{fr} & \textbf{it}${}^{\dagger}$ & \textbf{ja}${}^{\dagger}$ & \textbf{nl}${}^{\dagger}$ & \textbf{ru}${}^{\dagger}$ & \textbf{zh} & \textbf{bn} & \textbf{hi} & \textbf{id} & \textbf{ko}${}^{\dagger}$ & 
 \textbf{th}${}^{\dagger}$ & \textbf{uk}${}^{\dagger}$ & \textbf{vi}${}^{\dagger}$ & \textbf{ta} & \textbf{te} &  
 \multicolumn{1}{c}{\textbf{ur}} & \multicolumn{1}{c}{\textbf{Avg }$\downarrow$} \\

\midrule[0.8pt]

$\text{BLOOM}_{\text{560M}}$
&560M&$56.7$&$126.4$&$37.4$&$85.3$&$55.7$&$129.5$&$31.3$&$59.0$&$42.3$&$32.7$&$47.4$&$25.7$&$14.0$&$39.7$&$27.9$&$64.2$&$88.4$&$60.0$&$56.9$\\

$\text{\tiny \ \ \ + Pre-train\ }$
&560M&$39.0$&$27.6$&$22.0$&$17.8$&$15.0$&$16.6$&$8.2$&$36.0$&$26.5$&$20.9$&$32.6$&$8.4$&$4.4$&$6.8$&$18.5$&$31.0$&$26.9$&$29.8$&$21.6$\\ 

   \cdashlinelr{1-21}
$\text{X-ELM}$
&5.03B&$\textbf{35.2}$&$34.6$&$21.3$&$17.3$&$18.0$&$21.5$&$9.8$&$37.9$&$25.1$&$19.5$&$\textbf{28.8}$&$9.3$&$4.4$&$8.5$&$18.9$&$\textbf{26.9}$&$\textbf{23.0}$&$27.5$&$21.5$\\ 

$\text{Branch-Train-Mix}$
&1.57B&$39.4$&$25.8$&$21.6$&$17.1$&$14.7$&$15.3$&$8.0$&$35.8$&$26.4$&$21.0$&$32.1$&$8.1$&$4.1$&$6.6$&$18.1$&$31.4$&$27.0$&$30.0$&$21.2$\\ 

$\text{DMoE (2 Groups)}$
&635M&$37.9$&$25.0$&$21.5$&$17.1$&$15.0$&$14.5$&$7.4$&$34.4$&$25.4$&$20.5$&$31.3$&$7.9$&$3.8$&$5.9$&$18.4$&$30.3$&$26.7$&$28.6$&$20.6$\\ 

$\text{DMoE (3 Groups)}$
&710M&$37.0$&$25.3$&$21.8$&$17.2$&$14.4$&$14.4$&$\textbf{7.1}$&$35.0$&$24.6$&$19.9$&$30.4$&$7.7$&$3.7$&$\textbf{5.6}$&$18.4$&$29.3$&$26.3$&$27.4$&$20.3$\\

$\text{DMoE (6 Groups)}$
&937M&$\underline{36.2}$&$\underline{23.5}$&$\textbf{20.9}$&$16.0$&$\underline{13.8}$&$\underline{13.6}$&$7.3$&$\textbf{34.1}$&$\underline{23.9}$&$\textbf{19.1}$&$30.7$&$\underline{7.4}$&$3.9$&$5.9$&$\textbf{17.6}$&$27.9$&$23.4$&$\textbf{26.8}$&$\textbf{19.5}$\\

$\text{\tiny \ \ \ w/ Gemma Clusters}$
&937M&$36.7$&$\underline{23.5}$&$21.4$&$16.1$&$\underline{13.8}$&$\underline{13.6}$&$7.3$&$\textbf{34.1}$&$\textbf{23.7}$&$19.7$&$31.8$&$\underline{7.4}$&$\underline{3.6}$&$\textbf{5.6}$&$\underline{17.7}$&$27.8$&$\underline{23.3}$&$27.1$&$\underline{19.7}$\\

$\text{\tiny \ \ \ w/ LANG2VEC Clusters}$
&937M&$37.6$&$25.2$&$\textbf{20.9}$&$\underline{15.8}$&$\textbf{13.4}$&$\textbf{13.3}$&$7.6$&$\underline{34.2}$&$25.0$&$\underline{19.3}$&$30.3$&$\textbf{7.2}$&$\textbf{3.5}$&$6.2$&$\underline{17.7}$&$28.0$&$23.6$&$28.1$&$19.8$\\

$\text{\tiny \ \ \ w/ Random Clusters}$
&937M&$38.5$&$25.7$&$21.5$&$17.0$&$14.6$&$15.2$&$7.9$&$35.3$&$25.7$&$20.6$&$31.8$&$8.1$&$4.1$&$6.5$&$18.1$&$30.6$&$26.1$&$29.4$&$20.9$\\

$\text{\tiny \ \ \ w/o Class. Loss}$
&937M&$36.5$&$24.2$&$21.2$&$16.4$&$14.4$&$13.9$&$\underline{7.2}$&$34.8$&$24.6$&$19.7$&$30.9$&$7.7$&$3.8$&$\underline{5.8}$&$18.2$&$28.4$&$24.1$&$27.6$&$20.0$\\

$\text{DMoE (9 Groups)}$
&1.16B&$36.9$&$\textbf{23.1}$&$\underline{21.1}$&$\textbf{14.7}$&$14.0$&$\textbf{13.3}$&$7.5$&$34.6$&$25.1$&$19.6$&$\underline{29.7}$&$\textbf{7.2}$&$\underline{3.6}$&$6.1$&$17.9$&$\underline{27.6}$&$\textbf{23.0}$&$\underline{27.0}$&$\textbf{19.5}$\\

\midrule[0.8pt]

$\text{BLOOM}_{\text{1.7B}}$
&1.72B&$41.2$&$63.0$&$24.3$&$44.0$&$35.2$&$63.0$&$20.1$&$40.1$&$27.0$&$22.6$&$32.7$&$17.4$&$9.8$&$23.9$&$19.2$&$40.1$&$43.6$&$36.3$&$33.5$\\

$\text{\tiny \ \ \ + Pre-train\ }$
&1.72B&$\underline{25.1}$&$21.0$&$\underline{15.3}$&$15.5$&$11.7$&$17.4$&$7.0$&$\textbf{23.4}$&$\textbf{16.8}$&$15.3$&$23.0$&$7.8$&$4.3$&$8.5$&$\underline{13.1}$&$21.6$&$21.1$&$22.6$&$16.1$\\

   \cdashlinelr{1-21}
$\text{X-ELM}$
&15.50B&$\textbf{24.9}$&$20.2$&$\textbf{15.0}$&$\underline{12.1}$&$12.1$&$12.3$&$7.1$&$24.3$&$\underline{17.4}$&$\textbf{14.4}$&$\textbf{21.0}$&$7.5$&$3.8$&$6.0$&$\textbf{12.9}$&$\textbf{19.3}$&$\textbf{16.6}$&$\textbf{19.4}$&$14.8$\\

$\text{Branch-Train-Mix}$
&5.75B&$25.4$&$18.7$&$16.0$&$15.2$&$11.3$&$13.6$&$6.0$&$\underline{23.6}$&$17.6$&$15.3$&$24.1$&$7.4$&$3.7$&$7.0$&$13.4$&$23.1$&$21.1$&$20.5$&$15.7$\\

   % \cdashlinelr{1-16}

$\text{DMoE (3 Groups)}$
&2.33B&$26.3$&$17.5$&$16.0$&$12.5$&$11.0$&$10.5$&$\textbf{5.7}$&$25.2$&$17.5$&$15.1$&$22.4$&$6.3$&$\underline{3.2}$&$\textbf{4.5}$&$13.8$&$21.1$&$19.6$&$19.9$&$14.9$\\

$\text{DMoE (6 Groups)}$
&3.23B&$26.1$&$\underline{17.2}$&$15.8$&$12.4$&$\textbf{10.7}$&$\underline{10.2}$&$\underline{5.8}$&$24.9$&$\underline{17.4}$&$\underline{14.5}$&$22.8$&$\underline{6.1}$&$3.3$&$\underline{4.6}$&$13.3$&$\underline{20.3}$&$17.5$&$\underline{19.6}$&$\underline{14.6}$\\

$\text{DMoE (9 Groups)}$
&4.14B&$26.6$&$\textbf{16.8}$&$16.0$&$\textbf{11.5}$&$\underline{10.8}$&$\textbf{10.0}$&$5.9$&$25.1$&$17.8$&$14.8$&$\underline{22.2}$&$\textbf{5.9}$&$\textbf{3.1}$&$4.8$&$13.5$&$20.1$&$\underline{17.0}$&$19.8$&$\textbf{14.5}$\\

\midrule[0.8pt]

$\text{Gemma}_{\text{2B}}$
&2.51B&$54.8$&$12.5$&$23.6$&$13.4$&$11.1$&$11.4$&$5.1$&$69.1$&$68.9$&$44.2$&$45.5$&$5.6$&$2.8$&$4.0$&$19.8$&$70.5$&$62.8$&$54.2$&$32.2$\\

$\text{\tiny \ \ \ + Pre-train\ }$
&2.51B&$28.8$&$\underline{10.1}$&$\underline{17.1}$&$\underline{9.3}$&$\underline{7.3}$&$\underline{6.5}$&$\underline{4.1}$&$\underline{29.4}$&$21.7$&$14.9$&$\underline{21.9}$&$\underline{3.8}$&$\underline{2.3}$&$\textbf{3.1}$&$\underline{13.6}$&$20.4$&$17.0$&$20.4$&$14.0$\\

   \cdashlinelr{1-21}
$\text{X-ELM}$
&22.56B&$30.2$&$11.1$&$19.3$&$10.2$&$8.0$&$7.3$&$4.3$&$34.6$&$22.7$&$15.3$&$23.9$&$3.9$&$\underline{2.3}$&$3.3$&$14.4$&$19.9$&$16.5$&$22.8$&$15.0$\\

$\text{Branch-Train-Mix}$
&11.57B&$\underline{27.7}$&$10.7$&$18.2$&$9.7$&$7.5$&$6.8$&$4.2$&$30.6$&$\underline{18.5}$&$\underline{14.4}$&$23.8$&$3.9$&$\underline{2.3}$&$\underline{3.2}$&$\underline{13.6}$&$\underline{17.4}$&$\underline{14.5}$&$\underline{19.3}$&$\underline{13.7}$\\

   % \cdashlinelr{1-16}

$\text{DMoE (9 Groups)}$
&6.53B&$\textbf{24.8}$&$\textbf{9.9}$&$\textbf{16.8}$&$\textbf{9.1}$&$\textbf{7.0}$&$\textbf{6.3}$&$\textbf{4.0}$&$\textbf{27.8}$&$\textbf{17.6}$&$\textbf{12.9}$&$\textbf{19.6}$&$\textbf{3.6}$&$\textbf{2.2}$&$\textbf{3.1}$&$\textbf{12.2}$&$\textbf{15.3}$&$\textbf{12.8}$&$\textbf{17.3}$&$\textbf{12.4}$\\

\bottomrule[1.2pt]
\end{tabu}
\vspace{-2mm}
\caption{\label{tab:18langs_ppl} The normalized perplexity on the valid split of CulturaX. The perplexity is normalized to the vocabulary of Bloom following \citet{wei2023skywork}. ${}^{\dagger}$ denotes the language unseen in the pre-training of BLOOM. ``\textbf{High}'', ``\textbf{Medium}'', and ``\textbf{Low}'' indicates the available amount of linguistic resources. The best and second results are denoted in \textbf{bold} and \underline{underlined}, correspondingly.
}
\vspace{-5mm}
\end{table*}

%% file: tabs/add4langs.tex
\begin{table}[htp]

\renewcommand\arraystretch{1.4}

\centering
\scriptsize

\setlength{\tabcolsep}{0.7mm}

 \begin{tabu}{l|cccc|ccc}
 
 \toprule[1.2pt]
  
  \multicolumn{1}{c}{ } & \multicolumn{4}{c}{\textbf{New Languages}} & \multicolumn{3}{c}{\textbf{Old Languages}} \\

  \cmidrule(r){2-5} \cmidrule(r){6-8} \noalign{\smallskip}

% high: ru, de, nl, ar, fr, it, ja, zh
% medium: vi, id, uk, hi, bn, ko, th
% low: ta, te, ur

 \multicolumn{1}{c}{\textbf{Model}} & \textbf{be} & \textbf{ml} & \textbf{mr} & \textbf{sr}  & \textbf{High} & \textbf{Medium} & \textbf{Low} \\

\midrule[0.8pt]

$\text{Gemma}_{\text{2B}}$
&$10.0$&$7.1$&$11.4$&$12.5$&$26.9_{\pm 9.4}$&$15.1_{\pm 7.9}$&$10.5_{\pm 4.3}$\\

$\text{\ + Pre-train\ }$
&$9.3$&$11.7$&$11.2$&$17.1$&$15.1_{\pm 4.4}$&$\underline{8.5}_{\pm 3.7}$&$\underline{
5.4}_{\pm 1.7}$\\

$\text{\ \ w/ LAPT\ }$
&$\underline{6.4}$&$\underline{4.3}$&$\underline{6.1}$&$\underline{8.3}$&$16.5_{\pm 4.6}$&$10.6_{\pm 3.8}$&$7.3_{\pm 3.1} $\\

   \cdashlinelr{1-8}

$\text{DMoE}$
&$8.9$&$10.0$&$11.8$&$17.2$&$\textbf{14.4}_{\pm 4.2}$&$\textbf{7.7}_{\pm 3.4}$&$\textbf{4.8}_{\pm 1.6}$\\

$\text{\ \ w/ DLA}$
&$\textbf{6.2}$&$\textbf{4.0}$&$\textbf{5.4}$&$\textbf{8.2}$&$\underline{15.0}_{\pm 4.2}$&$\underline{8.5}_{\pm 3.2}$&$\underline{5.4}_{\pm 1.8}$\\

\bottomrule[1.2pt]
\end{tabu}
\vspace{-1mm}
\caption{\label{tab:18langs_add4} The perplexity after adding new languages. 
}
\vspace{-2mm}
\end{table}

%% file: tabs/main_128langs_ppl.tex
\begin{table*}[htp]

\renewcommand\arraystretch{1.1}

\centering
\scriptsize

\setlength{\tabcolsep}{0.4mm}

 \begin{tabu}{l|rrrrr|rrrrr|rrrrr|rrrrr|r}
 
 \toprule[1.2pt]
  
  \multicolumn{1}{c}{ } & \multicolumn{5}{c}{\textbf{High}} & \multicolumn{5}{c}{\textbf{Medium}} & \multicolumn{5}{c}{\textbf{Low}} & \multicolumn{5}{c}{\textbf{Extremely-Low}} & \multicolumn{1}{c}{\textbf{ALL 128L}} \\

  \cmidrule(r){2-6} \cmidrule(r){7-11} \cmidrule(r){12-16} \cmidrule(r){17-21} \cmidrule(r){22-22} \noalign{\smallskip}

% high: ru, de, nl, ar, fr, it, ja, zh
% medium: vi, id, uk, hi, bn, ko, th
% low: ta, te, ur

 \multicolumn{1}{c}{\textbf{Model}} & \textbf{ar} & \textbf{de}${}^{\dagger}$ & \textbf{en} & \textbf{it}${}^{\dagger}$ & \textbf{ja}${}^{\dagger}$ & \textbf{hi} & \textbf{id} & \textbf{th}${}^{\dagger}$ & \textbf{uk}${}^{\dagger}$ & \textbf{vi}${}^{\dagger}$ & \textbf{kk}${}^{\dagger}$ & \textbf{mn}${}^{\dagger}$ & 
 \textbf{my}${}^{\dagger}$ & \textbf{te} & \textbf{ur} & \textbf{br}${}^{\dagger}$ & \textbf{pa}${}^{\dagger}$ & \textbf{sw} & \textbf{ug}${}^{\dagger}$ & \multicolumn{1}{c}{\textbf{zu}} & \multicolumn{1}{c}{\textbf{Avg} $\downarrow$} \\

\midrule[0.8pt]

$\text{BLOOM}_{\text{560M}}$
&$42.4$&$111.3$&$66.8$&$82.4$&$55.5$&$30.6$&$41.3$&$13.7$&$44.5$&$21.8$&$29.0$&$31.2$&$6.1$&$91.4$&$72.5$&$261.7$&$131.6$&$224.9$&$44.5$&$1278.9$&$154.4_{\pm157.0}$\\

$\text{\ + Pre-train\ }$
&$34.4$&$23.8$&$44.6$&$20.5$&$12.9$&$20.5$&$26.9$&$3.6$&$7.3$&$15.1$&$5.7$&$6.8$&$\underline{2.8}$&$\underline{30.3}$&$37.6$&$40.2$&$\underline{32.5}$&$45.5$&$9.2$&$36.2$&$20.7_{\pm12.8}$\\

   \cdashlinelr{1-22}

$\text{Branch-Train-Mix}$
&$\underline{32.5}$&$\underline{21.0}$&$\textbf{40.3}$&$\underline{17.1}$&$\underline{12.1}$&$\underline{20.0}$&$\underline{26.1}$&$\underline{3.5}$&$\underline{6.6}$&$\textbf{14.6}$&$\underline{5.4}$&$\underline{6.3}$&$\underline{2.8}$&$31.6$&$\underline{37.1}$&$\underline{35.4}$&$34.1$&$\underline{43.5}$&$\underline{9.0}$&$\underline{31.2}$&$\underline{19.2}_{\pm12.2}$\\

% $\text{DaMoE (2 Groups)}$
% &$-$&$-$&$-$&$-$&$-$&$-$&$-$&$-$&$-$&$-$&$-$&$-$&$-$&$-$&$-$&$-$&$-$&$-$&$-$&$-$&$-$\\

$\text{DMoE}$
&$\textbf{32.1}$&$\textbf{19.5}$&$\underline{40.8}$&$\textbf{16.3}$&$\textbf{11.3}$&$\textbf{19.8}$&$\textbf{25.8}$&$\textbf{3.4}$&$\textbf{6.4}$&$\underline{14.7}$&$\textbf{5.2}$&$\textbf{6.2}$&$\textbf{2.7}$&$\textbf{29.3}$&$\textbf{35.2}$&$\textbf{31.4}$&$\textbf{30.5}$&$\textbf{39.7}$&$\textbf{8.1}$&$\textbf{28.3}$&$\textbf{17.7}_{\pm11.2}$\\

\midrule[0.8pt]

$\text{BLOOM}_{\text{1.7B}}$
&$30.4$&$56.4$&$45.3$&$44.9$&$35.0$&$20.8$&$27.8$&$9.7$&$26.7$&$15.3$&$19.1$&$21.1$&$4.4$&$46.8$&$44.0$&$113.2$&$61.7$&$80.3$&$28.4$&$260.8$&$71.9_{\pm52.1}$\\

$\text{\ + Pre-train\ }$
&$22.9$&$15.9$&$\underline{30.3}$&$14.1$&$9.7$&$\underline{14.7}$&$19.3$&$\underline{3.1}$&$5.4$&$\underline{11.3}$&$\underline{4.5}$&$\underline{5.4}$&$\textbf{2.5}$&$\underline{22.3}$&$26.8$&$27.9$&$\underline{23.9}$&$30.8$&$\underline{7.4}$&$26.0$&$14.9_{\pm8.7}$\\

   \cdashlinelr{1-22}
   
$\text{Branch-Train-Mix}$
&$\textbf{21.1}$&$\underline{15.3}$&$\textbf{29.7}$&$\underline{13.1}$&$\underline{9.4}$&$\textbf{14.1}$&$\textbf{18.6}$&$\underline{3.1}$&$\underline{5.3}$&$\textbf{10.9}$&$\underline{4.5}$&$\underline{5.4}$&$\underline{2.6}$&$22.4$&$\textbf{25.7}$&$\underline{26.2}$&$24.2$&$\underline{29.8}$&$7.5$&$\underline{24.1}$&$\underline{14.3}_{\pm8.4}$\\

   % \cdashlinelr{1-16}

% $\text{DaMoE (2 Groups)}$
% &$-$&$-$&$-$&$-$&$-$&$-$&$-$&$-$&$-$&$-$&$-$&$-$&$-$&$-$&$-$&$-$&$-$&$-$&$-$&$-$&$-$\\

$\text{DMoE}$
&$\underline{22.7}$&$\textbf{14.9}$&$31.3$&$\textbf{12.8}$&$\textbf{8.9}$&$\underline{14.7}$&$\underline{19.1}$&$\textbf{3.0}$&$\textbf{5.1}$&$\underline{11.3}$&$\textbf{4.3}$&$\textbf{5.1}$&$\textbf{2.5}$&$\textbf{22.1}$&$\underline{26.0}$&$\textbf{23.9}$&$\textbf{23.2}$&$\textbf{29.2}$&$\textbf{6.8}$&$\textbf{22.8}$&$\textbf{13.7}_{\pm8.1}$\\

\bottomrule[1.2pt]
\end{tabu}
\vspace{-2mm}
\caption{\label{tab:128_ppl} The perplexity of 20 languages on the valid split of MADLAD-400 \citep{kudugunta2023madlad}. Refer to Table \ref{tab:128_ppl_p1} to \ref{tab:128_ppl_p8} in Appendix \ref{app:128langs_all_res} for all results of 128 languages. ${}^{\dagger}$ denotes the language unseen in the pre-training of BLOOM. ``\textbf{High}''(>1\%), ``\textbf{Medium}''(>0.1\%), ``\textbf{Low}''(>0.01\%), and ``\textbf{Extremely-Low}''(<=0.01\%) indicates the available amount of linguistic resources on the CommonCrawl following \citet{lai-etal-2023-okapi}. 
}
\vspace{-2mm}
\end{table*}

%% file: tabs/main_128langs_in-context.tex
\begin{table*}[htp]

\renewcommand\arraystretch{1.1}

\centering
\scriptsize

\setlength{\tabcolsep}{1.2mm}

 \begin{tabu}{l|ccccc|ccccc}
 
 \toprule[1.2pt]
  
  \multicolumn{1}{c}{ } & \multicolumn{5}{c}{\textbf{Zero-shot Results}} & \multicolumn{5}{c}{\textbf{Few-shot Results}} \\

  \cmidrule(r){2-6} \cmidrule(r){7-11} \noalign{\smallskip}

% high: ru, de, nl, ar, fr, it, ja, zh
% medium: vi, id, uk, hi, bn, ko, th
% low: ta, te, ur

 \multicolumn{1}{c}{\textbf{Model}} & \textbf{XNLI} & \textbf{PAWS-X} & \textbf{XCOPA} & \textbf{XStoryCloze} & \textbf{XWinograd} & \textbf{XNLI} & \textbf{PAWS-X} & \textbf{XCOPA} & \textbf{XStoryCloze} & \textbf{XWinograd} \\

\midrule[0.8pt]

$\text{BLOOM}_{\text{560M}}$
&${36.2}_{\pm3.3}$&${51.5}_{\pm1.6}$&${53.9}_{\pm4.1}$&${53.5}_{\pm3.5}$&${53.7}_{\pm4.0}$&${34.4}_{\pm2.4}$&${51.1}_{\pm1.2}$&${53.4}_{\pm4.0}$&${52.6}_{\pm3.5}$&${53.3}_{\pm4.2}$\\

$\text{\ \ \ + Pre-train\ }$
&${37.1}_{\pm3.5}$&${52.9}_{\pm2.4}$&${53.6}_{\pm2.9}$&${\underline{53.8}}_{\pm2.6}$&${\underline{54.9}}_{\pm4.1}$&${34.7}_{\pm2.6}$&${\underline{51.6}}_{\pm1.1}$&${53.8}_{\pm2.4}$&${52.7}_{\pm2.8}$&${53.8}_{\pm5.1}$\\

   \cdashlinelr{1-11}
   
$\text{Branch-Train-Mix}$
&${\underline{37.2}}_{\pm4.1}$&${\underline{53.1}}_{\pm2.5}$&${\underline{54.1}}_{\pm2.7}$&${\underline{53.8}}_{\pm2.8}$&${54.4}_{\pm3.6}$&${\underline{35.3}}_{\pm2.6}$&${51.4}_{\pm2.1}$&${\underline{53.9}}_{\pm3.2}$&${\underline{53.1}}_{\pm2.8}$&${\underline{54.2}}_{\pm4.3}$\\

$\text{DMoE}$
&${\textbf{37.5}}_{\pm4.5}$&${\textbf{53.2}}_{\pm1.7}$&${\textbf{54.4}}_{\pm2.8}$&${\textbf{54.1}}_{\pm2.7}$&${\textbf{55.1}}_{\pm4.3}$&${\textbf{35.7}}_{\pm2.5}$&${\textbf{52.2}}_{\pm1.4}$&${\textbf{54.7}}_{\pm2.6}$&${\textbf{53.4}}_{\pm2.7}$&${\textbf{55.1}}_{\pm4.6}$\\

\midrule[0.8pt]

$\text{BLOOM}_{\text{1.7B}}$
&${\underline{39.2}}_{\pm5.4}$&${\underline{53.9}}_{\pm1.6}$&${55.1}_{\pm5.7}$&${56.0}_{\pm4.7}$&${55.1}_{\pm5.2}$&${\underline{37.1}}_{\pm4.5}$&${50.5}_{\pm1.1}$&${55.2}_{\pm5.8}$&${55.5}_{\pm5.2}$&${55.5}_{\pm5.2}$\\

$\text{\ \ \ + Pre-train\ }$
&${\underline{39.2}}_{\pm4.9}$&${53.7}_{\pm1.5}$&${55.0}_{\pm3.9}$&${\underline{56.4}}_{\pm3.6}$&${55.5}_{\pm4.6}$&${\underline{37.1}}_{\pm3.2}$&${\underline{51.8}}_{\pm1.7}$&${55.3}_{\pm4.5}$&${56.0}_{\pm3.6}$&${\underline{55.8}}_{\pm4.6}$\\

   \cdashlinelr{1-11}
   
$\text{Branch-Train-Mix}$
&${39.1}_{\pm5.1}$&${53.5}_{\pm1.6}$&${\underline{55.6}}_{\pm3.6}$&${\underline{56.4}}_{\pm3.3}$&${\underline{55.6}}_{\pm4.7}$&${36.8}_{\pm3.2}$&${51.3}_{\pm1.4}$&${\underline{55.5}}_{\pm4.3}$&${\textbf{56.2}}_{\pm4.0}$&${\underline{55.8}}_{\pm4.4}$\\

$\text{DMoE}$
&${\textbf{39.8}}_{\pm4.8}$&${\textbf{54.1}}_{\pm1.2}$&${\textbf{56.0}}_{\pm4.1}$&${\textbf{56.6}}_{\pm3.5}$&${\textbf{56.4}}_{\pm5.1}$&${\textbf{37.5}}_{\pm2.9}$&${\textbf{52.2}}_{\pm2.4}$&${\textbf{55.7}}_{\pm3.8}$&${\underline{56.1}}_{\pm3.4}$&${\textbf{56.4}}_{\pm5.1}$\\

\bottomrule[1.2pt]
\end{tabu}
\vspace{-2mm}
\caption{\label{tab:128langs_incontext} The in-context learning results of models after training on 128 languages. The number of demonstration samples in the ``Few-shot'' setting is set to four in this work. Table \ref{tab:all_xnli} to \ref{tab:all_xwinograd} report results of all languages. 
}
\vspace{-2mm}
\end{table*}

%% file: tabs/bloom1.7B_18langs_group.tex
\begin{table}[htp]

\renewcommand\arraystretch{1.3}

\centering
\scriptsize

\setlength{\tabcolsep}{0.8mm}

 \begin{tabu}{l|cccccccccccccccccc}
 
 \toprule[1.2pt]

% high: ru, de, nl, ar, fr, it, ja, zh
% medium: vi, id, uk, hi, bn, ko, th
% low: ta, te, ur

 \multicolumn{1}{c}{} & \textbf{ar}&\textbf{ur}&\textbf{bn}&\textbf{it}&\textbf{de}&\textbf{nl}&\textbf{ta}&\textbf{te}&\textbf{hi}&\textbf{id}&\textbf{fr}&\textbf{vi}&\textbf{ru}&\textbf{uk}&\textbf{th}&\textbf{ko}&\textbf{ja}&\textbf{zh}\\

\midrule[0.8pt]

% ('ar', 'id'):1
% ('ko', 'ur'):2
% ('bn', 'hi'):3
% ('it', 'th'):4
% ('de', 'nl'):5
% ('ta', 'te'):6
% ('fr', 'vi'):7
% ('ru', 'uk'):8
% ('ja', 'zh'):9

\textbf{9G}&\cellcolor[HTML]{31869B}1 &\cellcolor[HTML]{92CDDC}2 &\cellcolor[HTML]{B7DEE8}3 &\cellcolor[HTML]{DAEEF3}4 & 5 & 5 &\cellcolor[HTML]{FDE9D9}6 &\cellcolor[HTML]{FDE9D9}6 &\cellcolor[HTML]{B7DEE8}3 &\cellcolor[HTML]{31869B}1 &\cellcolor[HTML]{FCD5B4}7 &\cellcolor[HTML]{FCD5B4}7 &\cellcolor[HTML]{FAB58F}8&\cellcolor[HTML]{FAB58F}8&\cellcolor[HTML]{DAEEF3}4 &\cellcolor[HTML]{92CDDC}2 &\cellcolor[HTML]{E26B0A}9&\cellcolor[HTML]{E26B0A}9\\

\textbf{6G}&\cellcolor[HTML]{31869B}1&\cellcolor[HTML]{31869B}1&\cellcolor[HTML]{31869B}1&\cellcolor[HTML]{92CDDC}2&\cellcolor[HTML]{92CDDC}2&\cellcolor[HTML]{92CDDC}2&\cellcolor[HTML]{B7DEE8}3&\cellcolor[HTML]{B7DEE8}3&\cellcolor[HTML]{B7DEE8}3&\cellcolor[HTML]{FCD5B4}4&\cellcolor[HTML]{FCD5B4}4&\cellcolor[HTML]{FCD5B4}4&\cellcolor[HTML]{FAB58F}5&\cellcolor[HTML]{FAB58F}5&\cellcolor[HTML]{FAB58F}5&\cellcolor[HTML]{E26B0A}6&\cellcolor[HTML]{E26B0A}6 &\cellcolor[HTML]{E26B0A}6\\

\textbf{3G}&\cellcolor[HTML]{31869B}1 & 2 & 2 &\cellcolor[HTML]{E26B0A}3 &\cellcolor[HTML]{E26B0A}3 &\cellcolor[HTML]{E26B0A}3 & 2 & 2 &\cellcolor[HTML]{31869B}1 &\cellcolor[HTML]{31869B}1 &\cellcolor[HTML]{31869B}1 &\cellcolor[HTML]{31869B}1 & 2 &\cellcolor[HTML]{E26B0A}3 &\cellcolor[HTML]{E26B0A}3 & 2 &\cellcolor[HTML]{E26B0A}3 &\cellcolor[HTML]{31869B}1 \\

\textbf{2G}&\cellcolor[HTML]{31869B}1&\cellcolor[HTML]{31869B}1&\cellcolor[HTML]{31869B}1&\cellcolor[HTML]{E26B0A}2&\cellcolor[HTML]{E26B0A}2&\cellcolor[HTML]{E26B0A}2&\cellcolor[HTML]{31869B}1&\cellcolor[HTML]{E26B0A}2&\cellcolor[HTML]{31869B}1&\cellcolor[HTML]{31869B}1&\cellcolor[HTML]{31869B}1&\cellcolor[HTML]{31869B}1&\cellcolor[HTML]{E26B0A}2&\cellcolor[HTML]{E26B0A}2&\cellcolor[HTML]{E26B0A}2&\cellcolor[HTML]{E26B0A}2&\cellcolor[HTML]{E26B0A}2&\cellcolor[HTML]{31869B}1\\

\bottomrule[1.2pt]
\end{tabu}
\vspace{-2mm}
\caption{\label{tab:18langs_group_bloom1b7} The grouping results of $\text{BLOOM}_{\text{1.7B}}$, where ``\textbf{2G}'' denotes the result that divides into 2 groups. 
}
\vspace{-2mm}
\end{table}

%% file: tabs/gemma2B_18langs_group.tex
\begin{table}[htp]

\renewcommand\arraystretch{1.3}

\centering
\scriptsize

\setlength{\tabcolsep}{0.8mm}

 \begin{tabu}{l|cccccccccccccccccc}
 
 \toprule[1.2pt]

% high: ru, de, nl, ar, fr, it, ja, zh
% medium: vi, id, uk, hi, bn, ko, th
% low: ta, te, ur

 \multicolumn{1}{c}{} & \textbf{ar}&\textbf{ur}&\textbf{bn}&\textbf{it}&\textbf{de}&\textbf{nl}&\textbf{ta}&\textbf{te}&\textbf{hi}&\textbf{id}&\textbf{fr}&\textbf{vi}&\textbf{ru}&\textbf{uk}&\textbf{th}&\textbf{ko}&\textbf{ja}&\textbf{zh}\\

\midrule[0.8pt]

\textbf{9G}&\cellcolor[HTML]{31869B}1&\cellcolor[HTML]{31869B}1&\cellcolor[HTML]{92CDDC}2&\cellcolor[HTML]{FDE9D9}6&\cellcolor[HTML]{B7DEE8}3&\cellcolor[HTML]{B7DEE8}3&\cellcolor[HTML]{DAEEF3}4&\cellcolor[HTML]{DAEEF3}4&\cellcolor[HTML]{92CDDC}2&5&\cellcolor[HTML]{FDE9D9}6&5&\cellcolor[HTML]{FCD5B4}7&\cellcolor[HTML]{FCD5B4}7&\cellcolor[HTML]{FAB58F}8&\cellcolor[HTML]{FAB58F}8&\cellcolor[HTML]{E26B0A}9&\cellcolor[HTML]{E26B0A}9\\

\textbf{6G}&\cellcolor[HTML]{31869B}1&\cellcolor[HTML]{31869B}1&\cellcolor[HTML]{92CDDC}2&\cellcolor[HTML]{B7DEE8}3&\cellcolor[HTML]{FCD5B4}4&\cellcolor[HTML]{FCD5B4}4&\cellcolor[HTML]{92CDDC}2&\cellcolor[HTML]{92CDDC}2&\cellcolor[HTML]{31869B}1&\cellcolor[HTML]{B7DEE8}3&\cellcolor[HTML]{FCD5B4}4&\cellcolor[HTML]{FAB58F}5&\cellcolor[HTML]{B7DEE8}3&\cellcolor[HTML]{FAB58F}5&\cellcolor[HTML]{FAB58F}5&\cellcolor[HTML]{E26B0A}6&\cellcolor[HTML]{E26B0A}6&\cellcolor[HTML]{E26B0A}6\\

\textbf{3G}&\cellcolor[HTML]{31869B}1&\cellcolor[HTML]{31869B}1&\cellcolor[HTML]{31869B}1&2&2&2&\cellcolor[HTML]{31869B}1&\cellcolor[HTML]{31869B}1&\cellcolor[HTML]{31869B}1&2&2&\cellcolor[HTML]{E26B0A}3&2&\cellcolor[HTML]{E26B0A}3&\cellcolor[HTML]{E26B0A}3&\cellcolor[HTML]{E26B0A}3&\cellcolor[HTML]{E26B0A}3&\cellcolor[HTML]{E26B0A}3\\

\textbf{2G}&\cellcolor[HTML]{31869B}1&\cellcolor[HTML]{31869B}1&\cellcolor[HTML]{31869B}1&\cellcolor[HTML]{E26B0A}2&\cellcolor[HTML]{E26B0A}2&\cellcolor[HTML]{E26B0A}2&\cellcolor[HTML]{31869B}1&\cellcolor[HTML]{31869B}1&\cellcolor[HTML]{31869B}1&\cellcolor[HTML]{E26B0A}2&\cellcolor[HTML]{E26B0A}2&\cellcolor[HTML]{E26B0A}2&\cellcolor[HTML]{E26B0A}2&\cellcolor[HTML]{31869B}1&\cellcolor[HTML]{31869B}1&\cellcolor[HTML]{31869B}1&\cellcolor[HTML]{E26B0A}2&\cellcolor[HTML]{31869B}1\\

\bottomrule[1.2pt]
\end{tabu}
\vspace{-2mm}
\caption{\label{tab:18langs_group_gemma2b} The grouping results of $\text{Gemma}_{\text{2B}}$, where ``\textbf{2G}'' denotes the result that divides into 2 groups. 
}
\vspace{-2mm}
\end{table}

%% file: tabs/langs_128langs_16G.tex
\begin{table*}[htp]

\begin{minipage}[t]{0.49\linewidth}
       \setlength{\tabcolsep}{2mm}
	\centering
	\small
	\renewcommand\arraystretch{1.25}
	\begin{center}
		\begin{tabular}{cc}
			\toprule[1.2pt]  
               \multicolumn{1}{c}{\textbf{Index}} & \multicolumn{1}{c}{\textbf{Languages}} \\
                \midrule[0.8pt]
                    1&ceb, en, fil, hil, ilo, la, lg, so\\
                    2&fr, it, ny, sn, sw, xh, yo, zu\\
                    3&am, dv, he, ka, ko, lo, my, ti\\
                    4&ar, ca, es, eu, hi, id, pt, vi\\
                    5&cnh, cs, de, ha, kha, lus, nl, uz\\
                    6&kaa, kk, ky, mn, ru, sah, tt, tyv\\
                    7&da, et, fi, fo, gsw, is, no, se\\
                    8&av, be, ce, mk, sr, tg, udm, uk\\
			\bottomrule[1.2pt]
		\end{tabular}
	\end{center}
	\vspace{-5mm}
\end{minipage}
\begin{minipage}[t]{0.49\linewidth}
       \setlength{\tabcolsep}{2.5mm}
	\centering
	\small
	%\vspace{-0.4cm}
	\renewcommand\arraystretch{1.2}
	\begin{center}
		% \caption{Ablation study of different training methods on 5 datasets for $\text{BLOOM}_{\text{560M}}$.}
		\begin{tabular}{cc}
			\toprule[1.2pt]  
               \multicolumn{1}{c}{\textbf{Index}} & \multicolumn{1}{c}{\textbf{Languages}} \\
                \midrule[0.8pt]
                    9&el, grc, hu, os, pl, ro, tr, yi\\
                    10&bn, gu, kn, ml, mr, pa, ta, te\\
                    11&az, br, ckb, fa, ps, sd, ug, ur\\
                    12&ht, ig, jv, mg, ms, sl, su, vec\\
                    13&fy, haw, lv, mi, sm, st, tet, to\\
                    14&co, eo, gl, ja, ne, oc, yue, zh\\
                    15&ee, gd, hmn, lb, mt, om, rm, ts\\
                    16&bo, kbd, kl, km, pap, sa, th, tk\\
			\bottomrule[1.2pt]
		\end{tabular}
	\end{center}
	\vspace{-5mm}
\end{minipage}
\vspace{2mm}

\caption{\label{tab:128lang_16G} The 16 language groups divided for the 128 languages experiment. 
}

\vspace{-2mm}

\end{table*}

%% file: tabs/main_18langs_ppl_qwen.tex
\begin{table*}[htp]

\renewcommand\arraystretch{1.1}

\centering
\scriptsize

\setlength{\tabcolsep}{0.6mm}

 \begin{tabu}{lr|cccccccc|ccccccc|ccc|c}
 
 \toprule[1.2pt]
  
  \multicolumn{2}{c}{ } & \multicolumn{8}{c}{\textbf{High}} & \multicolumn{7}{c}{\textbf{Medium}} & \multicolumn{3}{c}{\textbf{Low}} & \multicolumn{1}{c}{ } \\

  \cmidrule(r){3-10} \cmidrule(r){11-17} \cmidrule(r){18-20} \noalign{\smallskip}

% high: ru, de, nl, ar, fr, it, ja, zh
% medium: vi, id, uk, hi, bn, ko, th
% low: ta, te, ur

 \multicolumn{1}{c}{\textbf{Model}} & \multicolumn{1}{c}{$\#$\textbf{Param.}} & \textbf{ar} & \textbf{de} & \textbf{fr} & \textbf{it} & \textbf{ja} & \textbf{nl} & \textbf{ru} & \textbf{zh} & \textbf{bn} & \textbf{hi} & \textbf{id} & \textbf{ko} & 
 \textbf{th} & \textbf{uk} & \textbf{vi} & \textbf{ta} & \textbf{te} &  
 \multicolumn{1}{c}{\textbf{ur}} & \multicolumn{1}{c}{\textbf{Avg }$\downarrow$} \\

\midrule[0.8pt]

$\text{Qwen2.5}_{\text{0.5B}}$
&494M&$110.8$&$147.3$&$9.5$&$133.8$&$19.8$&$20.2$&$\underline{3.7}$&$36.8$&$336.2$&$191.4$&$85.2$&$2.5$&$3.4$&$5.4$&$23.3$&$44.7$&$2.1e^5$&$689.1$&$1262.0$\\

$\text{\tiny \ \ \ + Pre-train\ }$
&494M&$75.7$&$89.2$&$8.3$&$75.3$&$\underline{15.9}$&$10.9$&$\textbf{3.4}$&$32.4$&$\underline{40.6}$&$\underline{32.8}$&$56.9$&$\underline{2.4}$&$\underline{2.8}$&$\underline{3.2}$&$18.7$&$\underline{9.0}$&$\underline{192.5}$&$\underline{48.8}$&$\underline{39.9}$\\

   \cdashlinelr{1-21}
$\text{X-ELM}$
&4.44B&$\underline{74.3}$&$\underline{87.9}$&$\underline{8.2}$&$\underline{72.9}$&$\textbf{15.3}$&$\underline{10.7}$&$\textbf{3.4}$&$\textbf{30.3}$&$51.9$&$40.2$&$\underline{55.5}$&$\textbf{2.3}$&$\underline{2.8}$&$\underline{3.2}$&$\underline{18.2}$&$11.9$&$404.1$&$57.3$&$52.8$\\

$\text{DMoE (9 Groups)}$
&1.75B&$\textbf{68.8}$&$\textbf{81.1}$&$\textbf{8.0}$&$\textbf{67.9}$&$\textbf{15.3}$&$\textbf{9.8}$&$\textbf{3.4}$&$\underline{31.6}$&$\textbf{33.6}$&$\textbf{27.8}$&$\textbf{51.8}$&$\textbf{2.3}$&$\textbf{2.7}$&$\textbf{3.0}$&$\textbf{17.7}$&$\textbf{8.6}$&$\textbf{173.0}$&$\textbf{38.7}$&$\textbf{35.8}$\\

\midrule[0.8pt]

$\text{Qwen}_{\text{1.5B}}$
&1.54B&$59.4$&$70.8$&$\underline{7.0}$&$64.4$&$13.2$&$12.2$&$\underline{3.0}$&$25.1$&$136.2$&$86.4$&$45.0$&$\underline{2.2}$&$\underline{2.8}$&$3.9$&$15.0$&$29.6$&$4809.7$&$246.9$&$312.9$\\

$\text{\tiny \ \ \ + Pre-train\ }$
&1.54B&$48.6$&$52.7$&$\textbf{6.6}$&$47.1$&$11.9$&$\underline{8.3}$&$\underline{3.0}$&$22.5$&$\underline{26.9}$&$\underline{22.6}$&$38.0$&$\textbf{2.1}$&$\textbf{2.5}$&$\underline{2.8}$&$13.6$&$\underline{7.4}$&$\underline{121.0}$&$\underline{32.7}$&$\underline{26.1}$\\

   \cdashlinelr{1-21}
$\text{X-ELM}$
&13.89B&$\textbf{47.4}$&$\underline{52.1}$&$\textbf{6.6}$&$\textbf{45.0}$&$\textbf{11.5}$&$\textbf{8.1}$&$\textbf{2.9}$&$\textbf{21.6}$&$31.3$&$25.9$&$\textbf{36.4}$&$\textbf{2.1}$&$\textbf{2.5}$&$\underline{2.8}$&$\textbf{13.2}$&$8.7$&$174.2$&$35.6$&$29.3$\\

   % \cdashlinelr{1-16}

$\text{DMoE (9 Groups)}$
&3.19B&$\underline{47.9}$&$\textbf{51.7}$&$\textbf{6.6}$&$\underline{45.9}$&$\underline{11.8}$&$\textbf{8.1}$&$\underline{3.0}$&$\underline{22.4}$&$\textbf{25.6}$&$\textbf{21.6}$&$\underline{37.4}$&$\textbf{2.1}$&$\textbf{2.5}$&$\textbf{2.7}$&$\underline{13.5}$&$\textbf{7.2}$&$\textbf{103.4}$&$\textbf{30.4}$&$\textbf{24.7}$\\

\bottomrule[1.2pt]
\end{tabu}
\vspace{-2mm}
\caption{\label{tab:18langs_ppl_qwen} The normalized perplexity on the valid split of CulturaX. The perplexity is normalized to the vocabulary of Bloom following \citet{wei2023skywork}. ``\textbf{High}'', ``\textbf{Medium}'', and ``\textbf{Low}'' indicates the available amount of linguistic resources. The best and second results are denoted in \textbf{bold} and \underline{underlined}, correspondingly.
}
\vspace{-2mm}
\end{table*}

%% file: tabs/128langs/part1.tex
\begin{table*}[htp]

\renewcommand\arraystretch{1.5}

\centering
\scriptsize

\setlength{\tabcolsep}{0.8mm}

 \begin{tabu}{l|rrrrrrrrrrrrrrrr}
 
 \toprule[1.2pt]
  
% part1: am$&$ar$&$av$&$az$&$be$&$bn$&$bo$&$br$&$ca$&$ce$&$ceb$&$ckb$&$cnh$&$co$&$cs$&$da

 \multicolumn{1}{c}{\textbf{Model}} & \textbf{am} & \textbf{ar} & \textbf{av} & \textbf{az} & \textbf{be} & \textbf{bn} & \textbf{bo} & \textbf{br} & \textbf{ca} & \textbf{ce} & \textbf{ceb} & \textbf{ckb} & \textbf{cnh} & \textbf{co} & \textbf{cs} & \textbf{da} \\

\midrule[0.8pt]

$\text{BLOOM}_{\text{560M}}$
&$13.84 $&$42.39 $&$48.53 $&$187.51 $&$65.85 $&$45.23 $&$7.17 $&$261.68 $&$41.64 $&$51.08 $&$225.93 $&$35.40 $&$445.37 $&$197.32 $&$146.79 $&$84.96 $\\

$\text{\ + Pre-train\ }$
&$4.39 $&$34.43 $&$8.27 $&$11.68 $&$8.63 $&$33.11 $&$3.44 $&$40.25 $&$24.85 $&$11.25 $&$20.18 $&$8.97 $&$32.45 $&$32.27 $&$17.03 $&$10.04 $\\

   \cdashlinelr{1-17}

$\text{Branch-Train-Mix}$
&$4.12 $&$32.53 $&$7.53 $&$10.85 $&$7.61 $&$32.62 $&$3.47 $&$35.44 $&$23.59 $&$9.62 $&$18.37 $&$8.46 $&$28.38 $&$28.55 $&$15.43 $&$8.67 $\\

$\text{DMoE}$
&$4.08 $&$32.06 $&$7.18 $&$9.82 $&$7.38 $&$31.69 $&$3.39 $&$31.42 $&$23.61 $&$9.39 $&$16.57 $&$8.07 $&$27.59 $&$25.97 $&$14.30 $&$7.98 $\\

\midrule[0.8pt]

$\text{BLOOM}_{\text{1.7B}}$
&$8.60 $&$30.41 $&$32.15 $&$72.69 $&$36.83 $&$28.67 $&$5.58 $&$113.16 $&$26.55 $&$34.38 $&$93.94 $&$24.93 $&$188.45 $&$103.17 $&$73.24 $&$42.47 $\\

$\text{\ + Pre-train\ }$
&$3.74 $&$22.93 $&$6.40 $&$8.68 $&$6.40 $&$22.04 $&$3.15 $&$27.93 $&$18.00 $&$8.30 $&$15.50 $&$7.51 $&$24.43 $&$22.78 $&$12.06 $&$6.97 $\\

   \cdashlinelr{1-17}
   
$\text{Branch-Train-Mix}$
&$3.71 $&$21.10 $&$6.25 $&$8.43 $&$6.16 $&$20.82 $&$3.27 $&$26.20 $&$17.12 $&$7.96 $&$14.83 $&$7.22 $&$23.23 $&$22.41 $&$11.82 $&$6.51 $\\

$\text{DMoE}$
&$3.62 $&$22.66 $&$5.94 $&$7.88 $&$5.94 $&$21.89 $&$3.12 $&$23.89 $&$17.88 $&$7.69 $&$13.80 $&$7.07 $&$21.90 $&$20.34 $&$11.15 $&$6.18 $\\

\bottomrule[1.2pt]
\end{tabu}
\vspace{-2mm}
\caption{\label{tab:128_ppl_p1} The perplexity of language ``am'' to ``da'' on the valid split of MADLAD-400 \citep{kudugunta2023madlad}.  
}
\vspace{-2mm}
\end{table*}

%% file: tabs/128langs/part2.tex
\begin{table*}[htp]

\renewcommand\arraystretch{1.5}

\centering
\scriptsize

\setlength{\tabcolsep}{0.7mm}

 \begin{tabu}{l|rrrrrrrrrrrrrrrr}
 
 \toprule[1.2pt]
  
% part2: de$&$dv$&$ee$&$el$&$en$&$eo$&$es$&$et$&$eu$&$fa$&$fi$&$fil$&$fo$&$fr$&$fy$&$gd

 \multicolumn{1}{c}{\textbf{Model}} & \textbf{de} & \textbf{dv} & \textbf{ee} & \textbf{el} & \textbf{en} & \textbf{eo} & \textbf{es} & \textbf{et} & \textbf{eu} & \textbf{fa} & \textbf{fi} & \textbf{fil} & \textbf{fo} & \textbf{fr} & \textbf{fy} & \textbf{gd} \\

\midrule[0.8pt]

$\text{BLOOM}_{\text{560M}}$
&$111.29 $&$4.96 $&$300.78 $&$18.28 $&$66.79 $&$266.55 $&$39.23 $&$310.09 $&$80.16 $&$161.83 $&$250.93 $&$234.67 $&$253.93 $&$43.25 $&$285.93 $&$172.37$\\

$\text{\ + Pre-train\ }$
&$23.83 $&$2.10 $&$20.79 $&$4.34 $&$44.58 $&$31.29 $&$24.03 $&$29.32 $&$39.50 $&$14.98 $&$24.25 $&$17.72 $&$24.34 $&$27.21 $&$24.39 $&$15.01 $\\

   \cdashlinelr{1-17}

$\text{Branch-Train-Mix}$
&$21.03 $&$2.11 $&$19.75 $&$4.18 $&$40.29 $&$26.97 $&$23.67 $&$24.47 $&$37.73 $&$14.51 $&$21.26 $&$15.86 $&$20.78 $&$25.53 $&$20.67 $&$12.95 $\\

$\text{DMoE}$
&$19.46 $&$2.05 $&$15.78 $&$4.02 $&$40.82 $&$22.84 $&$23.14 $&$22.67 $&$36.67 $&$13.40 $&$18.91 $&$14.18 $&$19.00 $&$25.97 $&$18.13 $&$11.55 $\\

\midrule[0.8pt]

$\text{BLOOM}_{\text{1.7B}}$
&$56.36 $&$3.49 $&$121.31 $&$11.27 $&$45.33 $&$126.18 $&$27.31 $&$151.08 $&$40.32 $&$86.79 $&$113.01 $&$98.05 $&$111.79 $&$28.28 $&$143.77 $&$84.11 $\\

$\text{\ + Pre-train\ }$
&$15.89 $&$1.97 $&$15.37 $&$3.53 $&$30.27 $&$21.05 $&$18.15 $&$19.85 $&$25.84 $&$11.96 $&$16.76 $&$13.23 $&$16.88 $&$20.03 $&$16.03 $&$11.38 $\\

   \cdashlinelr{1-17}
   
$\text{Branch-Train-Mix}$
&$15.34 $&$2.00 $&$14.43 $&$3.57 $&$29.69 $&$20.33 $&$17.43 $&$18.64 $&$24.97 $&$11.35 $&$16.27 $&$12.54 $&$15.78 $&$18.89 $&$14.92 $&$10.30 $\\

$\text{DMoE}$
&$14.86 $&$1.94 $&$13.35 $&$3.43 $&$31.34 $&$17.73 $&$18.19 $&$17.50 $&$25.37 $&$10.89 $&$15.07 $&$11.62 $&$14.72 $&$20.01 $&$13.55 $&$9.85 $\\

\bottomrule[1.2pt]
\end{tabu}
\vspace{-2mm}
\caption{\label{tab:128_ppl_p2} The perplexity of language ``de'' to ``gd'' on the valid split of MADLAD-400 \citep{kudugunta2023madlad}.  
}
\vspace{-2mm}
\end{table*}

%% file: tabs/128langs/part3.tex
\begin{table*}[htp]

\renewcommand\arraystretch{1.5}

\centering
\scriptsize

\setlength{\tabcolsep}{0.6mm}

 \begin{tabu}{l|rrrrrrrrrrrrrrrr}
 
 \toprule[1.2pt]
  
% part3: gl$&$grc$&$gsw$&$gu$&$ha$&$haw$&$he$&$hi$&$hil$&$hmn$&$ht$&$hu$&$id$&$ig$&$ilo$&$is

 \multicolumn{1}{c}{\textbf{Model}} & \textbf{gl} & \textbf{grc} & \textbf{gsw} & \textbf{gu} & \textbf{ha} & \textbf{haw} & \textbf{he} & \textbf{hi} & \textbf{hil} & \textbf{hmn} & \textbf{ht} & \textbf{hu} & \textbf{id} & \textbf{ig} & \textbf{ilo} & \textbf{is} \\

\midrule[0.8pt]

$\text{BLOOM}_{\text{560M}}$
&$121.12$&$19.74$&$143.45$&$180.03$&$559.77$&$73.99$&$20.93$&$30.55$&$220.05$&$120.38$&$391.80$&$156.40$&$41.35$&$262.93$&$271.81$&$207.29 $\\

$\text{\ + Pre-train\ }$
&$26.66$&$6.00$&$42.16$&$39.27$&$27.09$&$10.39$&$7.01$&$20.45$&$17.53$&$13.11$&$33.86$&$15.41$&$26.93$&$28.80$&$26.04$&$20.08 $\\

   \cdashlinelr{1-17}

$\text{Branch-Train-Mix}$
&$24.84$&$5.69$&$37.22$&$40.53$&$24.27$&$9.67$&$6.71$&$19.98$&$14.53$&$11.94$&$30.05$&$14.10$&$26.10$&$26.85$&$22.45$&$17.36 $\\

$\text{DMoE}$
&$23.70$&$5.45$&$35.14$&$37.51$&$22.80$&$9.00$&$6.44$&$19.78$&$13.22$&$10.78$&$26.19$&$13.00$&$25.81$&$23.77$&$19.76$&$15.98 $\\

\midrule[0.8pt]

$\text{BLOOM}_{\text{1.7B}}$
&$61.19 $&$12.99 $&$90.97 $&$67.49 $&$191.64 $&$47.25 $&$15.00 $&$20.82 $&$96.72 $&$79.33 $&$169.97 $&$70.31 $&$27.80 $&$92.00 $&$130.53 $&$92.95 $\\

$\text{\ + Pre-train\ }$
&$18.81 $&$4.71 $&$28.90 $&$27.69 $&$19.49 $&$8.48 $&$5.58 $&$14.70 $&$12.23 $&$10.47 $&$23.94 $&$10.98 $&$19.25 $&$20.90 $&$18.45 $&$14.12 $\\

   \cdashlinelr{1-17}
   
$\text{Branch-Train-Mix}$
&$18.04 $&$4.74 $&$27.91 $&$27.50 $&$18.94 $&$8.24 $&$5.55 $&$14.14 $&$10.97 $&$9.79 $&$21.31 $&$10.91 $&$18.65 $&$19.81 $&$17.42 $&$13.36 $\\

$\text{DMoE}$
&$17.84 $&$4.60 $&$26.20 $&$27.18 $&$17.73 $&$7.91 $&$5.41 $&$14.68 $&$10.44 $&$9.48 $&$20.24 $&$10.27 $&$19.08 $&$18.63 $&$15.59 $&$12.44 $\\

\bottomrule[1.2pt]
\end{tabu}
\vspace{-2mm}
\caption{\label{tab:128_ppl_p3} The perplexity of language ``gl'' to ``is'' on the valid split of MADLAD-400 \citep{kudugunta2023madlad}.  
}
\vspace{-2mm}
\end{table*}

%% file: tabs/128langs/part4.tex
\begin{table*}[htp]

\renewcommand\arraystretch{1.5}

\centering
\scriptsize

\setlength{\tabcolsep}{0.8mm}

 \begin{tabu}{l|rrrrrrrrrrrrrrrr}
 
 \toprule[1.2pt]
  
% part4: it$&$ja$&$jv$&$ka$&$kaa$&$kbd$&$kha$&$kk$&$kl$&$km$&$kn$&$ko$&$ky$&$la$&$lb$&$lg

 \multicolumn{1}{c}{\textbf{Model}} & \textbf{it} & \textbf{ja} & \textbf{jv} & \textbf{ka} & \textbf{kaa} & \textbf{kbd} & \textbf{kha} & \textbf{kk} & \textbf{kl} & \textbf{km} & \textbf{kn} & \textbf{ko} & \textbf{ky} & \textbf{la} & \textbf{lb} & \textbf{lg} \\

\midrule[0.8pt]

$\text{BLOOM}_{\text{560M}}$
&$82.38$&$55.55$&$277.50$&$12.20$&$50.53$&$30.80$&$200.91$&$28.98$&$231.11$&$10.09$&$196.03$&$24.03$&$47.03$&$111.79$&$261.41$&$369.16 $\\

$\text{\ + Pre-train\ }$
&$20.47$&$12.86$&$36.26$&$4.16$&$6.56$&$7.68$&$20.35$&$5.71$&$16.50$&$3.85$&$46.08$&$7.36$&$7.37$&$36.19$&$25.71$&$46.19 $\\

   \cdashlinelr{1-17}

$\text{Branch-Train-Mix}$
&$17.12$&$12.12$&$34.82$&$4.08$&$6.16$&$7.12$&$18.22$&$5.38$&$14.78$&$3.81$&$47.55$&$7.12$&$6.89$&$34.72$&$21.35$&$40.71 $\\

$\text{DMoE}$
&$16.33$&$11.34$&$30.43$&$3.89$&$5.95$&$6.35$&$17.05$&$5.20$&$13.57$&$3.63$&$43.80$&$6.74$&$6.61$&$32.04$&$18.92$&$36.13 $\\

\midrule[0.8pt]

$\text{BLOOM}_{\text{1.7B}}$
&$44.86 $&$35.01 $&$149.22 $&$10.42 $&$32.12 $&$20.19 $&$124.88 $&$19.14 $&$120.42 $&$8.26 $&$92.29 $&$16.12 $&$31.49 $&$66.75 $&$132.24 $&$141.18 $\\

$\text{\ + Pre-train\ }$
&$14.13 $&$9.67 $&$26.62 $&$3.63 $&$5.12 $&$5.86 $&$14.83 $&$4.52 $&$12.68 $&$3.36 $&$32.93 $&$6.00 $&$5.73 $&$27.12 $&$16.92 $&$33.36 $\\

   \cdashlinelr{1-17}
   
$\text{Branch-Train-Mix}$
&$13.10 $&$9.39 $&$25.49 $&$3.66 $&$5.14 $&$5.83 $&$14.51 $&$4.50 $&$12.57 $&$3.39 $&$32.53 $&$5.96 $&$5.73 $&$27.40 $&$15.66 $&$31.24 $\\

$\text{DMoE}$
&$12.83 $&$8.92 $&$23.54 $&$3.51 $&$4.86 $&$5.50 $&$13.19 $&$4.31 $&$11.14 $&$3.25 $&$32.03 $&$5.72 $&$5.44 $&$26.05 $&$14.69 $&$28.27 $\\

\bottomrule[1.2pt]
\end{tabu}
\vspace{-2mm}
\caption{\label{tab:128_ppl_p4} The perplexity of language ``it'' to ``lg'' on the valid split of MADLAD-400 \citep{kudugunta2023madlad}.  
}
\vspace{-2mm}
\end{table*}

%% file: tabs/128langs/part5.tex
\begin{table*}[htp]

\renewcommand\arraystretch{1.5}

\centering
\scriptsize

\setlength{\tabcolsep}{0.6mm}

 \begin{tabu}{l|rrrrrrrrrrrrrrrr}
 
 \toprule[1.2pt]
  
% part5: lo$&$lus$&$lv$&$mg$&$mi$&$mk$&$ml$&$mn$&$mr$&$ms$&$mt$&$my$&$ne$&$nl$&$no$&$ny

 \multicolumn{1}{c}{\textbf{Model}} & \textbf{lo} & \textbf{lus} & \textbf{lv} & \textbf{mg} & \textbf{mi} & \textbf{mk} & \textbf{ml} & \textbf{mn} & \textbf{mr} & \textbf{ms} & \textbf{mt} & \textbf{my} & \textbf{ne} & \textbf{nl} & \textbf{no} & \textbf{ny} \\

\midrule[0.8pt]

$\text{BLOOM}_{\text{560M}}$
&$8.72$&$312.58$&$169.08$&$189.81$&$129.33$&$62.40$&$116.36$&$31.15$&$127.22$&$98.74$&$97.91$&$6.08$&$125.47$&$118.66$&$213.49$&$286.37 $\\

$\text{\ + Pre-train\ }$
&$2.44$&$37.91$&$15.10$&$18.63$&$16.50$&$9.37$&$34.71$&$6.75$&$36.21$&$33.69$&$9.95$&$2.77$&$51.30$&$15.22$&$22.27$&$22.99 $\\

   \cdashlinelr{1-17}

$\text{Branch-Train-Mix}$
&$2.42$&$34.92$&$12.49$&$16.38$&$15.14$&$8.45$&$35.88$&$6.31$&$36.48$&$32.98$&$8.94$&$2.77$&$53.62$&$13.01$&$19.45$&$20.69 $\\

$\text{DMoE}$
&$2.37$&$32.79$&$10.65$&$15.08$&$13.82$&$8.05$&$33.29$&$6.19$&$34.98$&$31.17$&$7.42$&$2.69$&$49.34$&$12.22$&$17.92$&$18.37 $\\

\midrule[0.8pt]

$\text{BLOOM}_{\text{1.7B}}$
&$4.58 $&$159.69 $&$83.05 $&$82.55 $&$74.40 $&$36.15 $&$50.08 $&$21.05 $&$61.00 $&$51.52 $&$59.86 $&$4.41 $&$63.23 $&$58.05 $&$97.54 $&$126.71 $\\

$\text{\ + Pre-train\ }$
&$2.17 $&$28.49 $&$10.26 $&$14.08 $&$13.02 $&$6.96 $&$23.58 $&$5.37 $&$24.57 $&$23.49 $&$7.26 $&$2.52 $&$35.24 $&$10.26 $&$15.45 $&$17.08 $\\

   \cdashlinelr{1-17}
   
$\text{Branch-Train-Mix}$
&$2.20 $&$28.23 $&$9.05 $&$12.81 $&$12.46 $&$6.74 $&$23.43 $&$5.35 $&$24.15 $&$22.88 $&$6.67 $&$2.57 $&$35.40 $&$9.67 $&$14.52 $&$15.98 $\\

$\text{DMoE}$
&$2.13 $&$26.04 $&$8.45 $&$12.54 $&$11.84 $&$6.51 $&$23.58 $&$5.13 $&$24.06 $&$22.18 $&$6.27 $&$2.48 $&$34.91 $&$9.37 $&$13.86 $&$15.11 $\\

\bottomrule[1.2pt]
\end{tabu}
\vspace{-2mm}
\caption{\label{tab:128_ppl_p5} The perplexity of language ``lo'' to ``ny'' on the valid split of MADLAD-400 \citep{kudugunta2023madlad}.  
}
\vspace{-2mm}
\end{table*}

%% file: tabs/128langs/part6.tex
\begin{table*}[htp]

\renewcommand\arraystretch{1.5}

\centering
\scriptsize

\setlength{\tabcolsep}{0.6mm}

 \begin{tabu}{l|rrrrrrrrrrrrrrrr}
 
 \toprule[1.2pt]
  
% part6:

 \multicolumn{1}{c}{\textbf{Model}} & \textbf{oc} & \textbf{om} & \textbf{os} & \textbf{pa} & \textbf{pap} & \textbf{pl} & \textbf{ps} & \textbf{pt} & \textbf{rm} & \textbf{ro} & \textbf{ru} & \textbf{sa} & \textbf{sah} & \textbf{sd} & \textbf{se} & \textbf{sl}\\

\midrule[0.8pt]

$\text{BLOOM}_{\text{560M}}$
&$80.82$&$260.13$&$30.78$&$131.60$&$367.89$&$91.73$&$67.38$&$37.27$&$354.95$&$195.23$&$33.77$&$181.38$&$32.12$&$83.51$&$255.36$&$218.52 $\\

$\text{\ + Pre-train\ }$
&$20.72$&$27.23$&$8.08$&$32.55$&$28.52$&$12.84$&$10.99$&$23.22$&$27.98$&$16.72$&$8.13$&$43.64$&$6.88$&$13.97$&$27.01$&$22.97 $\\

   \cdashlinelr{1-17}

$\text{Branch-Train-Mix}$
&$19.13$&$23.77$&$7.56$&$34.14$&$25.53$&$11.91$&$10.65$&$22.23$&$23.60$&$15.09$&$7.40$&$47.25$&$6.51$&$14.48$&$23.05$&$19.99 $\\

$\text{DMoE}$
&$17.82$&$21.06$&$7.06$&$30.48$&$22.65$&$11.12$&$9.59$&$22.51$&$18.69$&$14.06$&$7.24$&$41.47$&$6.20$&$12.62$&$21.32$&$17.32 $\\

\midrule[0.8pt]

$\text{BLOOM}_{\text{1.7B}}$
&$45.64 $&$150.00 $&$21.03 $&$61.74 $&$187.88 $&$46.24 $&$44.31 $&$24.85 $&$183.77 $&$87.28 $&$21.70 $&$104.87 $&$22.39 $&$54.03 $&$156.62 $&$107.83 $\\

$\text{\ + Pre-train\ }$
&$14.69 $&$20.86 $&$6.20 $&$23.89 $&$19.32 $&$9.20 $&$8.68 $&$17.07 $&$18.42 $&$11.82 $&$6.16 $&$34.04 $&$5.45 $&$11.45 $&$18.82 $&$15.75 $\\

   \cdashlinelr{1-17}
   
$\text{Branch-Train-Mix}$
&$14.14 $&$19.52 $&$6.36 $&$24.24 $&$17.96 $&$9.25 $&$8.46 $&$16.35 $&$15.92 $&$11.42 $&$5.93 $&$36.76 $&$5.52 $&$11.82 $&$17.97 $&$14.49 $\\

$\text{DMoE}$
&$13.36 $&$17.80 $&$5.68 $&$23.17 $&$16.37 $&$8.66 $&$7.95 $&$17.17 $&$14.33 $&$10.93 $&$5.82 $&$32.97 $&$5.18 $&$10.71 $&$16.31 $&$13.48 $\\

\bottomrule[1.2pt]
\end{tabu}
\vspace{-2mm}
\caption{\label{tab:128_ppl_p6} The perplexity of language ``oc'' to ``sl'' on the valid split of MADLAD-400 \citep{kudugunta2023madlad}.  
}
\vspace{-2mm}
\end{table*}

%% file: tabs/128langs/part7.tex
\begin{table*}[htp]

\renewcommand\arraystretch{1.5}

\centering
\scriptsize

\setlength{\tabcolsep}{0.6mm}

 \begin{tabu}{l|rrrrrrrrrrrrrrrr}
 
 \toprule[1.2pt]
  
% part7:

 \multicolumn{1}{c}{\textbf{Model}} & \textbf{sm} & \textbf{sn} & \textbf{so} & \textbf{sr} & \textbf{st} & \textbf{su} & \textbf{sw} & \textbf{ta} & \textbf{te} & \textbf{tet} & \textbf{tg} & \textbf{th} & \textbf{ti} & \textbf{tk} & \textbf{to} & \textbf{tr}\\

\midrule[0.8pt]

$\text{BLOOM}_{\text{560M}}$
&$114.75$&$435.89$&$239.57$&$51.57$&$275.07$&$224.84$&$224.92$&$80.03$&$91.45$&$206.03$&$40.17$&$13.71$&$16.22$&$217.70$&$82.62$&$153.97 $\\

$\text{\ + Pre-train\ }$
&$15.29$&$28.67$&$18.92$&$9.57$&$18.37$&$27.30$&$45.46$&$37.65$&$30.32$&$16.00$&$7.22$&$3.64$&$5.13$&$14.84$&$14.60$&$14.62 $\\

   \cdashlinelr{1-17}

$\text{Branch-Train-Mix}$
&$13.78$&$25.29$&$16.27$&$8.58$&$16.60$&$25.26$&$43.52$&$38.29$&$31.56$&$14.34$&$6.62$&$3.49$&$4.83$&$13.70$&$12.54$&$13.49 $\\

$\text{DMoE}$
&$12.63$&$22.63$&$14.53$&$8.07$&$14.96$&$22.70$&$39.65$&$36.32$&$29.26$&$12.48$&$6.36$&$3.43$&$4.70$&$12.08$&$11.60$&$12.80 $\\

\midrule[0.8pt]

$\text{BLOOM}_{\text{1.7B}}$
&$75.21 $&$143.65 $&$131.67 $&$31.61 $&$115.55 $&$129.35 $&$80.34 $&$47.14 $&$46.77 $&$100.13 $&$25.36 $&$9.70 $&$10.17 $&$87.48 $&$49.50 $&$61.45 $\\

$\text{\ + Pre-train\ }$
&$12.08 $&$20.74 $&$13.86 $&$7.08 $&$13.90 $&$20.01 $&$30.75 $&$26.90 $&$22.30 $&$11.61 $&$5.56 $&$3.12 $&$4.31 $&$10.63 $&$11.29 $&$10.42 $\\

   \cdashlinelr{1-17}
   
$\text{Branch-Train-Mix}$
&$11.27 $&$19.18 $&$12.80 $&$6.88 $&$13.11 $&$18.71 $&$29.84 $&$26.40 $&$22.36 $&$10.45 $&$5.49 $&$3.09 $&$4.31 $&$10.80 $&$10.27 $&$10.20 $\\

$\text{DMoE}$
&$10.91 $&$18.48 $&$11.95 $&$6.52 $&$12.52 $&$17.68 $&$29.23 $&$26.61 $&$22.14 $&$9.76 $&$5.24 $&$3.03 $&$4.11 $&$9.39 $&$9.81 $&$9.75 $\\

\bottomrule[1.2pt]
\end{tabu}
\vspace{-2mm}
\caption{\label{tab:128_ppl_p7} The perplexity of language ``sm'' to ``tr'' on the valid split of MADLAD-400 \citep{kudugunta2023madlad}.  
}
\vspace{-2mm}
\end{table*}

%% file: tabs/128langs/part8.tex
\begin{table*}[htp]

\renewcommand\arraystretch{1.5}

\centering
\scriptsize

\setlength{\tabcolsep}{0.6mm}

 \begin{tabu}{l|rrrrrrrrrrrrrrrr}
 
 \toprule[1.2pt]
  
% part8:

 \multicolumn{1}{c}{\textbf{Model}} & \textbf{ts} & \textbf{tt} & \textbf{tyv} & \textbf{udm} & \textbf{ug} & \textbf{uk} & \textbf{ur} & \textbf{uz} & \textbf{vec} & \textbf{vi} & \textbf{xh} & \textbf{yi} & \textbf{yo} & \textbf{yue} & \textbf{zh} & \textbf{zu}\\

\midrule[0.8pt]

$\text{BLOOM}_{\text{560M}}$
&$175.78$&$39.70$&$40.14$&$54.51$&$44.53$&$44.49$&$72.47$&$388.73$&$319.37$&$21.85$&$606.11$&$17.27$&$257.05$&$71.22$&$36.26$&$1278.88 $\\

$\text{\ + Pre-train\ }$
&$13.05$&$7.46$&$8.37$&$8.87$&$9.21$&$7.30$&$37.61$&$18.66$&$73.68$&$15.12$&$34.44$&$5.10$&$28.08$&$25.72$&$15.07$&$36.17 $\\

   \cdashlinelr{1-17}

$\text{Branch-Train-Mix}$
&$11.26$&$6.97$&$7.54$&$7.95$&$8.99$&$6.61$&$37.13$&$16.74$&$64.42$&$14.62$&$32.00$&$4.86$&$27.41$&$25.82$&$14.83$&$31.24 $\\

$\text{DMoE}$
&$9.90$&$6.71$&$7.33$&$7.65$&$8.10$&$6.40$&$35.19$&$15.80$&$55.43$&$14.66$&$27.73$&$4.71$&$24.28$&$23.99$&$14.31$&$28.31 $\\

\midrule[0.8pt]

$\text{BLOOM}_{\text{1.7B}}$
&$85.50 $&$27.57 $&$28.15 $&$39.09 $&$28.37 $&$26.68 $&$44.00 $&$144.46 $&$188.36 $&$15.30 $&$154.98 $&$11.37 $&$100.09 $&$42.71 $&$25.86 $&$260.80 $\\

$\text{\ + Pre-train\ }$
&$9.51 $&$5.77 $&$6.38 $&$6.75 $&$7.35 $&$5.44 $&$26.84 $&$13.32 $&$46.42 $&$11.34 $&$25.61 $&$4.13 $&$21.09 $&$19.02 $&$11.70 $&$26.00 $\\

   \cdashlinelr{1-17}
   
$\text{Branch-Train-Mix}$
&$8.67 $&$5.74 $&$6.20 $&$6.54 $&$7.47 $&$5.28 $&$25.68 $&$13.08 $&$43.24 $&$10.88 $&$24.70 $&$4.17 $&$21.20 $&$17.44 $&$11.33 $&$24.14 $\\

$\text{DMoE}$
&$8.10 $&$5.45 $&$5.86 $&$6.30 $&$6.81 $&$5.15 $&$25.96 $&$12.12 $&$38.54 $&$11.27 $&$22.65 $&$3.96 $&$19.29 $&$18.25 $&$11.30 $&$22.82 $\\

\bottomrule[1.2pt]
\end{tabu}
\vspace{-2mm}
\caption{\label{tab:128_ppl_p8} The perplexity of language ``ts'' to ``zu'' on the valid split of MADLAD-400 \citep{kudugunta2023madlad}.  
}
\vspace{-2mm}
\end{table*}

%% file: tabs/in-context/xnli.tex
\begin{table*}[thp]

\renewcommand\arraystretch{1.3}

\centering
\scriptsize

\setlength{\tabcolsep}{1mm}

 \begin{tabu}{l|c|ccccccc|cccccc|ccc|c}
 
 \toprule[1.2pt]
  \multicolumn{2}{c}{ } & \multicolumn{7}{c}{\textbf{High}} & \multicolumn{6}{c}{\textbf{Medium}} & \multicolumn{3}{c}{\textbf{Low}} &\\
  \cmidrule(r){3-9} \cmidrule(r){10-15} \cmidrule(r){16-18} \noalign{\smallskip}
\multicolumn{1}{c}{\textbf{Model}}& \multicolumn{1}{c}{\textbf{\#shot}} & en & de${}^{\dagger}$ & es & eu & fr & ru${}^{\dagger}$  & zh & ar & bg${}^{\dagger}$  & el${}^{\dagger}$  & th${}^{\dagger}$  & tr${}^{\dagger}$  & vi & hi & sw & ur &\multicolumn{1}{c}{\textbf{Avg}}\\

   \midrule[0.8pt]
\multirow{2}{*}{$\text{BLOOM}_{\text{560M}}$}
            & 0 & $43.9$&$34.4$&$40.5$&$37.9$&$39.1$&$34.6$&$35.5$&$33.5$&$34.0$&$35.2$&$32.1$&$31.6$&$39.0$&$39.8$&$33.9$&$34.5$&$36.2$\\

            & 4 & $40.3$&$34.0$&$38.6$&$35.0$&$37.0$&$34.4$&$32.4$&$33.3$&$33.4$&$31.8$&$33.5$&$32.1$&$36.0$&$33.7$&$31.8$&$34.1$&$34.4$\\

    \cdashlinelr{1-19}

\multirow{2}{*}{$\text{BLOOM}_{\text{560M}}$ + Pre-train}
            & 0 & $43.5$&$37.6$&$41.6$&$40.6$&$40.9$&$36.1$&$33.4$&$33.5$&$34.1$&$33.1$&$33.8$&$33.4$&$42.0$&$38.5$&$33.9$&$37.1$&$37.1$\\

            & 4 & $40.3$&$33.9$&$38.5$&$36.7$&$38.4$&$32.7$&$35.5$&$33.4$&$31.6$&$33.4$&$31.8$&$32.8$&$36.3$&$35.3$&$32.2$&$32.1$&$34.7$\\

    \cdashlinelr{1-19}

\multirow{2}{*}{$\text{Branch-Train-Mix}$}
            & 0 & $47.4$&$36.0$&$41.9$&$41.2$&$41.1$&$36.7$&$33.8$&$33.5$&$34.1$&$34.9$&$32.6$&$34.1$&$41.7$&$35.5$&$33.5$&$36.7$&$37.2$\\

            & 4 & $41.6$&$35.8$&$39.8$&$35.0$&$38.6$&$36.1$&$34.9$&$32.7$&$34.2$&$33.1$&$34.3$&$33.1$&$35.8$&$33.9$&$32.0$&$33.9$&$35.3$\\

    \cdashlinelr{1-19}

\multirow{2}{*}{$\text{DMoE}$}
            & 0 & $48.3$&$36.3$&$43.8$&$38.6$&$42.9$&$37.3$&$33.4$&$33.3$&$34.5$&$34.1$&$34.0$&$33.9$&$43.0$&$37.7$&$33.2$&$36.2$&$\textbf{37.5}$\\

            & 4 & $41.2$&$35.2$&$40.5$&$35.2$&$39.0$&$35.8$&$35.2$&$33.9$&$35.4$&$33.9$&$33.3$&$33.6$&$37.7$&$35.1$&$32.1$&$34.0$&$\textbf{35.7}$\\

  \midrule[0.8pt]
\multirow{2}{*}{$\text{BLOOM}_{\text{1.7B}}$}
            & 0 & $49.2$&$36.6$&$47.7$&$47.0$&$45.2$&$37.8$&$34.9$&$33.3$&$35.6$&$33.6$&$33.7$&$35.3$&$42.9$&$42.2$&$34.1$&$38.4$&$39.2$\\

            & 4 & $46.3$&$34.7$&$43.1$&$40.9$&$45.0$&$35.3$&$38.0$&$32.9$&$35.0$&$33.1$&$32.9$&$31.2$&$37.9$&$38.4$&$32.6$&$35.5$&$37.1$\\

    \cdashlinelr{1-19}

\multirow{2}{*}{$\text{BLOOM}_{\text{1.7B}}$ + Pre-train}
            & 0 & $49.4$&$40.8$&$46.7$&$42.6$&$43.5$&$42.4$&$33.7$&$33.5$&$35.1$&$34.3$&$37.4$&$32.7$&$42.6$&$38.6$&$34.9$&$38.6$&$39.2$\\

            & 4 & $45.0$&$37.3$&$40.4$&$37.8$&$42.4$&$38.8$&$35.4$&$34.0$&$35.2$&$33.7$&$34.3$&$33.4$&$37.5$&$36.8$&$34.4$&$36.6$&$37.1$\\

    \cdashlinelr{1-19}

\multirow{2}{*}{$\text{Branch-Train-Mix}$}
            & 0 & $49.7$&$43.4$&$46.0$&$41.4$&$44.1$&$41.4$&$33.8$&$33.7$&$34.5$&$35.8$&$35.4$&$31.8$&$44.5$&$38.0$&$35.5$&$37.1$&$39.1$\\

            & 4 & $44.7$&$38.4$&$41.2$&$37.3$&$41.5$&$38.2$&$35.3$&$34.4$&$34.9$&$33.6$&$33.8$&$33.6$&$37.8$&$35.9$&$34.3$&$34.1$&$36.8$\\

    \cdashlinelr{1-19}

\multirow{2}{*}{$\text{DMoE}$}
            & 0 & $50.6$&$42.6$&$45.4$&$44.3$&$43.1$&$41.3$&$34.6$&$33.5$&$35.2$&$37.1$&$35.1$&$34.3$&$44.7$&$39.2$&$36.2$&$39.7$&$\textbf{39.8}$\\

            & 4 & $44.9$&$36.9$&$40.6$&$38.2$&$41.8$&$39.2$&$37.4$&$33.9$&$36.0$&$36.2$&$35.9$&$32.8$&$37.7$&$36.8$&$34.5$&$36.7$&$\textbf{37.5}$\\

\bottomrule[1.2pt]
\end{tabu}

\caption{\label{tab:all_xnli} In-context learning results on XNLI across all languages. ``\textbf{High}'', ``\textbf{Medium}'' and ``\textbf{Low}'' denotes the available amount of linguistic resources. ${}^{\dagger}$ denotes the unseen language in the pre-training corpus of BLOOM.
}

\vspace{-2mm}

\end{table*}

%% file: tabs/in-context/pawsx.tex
\begin{table*}[thp]

\renewcommand\arraystretch{1.3}

\centering
\scriptsize

\setlength{\tabcolsep}{4mm}

 \begin{tabu}{l|c|cccccc|c|c}
 
 \toprule[1.2pt]
  \multicolumn{2}{c}{ } & \multicolumn{6}{c}{\textbf{High}} & \multicolumn{1}{c}{\textbf{Medium}} &\\
  \cmidrule(r){3-8} \cmidrule(r){9-9} \noalign{\smallskip}
\multicolumn{1}{c}{\textbf{Model}}& \textbf{\#shot} & de${}^{\dagger}$ & en  & es & fr & ja${}^{\dagger}$ & zh & ko${}^{\dagger}$  &\multicolumn{1}{c}{\textbf{Avg}}\\

   \midrule[0.8pt]
   
\multirow{2}{*}{$\text{BLOOM}_{\text{560M}}$}
            & 0 & $49.4$&$49.9$&$50.4$&$52.8$&$52.8$&$54.1$&$51.0$&$51.5$\\
            & 4 & $52.5$&$50.0$&$49.7$&$51.9$&$51.3$&$52.5$&$49.7$&$51.1$\\
    \cdashlinelr{1-10}

\multirow{2}{*}{$\text{BLOOM}_{\text{560M}}$ + Pre-train}
            & 0 & $49.7$&$50.2$&$50.7$&$54.8$&$55.7$&$55.1$&$54.1$&$52.9$\\
            & 4 & $50.9$&$51.5$&$51.1$&$51.9$&$50.3$&$54.1$&$51.3$&$51.6$\\

    \cdashlinelr{1-10}

\multirow{2}{*}{$\text{Branch-Train-Mix}$}
            & 0 & $51.3$&$49.1$&$50.9$&$54.6$&$56.2$&$55.3$&$54.7$&$53.1$\\
            & 4 & $48.3$&$50.5$&$53.0$&$52.3$&$50.1$&$55.1$&$50.4$&$51.4$\\
    \cdashlinelr{1-10}

\multirow{2}{*}{$\text{DMoE}$}
            & 0 & $51.4$&$51.1$&$51.2$&$54.6$&$54.9$&$55.2$&$54.0$&$\textbf{53.2}$\\
            & 4 & $51.3$&$50.1$&$52.4$&$52.8$&$51.8$&$54.8$&$52.2$&$\textbf{52.2}$\\
  \midrule[0.8pt]
\multirow{2}{*}{$\text{BLOOM}_{\text{1.7B}}$}
            & 0 & $52.6$&$53.8$&$50.7$&$54.9$&$55.7$&$54.8$&$54.8$&$53.9$\\
            & 4 & $48.9$&$50.4$&$49.8$&$51.4$&$50.1$&$50.4$&$52.8$&$50.5$\\
    \cdashlinelr{1-10}

\multirow{2}{*}{$\text{BLOOM}_{\text{1.7B}}$ + Pre-train}
            & 0 & $53.3$&$51.1$&$52.7$&$54.9$&$55.9$&$54.9$&$53.5$&$53.7$\\
            & 4 & $51.4$&$48.6$&$51.6$&$53.2$&$54.0$&$53.2$&$50.8$&$51.8$\\
    \cdashlinelr{1-10}

\multirow{2}{*}{$\text{Branch-Train-Mix}$}
            & 0 & $51.3$&$53.1$&$52.8$&$54.8$&$55.8$&$55.3$&$51.7$&$53.5$\\
            & 4 & $50.0$&$49.4$&$52.2$&$53.2$&$50.3$&$52.9$&$51.2$&$51.3$\\
    \cdashlinelr{1-10}

\multirow{2}{*}{$\text{DMoE}$}
            & 0 & $53.0$&$53.7$&$52.8$&$54.8$&$55.9$&$55.4$&$53.0$&$\textbf{54.1}$\\
            & 4 & $48.8$&$49.2$&$50.9$&$54.6$&$54.7$&$54.6$&$52.5$&$\textbf{52.2}$\\
\bottomrule[1.2pt]
\end{tabu}

\caption{\label{tab:all_pawsx} In-context learning results on PAWS-X across all languages. ``\textbf{High}'' and ``\textbf{Medium}'' denotes the available amount of linguistic resources. ${}^{\dagger}$ denotes the unseen language in the pre-training corpus of BLOOM.
}

\vspace{-2mm}

\end{table*}

%% file: tabs/in-context/xcopa.tex
\begin{table*}[thp]

\renewcommand\arraystretch{1.3}

\centering
\scriptsize

\setlength{\tabcolsep}{2.2mm}

% \vspace{2mm}

 \begin{tabu}{l|c|c|ccccc|ccc|cc|c}
 
 \toprule[1.2pt]
  \multicolumn{2}{c}{ } & \multicolumn{1}{c}{\textbf{High}} & \multicolumn{5}{c}{\textbf{Medium}} & \multicolumn{3}{c}{\textbf{Low}} & \multicolumn{2}{c}{\textbf{Ex-Low}} &\\
  \cmidrule(r){3-3} \cmidrule(r){4-8} \cmidrule(r){9-11} \cmidrule(r){12-13} \noalign{\smallskip}
\multicolumn{1}{c}{\textbf{Model}}& \textbf{\#shot} & zh & id & it${}^{\dagger}$  & th${}^{\dagger}$  & tr${}^{\dagger}$  & vi & et${}^{\dagger}$  & sw & ta & ht${}^{\dagger}$  & qu${}^{\dagger}$  &\multicolumn{1}{c}{\textbf{Avg}}\\

   \midrule[0.8pt]
\multirow{2}{*}{$\text{BLOOM}_{\text{560M}}$}
            & 0 & $57.6$&$60.0$&$52.4$&$53.0$&$52.8$&$61.0$&$48.0$&$52.4$&$56.4$&$50.8$&$49.0$&$53.9$\\

            & 4 & $57.4$&$60.6$&$50.2$&$53.0$&$50.6$&$59.2$&$49.4$&$50.6$&$56.4$&$51.6$&$48.6$&$53.4$\\

    \cdashlinelr{1-14}

\multirow{2}{*}{$\text{BLOOM}_{\text{560M}}$ + Pre-train}
            & 0 & $53.6$&$57.2$&$54.4$&$53.8$&$51.6$&$58.8$&$52.0$&$51.4$&$57.0$&$49.0$&$50.4$&$53.6$\\

            & 4 & $54.4$&$56.8$&$54.0$&$53.8$&$52.8$&$58.0$&$52.2$&$50.8$&$56.6$&$52.2$&$49.8$&$53.8$\\

    \cdashlinelr{1-14}

\multirow{2}{*}{$\text{Branch-Train-Mix}$}
            & 0 & $55.8$&$57.4$&$54.4$&$56.2$&$53.6$&$58.8$&$50.2$&$53.0$&$54.6$&$49.8$&$51.6$&$54.1$\\

            & 4 & $55.0$&$57.2$&$53.2$&$55.4$&$52.8$&$61.2$&$50.8$&$52.0$&$55.6$&$49.6$&$50.4$&$53.9$\\

    \cdashlinelr{1-14}

\multirow{2}{*}{$\text{DMoE}$}
            & 0 & $55.8$&$59.0$&$54.2$&$55.6$&$53.2$&$58.6$&$53.2$&$51.2$&$55.6$&$52.4$&$49.2$&$\textbf{54.4}$\\

            & 4 & $56.6$&$57.4$&$53.6$&$57.0$&$52.8$&$59.6$&$52.6$&$51.0$&$56.0$&$52.4$&$52.4$&$\textbf{54.7}$\\

  \midrule[0.8pt]
\multirow{2}{*}{$\text{BLOOM}_{\text{1.7B}}$}
            & 0 & $61.4$&$63.2$&$52.4$&$53.2$&$53.0$&$66.2$&$47.4$&$51.8$&$56.4$&$50.4$&$50.8$&$55.1$\\

            & 4 & $63.8$&$62.0$&$51.2$&$53.0$&$52.0$&$66.2$&$49.2$&$52.0$&$57.0$&$51.0$&$50.2$&$55.2$\\

    \cdashlinelr{1-14}

\multirow{2}{*}{$\text{BLOOM}_{\text{1.7B}}$ + Pre-train}
            & 0 & $58.6$&$61.4$&$52.6$&$55.0$&$52.4$&$61.8$&$49.6$&$54.2$&$56.0$&$53.4$&$50.0$&$55.0$\\

            & 4 & $60.4$&$61.6$&$53.0$&$55.2$&$51.4$&$63.4$&$50.0$&$54.8$&$56.2$&$51.8$&$50.0$&$55.3$\\

    \cdashlinelr{1-14}

\multirow{2}{*}{$\text{Branch-Train-Mix}$}
            & 0 & $58.6$&$61.2$&$55.2$&$55.2$&$54.2$&$62.6$&$51.0$&$52.6$&$55.2$&$54.0$&$51.4$&$55.6$\\

            & 4 & $59.6$&$61.4$&$56.0$&$53.8$&$52.2$&$63.8$&$50.8$&$54.0$&$56.6$&$50.4$&$51.8$&$55.5$\\

    \cdashlinelr{1-14}

\multirow{2}{*}{$\text{DMoE}$}
            & 0 & $59.6$&$62.8$&$54.2$&$56.2$&$54.8$&$63.6$&$51.2$&$54.2$&$56.0$&$52.6$&$51.2$&$\textbf{56.0}$\\

            & 4 & $60.8$&$60.2$&$53.6$&$56.2$&$53.8$&$63.0$&$51.2$&$53.4$&$55.2$&$54.8$&$50.6$&$\textbf{55.7}$\\

\bottomrule[1.2pt]
\end{tabu}

\caption{\label{tab:all_xcopa} In-context learning results on XCOPA across all languages. ``\textbf{High}'', ``\textbf{Medium}'', ``\textbf{Low}'' and ``\textbf{Ex-Low}'' denotes the available amount of linguistic resources. ${}^{\dagger}$ denotes the unseen language in the pre-training corpus of BLOOM.
}

\vspace{-2mm}

\end{table*}

%% file: tabs/in-context/xstorycloze.tex
\begin{table*}[thp]

\renewcommand\arraystretch{1.3}

\centering
\scriptsize

\setlength{\tabcolsep}{2.4mm}

% \vspace{2mm}

 \begin{tabu}{l|c|cccc|cc|ccc|cc|c}
 
 \toprule[1.2pt]
  \multicolumn{2}{c}{ } & \multicolumn{4}{c}{\textbf{High}} & \multicolumn{2}{c}{\textbf{Medium}} & \multicolumn{3}{c}{\textbf{Low}} & \multicolumn{2}{c}{\textbf{Ex-Low}} &\\
  \cmidrule(r){3-6} \cmidrule(r){7-8} \cmidrule(r){9-11} \cmidrule(r){12-13} \noalign{\smallskip}
\multicolumn{1}{c}{\textbf{Model}}& \textbf{\#shot} & en & es & ru${}^{\dagger}$  & zh & ar & id & hi & sw & te & eu & my${}^{\dagger}$  &\multicolumn{1}{c}{\textbf{Avg}}\\

   \midrule[0.8pt]
\multirow{2}{*}{$\text{BLOOM}_{\text{560M}}$}
            & 0 & $59.9$&$55.9$&$48.4$&$55.1$&$52.5$&$55.3$&$55.1$&$49.9$&$55.1$&$53.5$&$47.3$&$53.5$\\
            & 4 & $59.0$&$54.3$&$48.6$&$54.3$&$49.9$&$54.9$&$53.3$&$49.6$&$56.5$&$51.8$&$46.9$&$52.6$\\
    \cdashlinelr{1-14}

\multirow{2}{*}{$\text{BLOOM}_{\text{560M}}$ + Pre-train}
            & 0 & $59.1$&$55.1$&$51.6$&$54.0$&$50.4$&$55.6$&$54.7$&$52.9$&$55.3$&$54.2$&$49.4$&$53.8$\\
            & 4 & $57.8$&$54.3$&$49.3$&$53.9$&$48.5$&$53.7$&$53.5$&$52.5$&$54.8$&$52.8$&$48.4$&$52.7$\\
    \cdashlinelr{1-14}

\multirow{2}{*}{$\text{Branch-Train-Mix}$}
            & 0 & $59.2$&$56.1$&$51.8$&$53.6$&$50.6$&$55.2$&$53.8$&$52.5$&$55.7$&$54.9$&$48.2$&$53.8$\\
            & 4 & $58.5$&$54.6$&$51.4$&$53.5$&$49.4$&$54.7$&$53.6$&$51.6$&$55.7$&$52.9$&$47.9$&$53.1$\\
    \cdashlinelr{1-14}

\multirow{2}{*}{$\text{DMoE}$}
            & 0 & $59.0$&$56.3$&$51.4$&$54.7$&$51.0$&$55.7$&$54.3$&$52.7$&$55.8$&$54.7$&$49.0$&$\textbf{54.1}$\\
            & 4 & $58.6$&$54.9$&$50.5$&$54.3$&$50.6$&$54.5$&$53.8$&$52.2$&$55.7$&$53.9$&$48.4$&$\textbf{53.4}$\\
  \midrule[0.8pt]
\multirow{2}{*}{$\text{BLOOM}_{\text{1.7B}}$}
            & 0 & $64.4$&$61.0$&$50.3$&$58.1$&$54.8$&$59.9$&$56.9$&$52.1$&$56.6$&$54.9$&$47.0$&$56.0$\\
            & 4 & $65.1$&$61.7$&$50.0$&$58.2$&$53.7$&$59.0$&$56.5$&$51.8$&$55.4$&$53.3$&$45.9$&$55.5$\\
    \cdashlinelr{1-14}

\multirow{2}{*}{$\text{BLOOM}_{\text{1.7B}}$ + Pre-train}
            & 0 & $63.7$&$60.6$&$52.4$&$57.0$&$54.5$&$58.0$&$55.7$&$54.5$&$57.1$&$56.7$&$49.9$&$56.4$\\
            & 4 & $63.1$&$60.0$&$51.7$&$56.7$&$54.7$&$58.3$&$55.3$&$55.0$&$57.0$&$54.9$&$49.4$&$56.0$\\
    \cdashlinelr{1-14}

\multirow{2}{*}{$\text{Branch-Train-Mix}$}
            & 0 & $62.6$&$60.2$&$54.0$&$57.2$&$55.1$&$58.2$&$56.3$&$55.0$&$57.2$&$55.3$&$49.2$&$56.4$\\
            & 4 & $64.5$&$60.1$&$52.8$&$56.5$&$53.9$&$58.6$&$55.9$&$54.9$&$57.6$&$55.1$&$48.3$&$\textbf{56.2}$\\
    \cdashlinelr{1-14}

\multirow{2}{*}{$\text{DMoE}$}
            & 0 & $63.4$&$60.4$&$54.3$&$57.0$&$53.6$&$58.8$&$56.8$&$55.1$&$57.8$&$55.5$&$49.4$&$\textbf{56.6}$\\
            & 4 & $62.3$&$59.6$&$53.3$&$57.1$&$55.3$&$57.6$&$56.1$&$54.9$&$57.4$&$55.1$&$48.4$&$56.1$\\
\bottomrule[1.2pt]
\end{tabu}

\caption{\label{tab:all_xstorycloze} In-context learning results on XStoryCloze across all languages. ``\textbf{High}'', ``\textbf{Medium}'', ``\textbf{Low}'' and ``\textbf{Ex-Low}'' denotes the available amount of linguistic resources. ${}^{\dagger}$ denotes the unseen language in the pre-training corpus of BLOOM.
}

\vspace{-2mm}

\end{table*}

%% file: tabs/in-context/xwinograd.tex
\begin{table*}[thp]

\renewcommand\arraystretch{1.3}

\centering
\scriptsize

\setlength{\tabcolsep}{2mm}

% \vspace{2mm}

 \begin{tabu}{l|c|ccccc|c|c}
 
 \toprule[1.2pt]
  \multicolumn{2}{c}{ } & \multicolumn{5}{c}{\textbf{High}} & \multicolumn{1}{c}{\textbf{Medium}} &\\
  \cmidrule(r){3-7} \cmidrule(r){8-8} \noalign{\smallskip}
\multicolumn{1}{c}{\textbf{Model}}& \textbf{\#shot} & en & fr & ru${}^{\dagger}$  & zh & ja${}^{\dagger}$  & pt &\multicolumn{1}{c}{\textbf{Avg}}\\

   \midrule[0.8pt]
\multirow{2}{*}{$\text{BLOOM}_{\text{560M}}$}
            & 0 & $54.0$&$51.8$&$50.8$&$51.7$&$62.3$&$51.3$&$53.7$\\
            & 4 & $53.6$&$48.2$&$50.1$&$52.4$&$61.7$&$53.6$&$53.3$\\
    \cdashlinelr{1-9}

\multirow{2}{*}{$\text{BLOOM}_{\text{560M}}$ + Pre-train}
            & 0 & $52.9$&$55.4$&$50.2$&$52.1$&$63.1$&$55.9$&$54.9$\\
            & 4 & $53.7$&$48.2$&$51.1$&$53.7$&$64.5$&$51.7$&$53.8$\\
    \cdashlinelr{1-9}

\multirow{2}{*}{$\text{Branch-Train-Mix}$}
            & 0 & $52.9$&$54.2$&$49.8$&$54.9$&$61.5$&$52.9$&$54.4$\\
            & 4 & $54.5$&$51.8$&$49.4$&$55.9$&$62.5$&$51.0$&$54.2$\\
    \cdashlinelr{1-9}

\multirow{2}{*}{$\text{DMoE}$}
            & 0 & $53.2$&$51.8$&$50.7$&$57.8$&$63.3$&$53.6$&$\textbf{55.1}$\\
            & 4 & $53.7$&$55.4$&$50.2$&$55.6$&$64.3$&$51.3$&$\textbf{55.1}$\\
  \midrule[0.8pt]
\multirow{2}{*}{$\text{BLOOM}_{\text{1.7B}}$}
            & 0 & $55.7$&$50.6$&$50.8$&$54.3$&$65.9$&$53.2$&$55.1$\\
            & 4 & $56.1$&$51.8$&$51.8$&$54.3$&$66.7$&$52.1$&$55.5$\\
    \cdashlinelr{1-9}

\multirow{2}{*}{$\text{BLOOM}_{\text{1.7B}}$ + Pre-train}
            & 0 & $55.1$&$51.8$&$51.8$&$54.6$&$65.3$&$54.4$&$55.5$\\
            & 4 & $55.4$&$53.0$&$51.2$&$54.6$&$65.5$&$55.1$&$55.8$\\
    \cdashlinelr{1-9}

\multirow{2}{*}{$\text{Branch-Train-Mix}$}
            & 0 & $55.1$&$51.8$&$50.6$&$55.2$&$65.3$&$55.5$&$55.6$\\
            & 4 & $55.7$&$53.0$&$50.8$&$55.6$&$64.9$&$54.8$&$55.8$\\
    \cdashlinelr{1-9}

\multirow{2}{*}{$\text{DMoE}$}
            & 0 & $54.6$&$50.6$&$52.6$&$57.1$&$66.7$&$57.0$&$\textbf{56.4}$\\
            & 4 & $56.0$&$53.0$&$51.2$&$56.5$&$67.1$&$54.8$&$\textbf{56.4}$\\            
\bottomrule[1.2pt]
\end{tabu}

\caption{\label{tab:all_xwinograd} In-context learning results on XWinograd across all languages. ``\textbf{High}'' and ``\textbf{Medium}'' denotes the available amount of linguistic resources. ${}^{\dagger}$ denotes the unseen language in the pre-training corpus of BLOOM.
}

\vspace{-2mm}

\end{table*}

%% file: tabs/licenses.tex
\begin{table}[htp]

 \setlength{\tabcolsep}{2mm}
	\centering
	\small
	\renewcommand\arraystretch{1.25}
	\begin{center}
		\begin{tabular}{ll}
			\toprule[1.2pt]  
               \multicolumn{1}{c}{\textbf{Name}} & \multicolumn{1}{c}{\textbf{License}} \\
                \midrule[0.8pt]
                Transformers      & Apache 2.0 license     \\
                X-ELM      & Apache 2.0 license     \\
                lm-evaluation-harness      & MIT license      \\
                matplotlib      & PSF license      \\
                Bloom      & BigScience RAIL 1.0 license      \\
                Gemma      & Gemma license      \\
                CulturaX      & ODC-BY and CC0 license  \\
                MADLAD-400      & CC-BY-4.0 license  \\
			\bottomrule[1.2pt]
		\end{tabular}
	\end{center}
	\vspace{-2mm}
    \caption{\label{tab:license} Licenses of scientific artifacts involved in this work.}
	\vspace{-5mm}
\end{table}

%% file: tabs/lang_codes.tex
\begin{table*}[htp]

% \vspace{0.2cm}

\begin{minipage}[t]{0.49\linewidth}
       \setlength{\tabcolsep}{1mm}
	\centering
	\scriptsize
	\renewcommand\arraystretch{1.25}
	\begin{center}
		% \caption{Ablation study of different training methods on 5 datasets for $\text{XGLM}_{\text{564M}}$.}
		\begin{tabular}{ccc}
			\toprule[1.2pt]  
               \multicolumn{1}{c}{\textbf{ISO 639-1/2}} & \multicolumn{1}{c}{\textbf{Language}}    &\multicolumn{1}{c}{\textbf{Family}} \\
                \midrule[0.8pt]
                    am	&Amharic	&Afro-Asiatic, Semitic\\
                    ar${}^{\star}$	&Arabic	&Afro-Asiatic, Semitic\\
                    av	&Avaric	&Northeast Caucasian, Avar-Andic\\
                    az	&Azerbaijani	&Turkic, Common Turkic\\
                    be	&Belarusian	&Indo-European, Balto-Slavic\\
                    bn${}^{\star}$	&Bangla	&Indo-European, Indo-Iranian\\
                    bo	&Tibetan	&Sino-Tibetan, Tibeto-Burman\\
                    br	&Breton	&Indo-European, Celtic\\
                    ca	&Catalan	&Indo-European, Italic\\
                    ce	&Chechen	&Northeast Caucasian, Nakh\\
                    ceb	&Cebuano	&Austronesian, Malayo-Polynesian\\
                    ckb	&Central Kurdish	&Indo-European, Indo-Iranian\\
                    cnh	&Chin Haka	&Sino-Tibetan, Tibeto-Burman\\
                    co	&Corsican	&Indo-European, Italic\\
                    cs	&Czech	&Indo-European, Balto-Slavic\\
                    da	&Danish	&Indo-European, Germanic\\
                    de${}^{\star}$	&German	&Indo-European, Germanic\\
                    dv	&Divehi	&Indo-European, Indo-Iranian\\
                    ee	&Ewe	&Niger-Congo, Atlantic-Congo\\
                    el	&Greek	&Indo-European, Graeco-Phrygian\\
                    en	&English	&Indo-European, Germanic\\
                    eo	&Esperanto	&Indo-European, Italic\\
                    es	&Spanish	&Indo-European, Italic\\
                    et	&Estonian	&Uralic, Finno-Ugric\\
                    eu	&Basque	&Language isolate\\
                    fa	&Persian	&Indo-European, Indo-Iranian\\
                    fi	&Finnish	&Uralic, Finno-Ugric\\
                    fil	&Filipino	&Austronesian, Malayo-Polynesian\\
                    fo	&Faroese	&Indo-European, Germanic\\
                    fr${}^{\star}$	&French	&Indo-European, Italic\\
                    fy	&Western Frisian	&Indo-European, Germanic\\
                    gd	&Scottish Gaelic	&Indo-European, Celtic\\
                    gl	&Galician	&Indo-European, Italic\\
                    grc	&Ancient Greek	&Indo-European, Hellenic\\
                    gsw	&Swiss German	&Indo-European, Germanic\\
                    gu	&Gujarati	&Indo-European, Indo-Iranian\\
                    ha	&Hausa	&Afro-Asiatic, Chadic\\
                    haw	&Hawaiian	&Austronesian, Malayo-Polynesian\\
                    he	&Hebrew	&Afro-Asiatic, Semitic\\
                    hi${}^{\star}$	&Hindi	&Indo-European, Indo-Iranian\\
                    hil	&Hiligaynon	&Austronesian, Malayo-Polynesian\\
                    hmn	&Hmong	&Hmong-Mien, Hmongic\\
                    ht	&Haitian Creole	&French Creole, Circum-Caribbean French\\
                    hu	&Hungarian	&Uralic, Finno-Ugric\\
                    id${}^{\star}$	&Indonesian	&Austronesian, Malayo-Polynesian\\
                    ig	&Igbo	&Niger-Congo, Atlantic-Congo\\
                    ilo	&Iloco	&Austronesian, Malayo-Polynesian\\
                    is	&Icelandic	&Indo-European, Germanic\\
                    it${}^{\star}$	&Italian	&Indo-European, Italic\\
                    ja${}^{\star}$	&Japanese	&Japonic\\
                    jv	&Javanese	&Austronesian, Malayo-Polynesian\\
                    ka	&Georgian	&Kartvelian, Karto-Zan\\
                    kaa	&Karakalpak	&Turkic, Common Turkic\\
                    kbd	&Kabardian	&Northwest Caucasian, Circassian\\
                    kha	&Khasi	&Austroasiatic, Khasi-Palaungic\\
                    kk	&Kazakh	&Turkic, Common Turkic\\
                    kl	&Greenlandic	&Eskaleut, Eskimo\\
                    km	&Khmer	&Austroasiatic, Khmer\\
                    kn	&Kannada	&Dravidian, Proto-Dravidian\\
                    ko${}^{\star}$	&Korean	&Koreanic, Korean\\
                    ky	&Kyrgyz	&Turkic, Common Turkic\\
                    la	&Latin	&Indo-European, Italic\\
                    lb	&Luxembourgish	&Indo-European, Germanic\\
                    lg	&Ganda	&Niger-Congo, Atlantic-Congo\\
			\bottomrule[1.2pt]
		\end{tabular}
	\end{center}
	% \vspace{-5mm}
\end{minipage}
\begin{minipage}[t]{0.49\linewidth}
       \setlength{\tabcolsep}{2.5mm}
	\centering
	\scriptsize
	%\vspace{-0.4cm}
	\renewcommand\arraystretch{1.25}
	\begin{center}
		% \caption{Ablation study of different training methods on 5 datasets for $\text{BLOOM}_{\text{560M}}$.}
		\begin{tabular}{ccc}
			\toprule[1.2pt]  
               \multicolumn{1}{c}{\textbf{ISO 639-1/2}} & \multicolumn{1}{c}{\textbf{Language}}    &\multicolumn{1}{c}{\textbf{Family}} \\
                \midrule[0.8pt]
                    lo	&Lao	&Kra-Dai, Tai\\
                    lus	&Mizo	&Sino-Tibetan, Tibeto-Burman\\
                    lv	&Latvian	&Indo-European, Balto-Slavic\\
                    mg	&Malagasy	&Austronesian, Malayo-Polynesian\\
                    mi	&Maori	&Austronesian, Malayo-Polynesian\\
                    mk	&Macedonian	&Indo-European, Balto-Slavic\\
                    ml	&Malayalam	&Dravidian, Southern\\
                    mn	&Mongolian	&Mongolic, Central Mongolic\\
                    mr	&Marathi	&Indo-European, Indo-Iranian\\
                    ms	&Malay	&Austronesian, Malayo-Polynesian\\
                    mt	&Maltese	&Afro-Asiatic, Semitic\\
                    my	&Burmese	&Sino-Tibetan, Tibeto-Burman\\
                    ne	&Nepali	&Indo-European, Indo-Iranian\\
                    nl${}^{\star}$	&Dutch	&Indo-European, Germanic\\
                    no	&Norwegian	&Indo-European, Germanic\\
                    ny	&Chewa	&Niger-Congo, Atlantic-Congo\\
                    oc	&Occitan	&Indo-European, Italic\\
                    om	&Oromo	&Afro-Asiatic, Cushitic\\
                    os	&Ossetian	&Indo-European, Indo-Iranian\\
                    pa	&Punjabi	&Indo-European, Indo-Iranian\\
                    pap	&Papiamento	&Portuguese Creole, Afro-Portuguese\\
                    pl	&Polish	&Indo-European, Balto-Slavic\\
                    ps	&Pashto	&Indo-European, Indo-Iranian\\
                    pt	&Portuguese	&Indo-European, Italic\\
                    rm	&Romansh	&Indo-European, Italic\\
                    ro	&Romanian	&Indo-European, Italic\\
                    ru${}^{\star}$	&Russian	&Indo-European, Balto-Slavic\\
                    sa	&Sanskrit	&Indo-European, Indo-Iranian\\
                    sah	&Yakut	&Turkic, Common Turkic\\
                    sd	&Sindhi	&Indo-European, Indo-Iranian\\
                    se	&Northern Sami	&Uralic, Sami\\
                    sl	&Slovenian	&Indo-European, Balto-Slavic\\
                    sm	&Samoan	&Austronesian, Malayo-Polynesian\\
                    sn	&Shona	&Niger-Congo, Atlantic-Congo\\
                    so	&Somali	&Afro-Asiatic, Cushitic\\
                    sr	&Serbian	&Indo-European, Balto-Slavic\\
                    st	&Sotho	&Niger-Congo, Atlantic-Congo\\
                    su	&Sundanese	&Austronesian, Malayo-Polynesian\\
                    sw	&Swahili	&Niger-Congo, Atlantic-Congo\\
                    ta${}^{\star}$	&Tamil	&Dravidian, Southern\\
                    te${}^{\star}$	&Telugu	&Dravidian, Southern\\
                    tet	&Tetum	&Austronesian, Malayo-Polynesian\\
                    tg	&Tajik	&Indo-European, Indo-Iranian\\
                    th${}^{\star}$	&Thai	&Kra-Dai, Tai\\
                    ti	&Tigrinya	&Afro-Asiatic, Semitic\\
                    tk	&Turkmen	&Turkic, Common Turkic\\
                    to	&Tongan	&Austronesian, Malayo-Polynesian\\
                    tr	&Turkish	&Turkic, Common Turkic\\
                    ts	&Tsonga	&Niger-Congo, Atlantic-Congo\\
                    tt	&Tatar	&Turkic, Common Turkic\\
                    tyv	&Tuvan	&Turkic, Common Turkic\\
                    udm	&Udmurt	&Uralic, Permic\\
                    ug	&Uyghur	&Turkic, Common Turkic\\
                    uk${}^{\star}$	&Ukrainian	&Indo-European, Balto-Slavic\\
                    ur${}^{\star}$	&Urdu	&Indo-European, Indo-Iranian\\
                    uz	&Uzbek	&Turkic, Common Turkic\\
                    vec	&Venetian	&Indo-European, Italic\\
                    vi${}^{\star}$	&Vietnamese	&Austroasiatic, Vietic\\
                    xh	&Xhosa	&Niger-Congo, Atlantic-Congo\\
                    yi	&Yiddish	&Indo-European, Germanic\\
                    yo	&Yoruba	&Niger-Congo, Atlantic-Congo\\
                    yue	&Yue Chinese	&Sino-Tibetan, Sinitic\\
                    zh${}^{\star}$	&Chinese	&Sino-Tibetan, Sinitic\\
                    zu	&Zulu	&Niger-Congo, Atlantic-Congo\\
			\bottomrule[1.2pt]
		\end{tabular}
	\end{center}
	% \vspace{-5mm}
\end{minipage}

\caption{\label{tab:lang_codes} Details of language codes in this work. ${}^{\star}$ denotes the language used in the 18 languages experiment. 
}
% \vspace{-2mm}

\end{table*}